\documentclass[10pt,a4paper]{article}

\usepackage[utf8]{inputenc}
\usepackage[T1]{fontenc}
\usepackage{amsmath}
\usepackage{newtxtext,newtxmath}   
\usepackage{titlesec}
\usepackage{algorithm}
\usepackage{algpseudocode}
\usepackage{listings}
\usepackage{graphicx}
\usepackage[table]{xcolor}
\usepackage{colortbl}
\usepackage{booktabs}
\usepackage{array}
\newcolumntype{P}[1]{>{\raggedright\arraybackslash}p{#1}}
\usepackage{geometry}
\usepackage{hyperref}
\hypersetup{colorlinks=true, linkcolor=blue!55!black, citecolor=blue!55!black, urlcolor=blue!55!black}
\usepackage{enumitem}
\usepackage{caption}
\usepackage[numbers]{natbib}
\usepackage{float}
\usepackage{placeins}
\usepackage{tikz}
\usetikzlibrary{shapes.geometric, arrows.meta, positioning, fit, backgrounds, calc, shadows.blur}
\usepackage{pgfplots}
\pgfplotsset{compat=1.18}
\usepgfplotslibrary{groupplots}

\definecolor{inputbg}{HTML}{2C3E50}
\definecolor{outputbg}{HTML}{1ABC9C}
\definecolor{s1col}{HTML}{3498DB}
\definecolor{s2col}{HTML}{E67E22}
\definecolor{s3col}{HTML}{9B59B6}
\definecolor{s4col}{HTML}{27AE60}
\definecolor{s5col}{HTML}{E74C3C}
\definecolor{bugcol}{HTML}{FFC107}
\definecolor{cardbg}{HTML}{FAFAFA}
\definecolor{ontocol}{HTML}{8E44AD}
\definecolor{neo4jcol}{HTML}{16A085}
\definecolor{routecol}{HTML}{2980B9}
\tikzset{
  gcard/.style={rectangle, rounded corners=5pt, align=center, font=\footnotesize,
                minimum height=0.9cm},
  ginput/.style ={gcard, fill=inputbg,  draw=inputbg!80!black,  text=white, font=\bfseries\footnotesize},
  goutput/.style={gcard, fill=outputbg, draw=outputbg!80!black, text=white, font=\bfseries\footnotesize},
  gonto/.style  ={gcard, fill=ontocol!88, draw=ontocol!70!black, text=white},
  groute/.style ={gcard, fill=routecol!90, draw=routecol!70!black, text=white},
  gproc/.style  ={gcard, fill=s1col!14, draw=s1col!55, text=black!85},
  gproc2/.style ={gcard, fill=s3col!14, draw=s3col!55, text=black!85},
  gdata/.style  ={gcard, fill=cardbg,  draw=black!35, text=black!80},
  gmerge/.style ={gcard, fill=s4col!16, draw=s4col!60, text=black!85},
  greview/.style={gcard, fill=bugcol!20, draw=bugcol!70!black, text=black!85},
  gdistinct/.style={gcard, fill=black!6, draw=black!35, text=black!70},
  gdb/.style={cylinder, shape border rotate=90, aspect=0.25, draw=neo4jcol!70!black,
              fill=neo4jcol!16, text=black!82, align=center, font=\scriptsize,
              minimum width=2.3cm, minimum height=1.4cm},
  gdiamond/.style={diamond, aspect=2.2, draw=s2col!75!black, fill=s2col!16,
                   text=black!85, align=center, font=\scriptsize, inner sep=1pt},
  gnote/.style={rectangle, rounded corners=4pt, draw=black!30, fill=black!4,
                font=\scriptsize, align=center, inner sep=4pt},
  gflow/.style={-{Stealth[length=3mm, width=2mm]}, thick, draw=black!45},
}

\tikzset{
  base/.style={
    rectangle, rounded corners=5pt, minimum width=9cm, minimum height=1cm,
    align=center, font=\small
  },
  inputcard/.style  ={base, fill=inputbg,  draw=inputbg!80!black, text=white, font=\bfseries\small},
  outputcard/.style ={base, fill=outputbg, draw=outputbg!80!black, text=white, font=\bfseries\small},
  s1card/.style     ={base, fill=s1col!90, draw=s1col!70!black, text=white},
  s2card/.style     ={base, fill=s2col!90, draw=s2col!70!black, text=white},
  s3card/.style     ={base, fill=s3col!85, draw=s3col!70!black, text=white},
  s4card/.style     ={base, fill=s4col!85, draw=s4col!60!black, text=white},
  s5card/.style     ={base, fill=s5col!85, draw=s5col!70!black, text=white},
  bugcard/.style    ={rectangle, rounded corners=4pt, draw=bugcol!80!black, fill=bugcol!15,
                      text=black!80, font=\scriptsize, minimum width=4.5cm, align=center,
                      dashed, thick},
  stagebox/.style   ={rectangle, rounded corners=10pt, draw=#1!60, fill=#1!6,
                      inner sep=10pt},
  flowline/.style   ={-{Stealth[length=3mm, width=2mm]}, thick, draw=black!45},
  bugline/.style    ={-{Stealth[length=2.5mm]}, dashed, thick, draw=bugcol!70!black},
  stitle/.style     ={font=\bfseries\footnotesize, text=#1!80!black},
}

\makeatletter
\renewcommand\normalsize{%
  \@setfontsize\normalsize{10}{11}%
  \abovedisplayskip 9\p@ \@plus2\p@ \@minus5\p@
  \belowdisplayskip \abovedisplayskip
  \abovedisplayshortskip \z@ \@plus3\p@
  \belowdisplayshortskip 6\p@ \@plus3\p@ \@minus3\p@}
\makeatother
\normalsize
\titleformat{\section}
  {\normalfont\fontsize{12}{14}\selectfont\bfseries}{\thesection}{1em}{}
\titleformat{\subsection}
  {\normalfont\fontsize{10}{12}\selectfont\bfseries}{\thesubsection}{1em}{}
\titleformat{\subsubsection}
  {\normalfont\fontsize{10}{12}\selectfont\bfseries}{\thesubsubsection}{1em}{}
\titlespacing*{\section}{0pt}{2.0ex plus 0.5ex minus 0.2ex}{1.0ex plus 0.2ex}
\titlespacing*{\subsection}{0pt}{1.6ex plus 0.4ex minus 0.2ex}{0.8ex plus 0.2ex}
\titlespacing*{\subsubsection}{0pt}{1.3ex plus 0.3ex minus 0.2ex}{0.6ex plus 0.2ex}

\definecolor{codebg}{RGB}{245,245,245}
\definecolor{codegreen}{RGB}{0,128,0}
\definecolor{codegray}{RGB}{128,128,128}
\definecolor{codepurple}{RGB}{128,0,128}

\lstdefinestyle{pythonstyle}{
    backgroundcolor=\color{codebg},
    commentstyle=\color{codegreen},
    keywordstyle=\color{blue},
    stringstyle=\color{codepurple},
    basicstyle=\ttfamily\footnotesize,
    breaklines=true,
    frame=single,
    rulecolor=\color{codegray},
    numbers=left,
    numberstyle=\tiny\color{codegray},
    language=Python,
    showstringspaces=false,
    tabsize=4
}

\lstdefinestyle{jsonstyle}{
    backgroundcolor=\color{codebg},
    basicstyle=\ttfamily\footnotesize,
    breaklines=true,
    frame=single,
    rulecolor=\color{codegray},
    numbers=left,
    numberstyle=\tiny\color{codegray},
    showstringspaces=false,
    tabsize=2
}

\newif\ifanon
\anonfalse

\newcommand{\titletext}{An Ontology-Guided, Deduplication-Aware Extraction Layer for Knowledge Graph Construction from Heterogeneous Documents}

\newcommand{\makenipstitle}{%
  \par\noindent\hrule height 4pt\relax
  \vspace{0.25in}
  \begin{center}
    {\fontsize{17}{21}\selectfont\bfseries \titletext\par}
  \end{center}
  \vspace{0.25in}
  \par\noindent\hrule height 1pt\relax
  \vspace{0.30in}
  \begin{center}
    \ifanon
      {\fontsize{12}{14}\selectfont\bfseries Anonymous Author(s)}\\[2pt]
      {\fontsize{10}{12}\selectfont Affiliation\\Address\\\texttt{email}}\\[2pt]
      {\fontsize{10}{12}\selectfont Paper under double-blind review}
    \else
      \begin{tabular}{c@{\hskip 1.4em}c@{\hskip 1.4em}c}
        {\fontsize{12}{14}\selectfont\bfseries Vaibhav Dangaich} &
        {\fontsize{12}{14}\selectfont\bfseries Kevin Lewis} &
        {\fontsize{12}{14}\selectfont\bfseries Kundeshwar Pundalik} \\[3pt]
        {\fontsize{8.5}{10}\selectfont\texttt{vaibhavdangaich@gmail.com}} &
        {\fontsize{8.5}{10}\selectfont\texttt{kevin.lewis@konectu.in}} &
        {\fontsize{8.5}{10}\selectfont\texttt{kundeshwar@konectu.in}}
      \end{tabular}
    \fi
  \end{center}
  \vspace{0.25in}
}
\date{}

\begin{document}

\makenipstitle
\thispagestyle{empty}

\begin{center}
  {\bfseries\large Abstract}
\end{center}
\vspace{-0.5em}
\begin{quote}
Large language models extract entities and relationships from unstructured documents fluently but inconsistently: type vocabularies fracture across documents, the same person surfaces under several name variants, relationships duplicate, and distinct individuals who share a name risk silent conflation. This paper presents the design, implementation, and empirical refinement of a production extraction layer that converts a live document stream into a validated knowledge graph aligned to a formal ontology. The system consumes document metadata from Kafka, routes PDF, spreadsheet, Office, and image content through handlers built for each format, and extracts entities and relationships in two passes using a locally hosted Qwen3.5-9B model tuned on the ontology. Its distinguishing component is ontology-guided extraction: the relevant slice of a curated ontology is retrieved live from a graph database by embedding similarity and injected into the extraction prompt, reducing catalog overhead by about 94 percent relative to static domain slices. Extracted results then pass through a refinement pipeline of five stages: deterministic cleaning, merging across chunks, a second pass for relationships, six deduplication algorithms that require no model inference, and an embedding resolution subsystem whose conflict guard no similarity score can override. Evaluation on intelligence corpora improved search recall from roughly 70 to 95 percent with no false merges, and corrected seven classes of silent quality defect, ranging from a bug that truncated source text by a single character to the systematic duplication of entities that carried title prefixes.
\end{quote}

\vspace{0.4em}

\clearpage

\section{Introduction}

This paper presents an extraction system that transforms a real-time stream of heterogeneous documents into a validated, ontology-aligned knowledge graph. It operates as a real-time Kafka consumer \cite{kreps2011kafka} that ingests document metadata from an upstream ingestion service, retrieves the corresponding raw files from disk, and employs a locally hosted Large Language Model (LLM), deployed on a GCP virtual machine via vLLM \cite{kwon2023vllm} with an OpenAI-compatible API, to extract structured entities and relationships from unstructured text, PDF, and image content. The extraction model is \texttt{Konect-U/Qwen3.5-9B-AWQ-4bit-Ontology}, a 4-bit AWQ-quantised, ontology-tuned Qwen3.5-9B model, and ontology retrieval (Section~\ref{sec:ontology-guided}) and embedding-based resolution (Section~\ref{sec:embedding-resolution}) use the companion embedding model \texttt{Konect-U/Qwen3-Embedding-0.6B-Ontology}. This approach builds upon the growing body of work demonstrating the efficacy of LLMs for knowledge graph construction \cite{pan2024unifying, zhu2024llms_kg}. The system supports provider switching between these local self-hosted models and cloud-hosted models through environment configuration, enabling seamless fallback without code changes. The output is a validated JSON knowledge graph conforming to a strict Pydantic-enforced schema \cite{colvin2021pydantic}, enriched with entity deduplication metadata. This output is persisted both to the terminal for operator visibility and to a designated output directory as individual JSON files for downstream graph database ingestion.

The system is implemented in Python. An earlier iteration was organised as three monolithic files (\texttt{consumer.py}, a $\sim$1,860-line \texttt{model.py}, and \texttt{str\_op\_schema.py}); that design became difficult to extend as new document formats and an ontology-grounding stage were added, and the single \texttt{model.py} entangled prompting, chunking, multi-phase extraction, retry logic, and deduplication in one namespace. The current system is therefore refactored into a modular package (\texttt{src/extraction/}) of roughly fifty focused modules grouped by responsibility:

\begin{itemize}\sloppy
    \item \textbf{consumer/} (\texttt{kafka.py}, \texttt{record\_processor.py}, \texttt{output.py}): Kafka consumption, message normalisation, MIME/extension-based routing, bounded-concurrency worker pool, and sequence-ordered output buffering.
    \item \textbf{core/} (\texttt{entity\_extractor.py}, \texttt{relationship\_extractor.py}, \texttt{media\_extractor.py}, \texttt{plain\_text.py}, \texttt{prompts.py}): the two-phase LLM extraction pipeline, prompt construction, and relationship second-pass quality guards.
    \item \textbf{ontology/} (\texttt{catalog\_injection.py}, \texttt{graph\_retriever.py}, \texttt{context\_composer.py}): \emph{ontology-guided extraction}: live retrieval of the relevant ontology slice from a Neo4j ontology graph and its injection into the extraction prompt (Section~\ref{sec:ontology-guided}).
    \item \textbf{pdf/}, \textbf{xlsx/}, \textbf{docx/}, \textbf{pptx/}, \textbf{vision/}: per-format extraction handlers for born-digital and scanned PDF, spreadsheets, Word, PowerPoint, and images (Section~\ref{sec:multiformat}).
    \item \textbf{dedup/} (\texttt{alias\_resolver.py}, \texttt{name\_similarity.py}, \texttt{phonetic.py}, \texttt{candidates.py}, \texttt{merger.py}): the rule-based deduplication algorithms, now augmented with phonetic matching (Double Metaphone, Soundex, Jaro--Winkler).
    \item \textbf{resolution/} (\texttt{engine.py}, \texttt{blocking.py}, \texttt{scorer.py}, \texttt{embedder.py}, \texttt{decision.py}, \texttt{store.py}): an \emph{embedding-based entity-resolution} subsystem with FAISS blocking, multi-signal scoring, and a thresholded decision engine (Section~\ref{sec:embedding-resolution}).
    \item \textbf{parsing/} (\texttt{retry.py}, \texttt{json\_parser.py}, \texttt{result\_finalizer.py}, \texttt{entity\_type.py}, \texttt{relationship\_normalizer.py}) and \textbf{schemas/} (\texttt{extraction.py}, \texttt{disambiguation.py}): strict-retry parsing, result finalisation, and the Pydantic data models that enforce structural guarantees on every extraction result.
    \item \textbf{providers/} (\texttt{local.py}, \texttt{gemini.py}, \texttt{base.py}): pluggable LLM backends with environment-driven provider switching, per-request timeouts, and retry budgets.
\end{itemize}

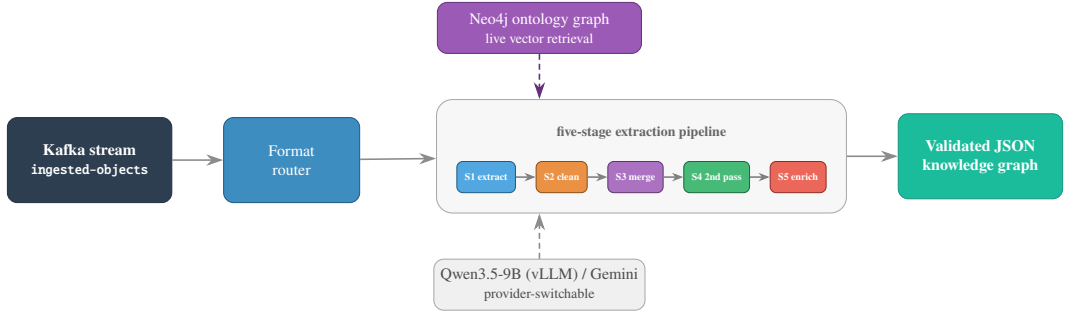
\begin{figure}[htbp]
\centering
\resizebox{\textwidth}{!}{%
\begin{tikzpicture}[node distance=0.55cm and 0.8cm,
  minichip/.style={rectangle, rounded corners=3pt, draw=#1!60!black, fill=#1!85,
                   text=white, font=\bfseries\tiny, inner sep=4pt, minimum height=0.55cm},
]
  \node[ginput, minimum width=2.9cm, minimum height=1.5cm] (kafka)
    {Kafka stream\\{\scriptsize\texttt{ingested-objects}}};
  \node[groute, minimum width=2.4cm, minimum height=1.5cm, right=0.9cm of kafka] (router)
    {Format\\router};

  \node[minichip=s1col, right=1.7cm of router, yshift=-0.3cm] (c1) {S1 extract};
  \node[minichip=s2col, right=0.35cm of c1] (c2) {S2 clean};
  \node[minichip=s3col, right=0.35cm of c2] (c3) {S3 merge};
  \node[minichip=s4col, right=0.35cm of c3] (c4) {S4 2nd pass};
  \node[minichip=s5col, right=0.35cm of c4] (c5) {S5 enrich};
  \node[font=\bfseries\scriptsize, text=black!60] (pipelabel)
    at ($(c1.north west)!0.5!(c5.north east)+(0,0.5)$) {five-stage extraction pipeline};
  \foreach \a/\b in {c1/c2, c2/c3, c3/c4, c4/c5}
    \draw[-{Stealth[length=2mm]}, thick, draw=black!50] (\a)--(\b);
  \begin{scope}[on background layer]
    \node[rectangle, rounded corners=8pt, draw=black!35, fill=black!3,
          inner sep=10pt, fit=(c1)(c5)(pipelabel)] (pipe) {};
  \end{scope}

  \node[goutput, minimum width=2.9cm, minimum height=1.5cm, right=0.9cm of pipe] (out)
    {Validated JSON\\knowledge graph};

  \node[gonto, minimum width=3.6cm, above=0.8cm of pipe, xshift=-1.8cm] (onto)
    {Neo4j ontology graph\\{\scriptsize live vector retrieval}};
  \node[gdistinct, minimum width=3.6cm, below=0.8cm of pipe, xshift=-1.8cm] (llm)
    {Qwen3.5-9B (vLLM) / Gemini\\{\scriptsize provider-switchable}};

  \draw[gflow] (kafka)--(router);
  \draw[gflow] (router)--(pipe);
  \draw[gflow] (pipe)--(out);
  \draw[gflow, densely dashed, draw=ontocol!70!black]
    (onto.south) -- ($(pipe.north)+(-1.8,0)$);
  \draw[gflow, densely dashed, draw=black!45]
    (llm.north) -- ($(pipe.south)+(-1.8,0)$);
\end{tikzpicture}%
}
\caption{High-level architecture of the extraction layer. The Kafka consumer routes each document by MIME type into the five-stage extraction pipeline (detailed in Fig.~\ref{fig:pipeline}); the ontology graph steers the type vocabulary at Stage~1; the LLM provider is switchable between the local vLLM endpoint and Gemini; the final output is a validated JSON knowledge graph.}
\label{fig:system-overview}
\end{figure}

\subsection{Contributions}

This paper makes four contributions. First, \emph{ontology-guided extraction} (Section~\ref{sec:ontology-guided}) retrieves the relevant slice of a curated ontology live from a graph database, using vector search over class definitions, and injects it into the extraction prompt, so the model emits types drawn from the formal schema rather than free-form labels, in the spirit of retrieval-augmented generation \cite{lewis2020rag}. Second, \emph{multi-format handling} (Section~\ref{sec:multiformat}) extends extraction beyond PDF and plain text to spreadsheets, Word, PowerPoint, and images, with a deterministic plan-then-execute strategy for tabular data and a six-signal per-page classifier that routes individual PDF pages to text, OCR, or skip paths. Third, a \emph{layered deduplication subsystem} (Section~\ref{sec:dedupres}) pairs six zero-inference rule-based algorithms with an embedding-based resolution stage whose hard-conflict guard no similarity score can override. Fourth, an \emph{empirical evaluation} (Sections~\ref{sec:eval-dedup}--\ref{sec:eval-ocr}) on intelligence-domain corpora quantifies the impact of these mechanisms and documents the upstream quality defects they expose and correct.

The entity deduplication system addresses a critical gap in knowledge graph construction: the handling of name variations and the prevention of false entity merges. This challenge has been extensively studied in the entity resolution literature \cite{christophides2021entity_resolution, papadakis2020blocking}, yet existing approaches typically require dedicated training data or pre-linked knowledge bases. Our contribution is a zero-overhead, rule-based deduplication pipeline that operates as a post-processing layer over LLM-extracted entities.

\paragraph{The Initial Challenge.}
Prior to this enhancement, the system exhibited several limitations:

\begin{itemize}
    \item Extracted entities possessed only a single alias (the base name as it appeared in the text).
    \item Spelling variations (e.g., ``John Doe'' vs.\ ``Jon Doe'') were instantiated as entirely distinct entities.
    \item The system lacked the capability to detect similarly named entities that might represent duplicates.
    \item Consequently, graph searches missed roughly a quarter of valid entity matches due to these name variations.
    \item Conversely, identical names appearing in disparate contexts were at risk of being incorrectly merged into a single entity (e.g., ``Elena Petrov,'' a combustion scientist at Khamsin Institute, and ``Elena Petrova,'' a malware analyst at Vektor Signal, could be falsely consolidated despite representing different individuals).
\end{itemize}

\paragraph{The Applied Solution.}
A six-stage post-extraction enhancement pipeline (detailed in Section~\ref{sec:dedup}) addresses these limitations: algorithmic alias expansion to five-plus variants per entity, fuzzy source-text mining for spelling variations \cite{ratcliff1988pattern}, linguistically aware similarity scoring \cite{knight1998transliteration}, context-validated duplicate detection that matches roles, organisations, and locations before any merge, qualifier-preserving relationship deduplication, and sequence-ordered parallel output buffering.

\paragraph{Results.} This methodology improved search recall from approximately 70\% to 95\% while maintaining a zero false-positive rate.


\section{Related work}

\paragraph{LLMs for knowledge graph construction.}
Since the introduction of the Transformer architecture \cite{vaswani2017attention} and the emergence of few-shot prompting in large models \cite{brown2020gpt3}, a growing body of work demonstrates that large language models can populate knowledge graphs directly from text \cite{pan2024unifying, zhu2024llms_kg}, whether through zero-shot prompting \cite{wei2023zeroshot_ie}, NER-style prompting \cite{wang2023gptner}, generative end-to-end relation extraction \cite{cabot2021rebel}, or revisited relation extraction \cite{wadhwa2023relation_extraction}. These studies consistently report the failure modes our system confronts in production: fragmented type vocabularies, hallucinated relations, and a bias toward entity description over relation extraction. Our contribution is not a new extraction model but an architecture that constrains and repairs a stock model's output at every stage.

\paragraph{Ontology grounding and retrieval augmentation.}
An ontology is a formal specification of a shared conceptualisation \cite{gruber1993ontology}; grounding extraction in such a schema is what keeps the emitted type vocabulary coherent. Retrieval-augmented generation \cite{lewis2020rag} conditions generation on retrieved free text; we apply the same principle to a formal class hierarchy, retrieving the relevant ontology slice by embedding similarity and injecting it into the extraction prompt. The centrality penalty we apply to generic classes adapts PageRank \cite{page1999pagerank} to schema retrieval. To our knowledge the combination of live graph retrieval, subclass expansion, and predicate full-text search for extraction-time grounding has not been described before.

\paragraph{Entity resolution and name matching.}
Entity resolution traces to the probabilistic record-linkage theory of Fellegi and Sunter \cite{fellegi1969theory} and has since matured into a broad field \cite{elmagarmid2007duplicate, christophides2021entity_resolution}, with blocking surveys \cite{papadakis2020blocking} and learned matchers from Magellan \cite{mudgal2018deep_entity} to transformer-based matchers \cite{brunner2020transformer}. Classical name-matching work covers personal-name abbreviation \cite{christen2006name_matching}, string-metric comparisons \cite{cohen2003string_metrics, navarro2001approximate}, the Jaro--Winkler measure our scorer relies on \cite{winkler1990string}, phonetic codes \cite{philips2000metaphone}, and cross-lingual matching \cite{knight1998transliteration, freeman2006crosslingual}; entity linking and disambiguation supply the contextual-evidence principle \cite{shen2015entity_linking, cucerzan2007disambiguation}. The embedding-based layer follows the modern practice of dense sentence representations for semantic similarity \cite{reimers2019sentencebert}. Unlike learned matchers, our six rule-based algorithms require no training data and run at zero inference cost, while the embedding-based layer adopts the canonical blocking--matching pipeline with a hard-conflict guard that no similarity score can override.

\section{System overview}

The extraction layer runs as a real-time Kafka consumer that receives document metadata from the upstream ingestion phase, resolves either file-based records (metadata pointers to files on disk) or database-embedded records (full text inline), and normalises both into a single internal representation before extraction. Consumer configuration, session-timeout tuning, and parallelism settings are operational rather than scientific concerns and are collected in Appendix~\ref{app:ops}. The subsections below describe the extraction strategy itself: the two-phase prompt design, the quality gate that protects the relationship pass, the chunking scheme, and the error-recovery behaviour.

\subsection{Two-phase extraction pipeline}

LLMs tend to prioritise entity descriptions over relationships \cite{wadhwa2023relation_extraction}, so extraction is split into two calls: Phase~1 extracts entities and preliminary relationships in the zero-shot paradigm \cite{wei2023zeroshot_ie, wang2023gptner}; Phase~2 re-reads the text alongside the extracted entity catalog and focuses solely on relationships, including temporal and contextual qualifiers. The outputs are consolidated by qualifier-preserving deduplication (Fig.~\ref{alg:rel-dedup}), after which Phase~3 (post-extraction enhancement, Figs.~\ref{alg:alias-expansion}--\ref{alg:context-dedup}) refines the entity data.

\subsection{Quality-gated relationship second pass}
\label{sec:quality-gate}

Ratio-based skip logic (``if $|\rho| \geq 50$, skip the second pass'') can false-trigger on documents where the existing relationships are disproportionately concentrated on one type or where entire relation families are structurally absent. To protect against these under-extraction cases, a \emph{quality-gate} evaluation precedes the skip decision, implementing four domain-informed checks:

\begin{enumerate}
    \item \textbf{Financial connectivity.} If financial entities are present alongside non-financial ones but zero cross-type relationships exist, the second pass is forced. This is a structurally common pattern where financial nodes appear as sources of funding but the LLM failed to connect them.
    \item \textbf{Financial relation types.} Even when the cross-type edges exist, if no relationship label matches \texttt{FUND|BUDGET|APPROV} (case-insensitive), the second pass is forced. This catches cases where financial relationships were extracted under non-standard labels that evade the cross-type edge check.
    \item \textbf{Person relation diversity.} When the entity set contains $\geq 8$ persons, fewer than 4 non-\texttt{REPORTED\_TO} person relationships triggers forcing. This prevents the known ``all roads lead to REPORTED\_TO'' LLM failure mode, where the model collapses all inter-person links into a single structural type.
    \item \textbf{Relationship type concentration.} If more than 30\% of relationships share a single type and fewer than 6 distinct types are present, the document likely received a shallow extraction that over-indexed on the most frequent signal in the text; the second pass is forced to surface the other families.
\end{enumerate}

The gate returns a \texttt{(should\_force, reasons)} pair; when forcing, the reasons list is logged so the forcing decision is auditable. Crucially, the gate \emph{complements} the ratio threshold: a document with 60 relationships of type \texttt{REPORTED\_TO} only still triggers check~3 and check~4, even though it clears the raw count. This heuristic-driven quality gate bridges the gap between shallow ``good enough'' metrics and domain-specific extraction completeness.

\subsection{Chunk-based extraction and merging}

For extensive documents, content is partitioned to respect LLM context limits. The chunking strategy differs by content type:

\paragraph{PDF and OCR text.} Pages are grouped into fixed-size windows (default \texttt{chunk\_pages=5}, one-page neighbor overlap on each side).

\paragraph{Plain text and database records.} Character-range chunking is used with configurable overlap (default: 8{,}000-char chunks, 1{,}000-char overlap). Each chunk is prefixed with an explicit overlap annotation, ``\texttt{Primary target chars: $s$--$e$. Overlap chars: 1000.}'', which tells the model it is looking at partially replicated context from the previous call, reducing relationship hallucination at chunk boundaries. This overlap-awareness matters: without the annotation, the model treats the overlap region as a novel document section and tends to re-extract entities and re-invent edges rather than recognising the continuation.

Following chunk-level extraction across all formats:
\begin{itemize}
    \item Entities are merged based on a canonical key.
    \item Relationship references are remapped to these canonical IDs.
    \item Phase~3 enhancements are applied to the consolidated entity set, ensuring that name variations discovered in any specific chunk propagate throughout the document's extracted data.
\end{itemize}

\subsection{Robustness and error recovery}

The pipeline includes retry mechanisms and automated JSON repair functions to handle malformed LLM outputs. The JSON parser handles both closed markdown fences (\verb|```json ... ```|) and unclosed fences (output truncated mid-stream), applying a three-stage recovery: closed-fence extraction $\to$ open-fence extraction $\to$ raw JSON search, so truncated LLM responses yield the maximum recoverable structure rather than a hard failure. When the main relationship extraction call fails due to entity count exceeding the model's capacity, a simplified fallback retries with the top-20 entities and a capped 30{,}000-character evidence window, recovering partial relationship coverage rather than returning an empty result.

\paragraph{Failure tracking for re-runs.}
Chunks that hit unrecoverable errors (e.g.\ context-window overflow on the local vLLM endpoint) are appended as self-contained JSONL records to a persistent failure log (\texttt{extraction\_failures.jsonl}), keyed by file path, object ID, stage, and the specific failed chunk or page indices. The log is written under a module-level mutex so concurrent workers never interleave partial records. Failed documents can be re-run as a batch from the log without re-scanning the full corpus, a critical operational property in production pipelines where transient GPU memory pressure may temporarily fail a document that would extract correctly on a retry.

Furthermore, the Phase~3 enhancement algorithms are designed to degrade gracefully; for instance, if properties are missing, duplicate detection proceeds with adjusted confidence levels, ensuring continued operation.

\section{Ontology-guided extraction via live graph retrieval}
\label{sec:ontology-guided}

\paragraph{The Problem with Un-Guided Extraction.}
A general-purpose LLM, asked to extract entities and relationships, invents its own type vocabulary on a per-document and even per-chunk basis. The same real-world class surfaces as \texttt{GovBody}, \texttt{Government Organization}, \texttt{Govt.\ Agency}, and \texttt{Ministry} across documents; relationship labels proliferate similarly. The downstream knowledge graph then carries a fractured, unaligned schema that frustrates ontology-conformant querying and analytics. To make extraction more solid and predictable, we constrain the model by injecting a description of the relevant ontology classes and predicates into the extraction prompt. The question is \emph{which} part of the ontology to inject, and \emph{how} to choose it; this is where the design evolved from static slices to live graph retrieval.

\paragraph{The First Approach: Static Catalog Slices.}
The initial design pre-compiled the ontology into a set of \emph{domain slices}. An offline build step (\texttt{build\_ontology\_catalog.py}) walked the ontology and emitted one prompt-ready text file per domain: \texttt{catalog\_cdr.txt}, \texttt{catalog\_south\_asia.txt}, a cross-domain \texttt{catalog\_bridge.txt} carrying shared predicates (\texttt{HAS\_PHONE}, \texttt{LOCATED\_IN}, \dots), and a full \texttt{catalog\_full.txt} for inspection. At extraction time a read-side loader (\texttt{catalog\_loader.py}) memoised each slice and estimated its size with a chars/4 token heuristic against a soft budget ($\sim$20{,}000 tokens; the worst routed case of three domains plus the bridge slice came to $\sim$17{,}700 tokens).

The \emph{selection} of slices per file was performed by a deterministic router (\texttt{file\_router.py}) implementing a three-tier keyword decision tree, comparing the file against a curated, domain-unique keyword vocabulary:
\begin{enumerate}
    \item \textbf{Strong filename match} (a domain keyword present in the filename, e.g.\ \texttt{07\_cdr\_data.csv} $\to$ \texttt{cdr}): trust the filename and skip the content scan.
    \item \textbf{Weak filename match} (an ambiguous single hit): read a $\sim$2\,KB content sample and combine filename and content signals.
    \item \textbf{No filename match}: scan content only; if that too is inconclusive, fall back to the bridge plus a broad regional slice for default coverage.
\end{enumerate}
The chosen slices were then concatenated into the system prompt, and a \texttt{FORCE\_CATALOG\_SLICES} override allowed pinning specific slices for testing. This was a clear improvement over un-guided extraction, since it bounded the type vocabulary and made outputs far more predictable, but it carried structural limitations:
\begin{itemize}
    \item \textbf{Coarse, all-or-nothing granularity.} Selection was at the domain level: the entire slice was injected even when a document touched only a handful of its classes, spending context budget on irrelevant types.
    \item \textbf{Brittle routing.} Keyword matching depends on filename conventions and a hand-curated vocabulary; opaque hashed filenames, multilingual content, or unanticipated synonyms misroute the file and inject the wrong (or default) slice.
    \item \textbf{Maintenance drift.} The slice text files are a materialised snapshot. Every ontology edit requires rerunning \texttt{build\_ontology\_catalog.py}; between rebuilds the injected catalog silently diverges from the live schema.
    \item \textbf{Budget pressure.} Injecting whole domains ($\sim$17.7\,K tokens in the worst case) crowds out document content in a 32\,K-token window.
\end{itemize}

\paragraph{The Current Approach: Live Graph Retrieval.}
We replace the static slice with \emph{live, content-conditioned retrieval} of the ontology, in the spirit of retrieval-augmented generation \cite{lewis2020rag} but targeting a formal class hierarchy rather than free text. A curated ontology is materialised as a Neo4j graph in which every class and predicate carries a natural-language definition and a pre-computed embedding. At extraction time, the relevant slice of that ontology is fetched on demand and injected into the system prompt, so the model is steered to emit types drawn from the formal schema. The mechanism (\texttt{ontology/graph\_retriever.py}, \texttt{catalog\_injection.py}, \texttt{context\_composer.py}) comprises five elements:

\begin{enumerate}
    \item \textbf{Multi-span windowing.} The content sample is split into overlapping windows ($\sim$1{,}500 characters, 300-character overlap, up to four windows) rather than embedded as one vector. A single vector over a multi-entity chunk collapses to a blended centroid that retrieves nothing specifically; per-window embeddings preserve local topicality.
    \item \textbf{Vector retrieval over class definitions.} Each window is embedded and used to query a Neo4j vector index (\texttt{db.index.vector.queryNodes}) over the ontology's class/predicate embeddings, returning candidate classes with their cosine score, definition, alternative labels, and graph centrality.
    \item \textbf{PageRank-penalised scoring.} High-centrality ancestor classes (\texttt{Entity}, \texttt{Organization}, \texttt{Location}) match almost everything and would dominate every retrieval. We down-weight them using a PageRank \cite{page1999pagerank} penalty,
    \[
        \text{score}(c) \;=\; \text{cos}(q, c)\,\bigl(1 - \lambda\,\text{pr}(c)\bigr), \qquad \lambda = 0.4,
    \]
    so that a generic class must clear a higher cosine bar than a specific one to be injected.
    \item \textbf{Dynamic cut-off under a token budget.} Rather than a fixed top-$k$, every class above an empirically motivated floor ($\text{min\_score}=0.72$) is taken, then trimmed to a token budget ($\sim$5{,}000 tokens, apportioned $70\%$ to classes and $30\%$ to predicates). A $0.5$ cosine floor admitted too many ``vaguely related'' classes; $0.72$ retains genuinely relevant ones.
    \item \textbf{Version-keyed caching and a circuit breaker.} Retrievals are cached under a key combining the content hash and an ontology-version hash refreshed every ten minutes, so editing the ontology transparently invalidates stale entries. If the graph is unreachable the circuit opens for sixty seconds and extraction proceeds \emph{un-guided} rather than blocking; ontology grounding is an enhancement, never a hard dependency.
\end{enumerate}

\paragraph{Novel-Type Flagging.}
Grounding is steering, not a hard constraint: the prompt instructs the model to use a catalogue type where one fits and otherwise to mark the emission as novel. After extraction, \texttt{flag\_novel\_types()} compares every emitted entity and relationship type against the full ontology vocabulary (after stripping pluralisation and wrapper suffixes such as \texttt{(records)}, \texttt{(entities)}) and sets \texttt{\_novel\_type}/\texttt{\_novel\_predicate} markers on those absent from the schema. These surface downstream as candidate ontology extensions for human review, so the schema can grow from real data without the model silently fragmenting the existing vocabulary.

\paragraph{Engineering Note.}
Because the Neo4j async driver and the embedding HTTP client bind to the event loop on which they are first used, the retriever runs its coroutines on a single dedicated daemon event loop (\texttt{\_run\_coro} with a 15-second timeout) rather than calling \texttt{asyncio.run()} per extraction, which would create a fresh loop each time and trigger ``future attached to a different loop'' errors from the cached singletons.

\begin{figure}[htbp]
\centering
\resizebox{\textwidth}{!}{%
\begin{tikzpicture}[node distance=0.5cm]
  \node[ginput, minimum width=6.4cm] (chunk) at (0,0) {Document chunk (content sample)};
  \node[gproc, minimum width=6.4cm, below=of chunk] (win)
    {Multi-span windowing $+$ embedding\\[-1pt]{\scriptsize $\sim$4 overlapping windows, 300-char overlap}};
  \node[gproc, minimum width=6.4cm, below=of win] (query)
    {Vector query over class/predicate embeddings\\[-1pt]{\scriptsize \texttt{db.index.vector.queryNodes}}};
  \node[gproc2, minimum width=6.4cm, below=of query] (score)
    {PageRank-penalised scoring\\[-1pt]{\scriptsize $\mathrm{score}(c)=\mathrm{cos}(q,c)\,\bigl(1-\lambda\,\mathrm{pr}(c)\bigr),\quad \lambda = 0.4$}};
  \node[gproc2, minimum width=6.4cm, below=of score] (cut)
    {Dynamic cut-off\\[-1pt]{\scriptsize keep $\geq 0.72$, trim to token budget ($\sim$5k)}};

  \node[goutput, minimum width=6.4cm] (out) at ($(chunk)+(9.9,0)$)
    {Ontology-aligned entities \& relationships};
  \node[greview, minimum width=6.4cm, below=of out] (novel)
    {Novel-type flagging\\[-1pt]{\scriptsize $\rightarrow$ proposed ontology extensions}};
  \node[groute, minimum width=6.4cm, below=of novel] (prompt)
    {Injected into extraction prompt\\[-1pt]{\scriptsize $\rightarrow$ LLM extraction}};
  \node[gonto, minimum width=6.4cm, below=of prompt] (cat)
    {Ontology catalog block\\[-1pt]{\scriptsize only the relevant classes $+$ predicates}};

  \node[gdb, minimum width=2.0cm] (db) at ($(query)+(4.95,0)$) {Neo4j\\ontology\\graph};
  \node[gnote, above=0.4cm of db] (note) {version cache\\$+$ circuit breaker};

  \foreach \a/\b in {chunk/win, win/query, query/score, score/cut}
    \draw[gflow] (\a)--(\b);
  \draw[gflow, rounded corners=8pt] (cut.south) -- ++(0,-0.5) -| (cat.south);
  \foreach \a/\b in {cat/prompt, prompt/novel, novel/out}
    \draw[gflow] (\a)--(\b);
  \draw[gflow] (query.east) -- (db.west);
  \draw[gflow] (db.south) |- (score.east);
  \draw[densely dashed, draw=black!35] (note)--(db);
\end{tikzpicture}%
}
\caption{Ontology-guided extraction via live graph retrieval, read as a U-shaped flow: the left column retrieves (window, embed, query, score, threshold), the bottom arrow hands the surviving classes across, and the right column injects (catalog block, prompt, novel-type flagging, aligned output). The Neo4j ontology graph sits between the columns; retrieval is cached by ontology version and protected by a circuit breaker, so a graph outage degrades to un-guided extraction rather than blocking.}
\label{fig:ontology-guided}
\end{figure}
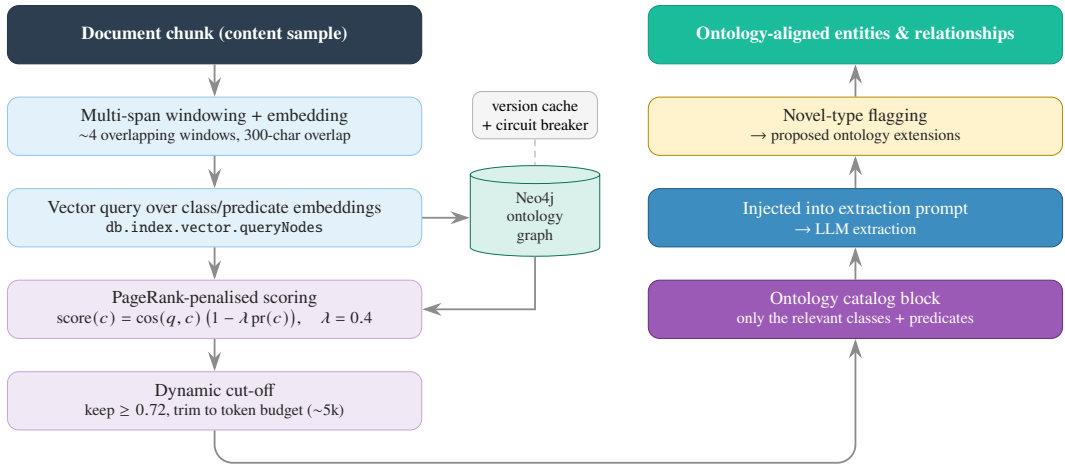

\paragraph{Impact.} Content-conditioned retrieval injects only the handful of classes a document actually needs, aligning emitted types with the formal ontology while keeping prompt overhead bounded. This is the ``grounding at extraction'' stage that lets the rest of the pipeline reason over a coherent schema rather than a bag of ad-hoc labels. Fig.~\ref{fig:ontology-guided} summarises the mechanism end-to-end.

\begin{table}[h]
\centering
\caption{Ontology injection: static catalog slices versus live graph retrieval.}
\label{tab:ontology-injection}
\begin{tabular}{@{}P{3.0cm}P{4.6cm}P{4.6cm}@{}}
\toprule
\rowcolor{ontocol!15}\textbf{Dimension} & \textbf{Static catalog slices (earlier)} & \textbf{Live graph retrieval (current)} \\
\midrule
Granularity        & Domain-level: the whole \texttt{catalog\_<domain>.txt} is injected & Class/predicate-level: only the matched classes \\
\addlinespace
Selection signal   & Filename $+$ $\sim$2\,KB content keyword match against a curated vocabulary & Per-window embedding similarity over class definitions \\
\addlinespace
Relevance ranking  & None within a slice (all-or-nothing) & Cosine score, PageRank-penalised, thresholded at $0.72$ \\
\addlinespace
Freshness          & Requires rebuilding slice files; drifts from the live schema & Always current; cached by ontology-version hash \\
\addlinespace
Token cost         & Whole domain(s); worst case $\sim$17.7\,K tokens & Bounded budget of only relevant classes ($\sim$5\,K) \\
\addlinespace
Failure mode       & Misroutes opaque filenames/synonyms; coarse over-injection & Degrades to un-guided extraction via the circuit breaker \\
\bottomrule
\end{tabular}
\end{table}

\paragraph{Measured Impact on Injection Size.}
The shift from domain slices to class-level retrieval produces a large, directly measurable reduction in the catalog prompt overhead. On a representative document, the static router selected multiple domain slices (the matched domains plus the cross-domain bridge slice), injecting roughly \textbf{11,200 tokens} of catalog text into the system prompt, most of it classes the document never used, carried only because they belonged to a selected domain. Live graph retrieval, by contrast, injected only the handful of classes and predicates the document's own content actually retrieved: approximately \textbf{700 tokens} for the same document. That is a $\sim$\textbf{94\% reduction} (about a $16\times$ smaller catalog block), summarised in Table~\ref{tab:injection-size}.

\begin{table}[h]
\centering
\caption{Catalog injection size for a representative document: domain slices versus graph retrieval.}
\label{tab:injection-size}
\small
\begin{tabular}{@{}lccc@{}}
\toprule
\rowcolor{s4col!15}\textbf{Metric} & \textbf{Static slices} & \textbf{Graph retrieval} & \textbf{Change} \\
\midrule
Catalog tokens injected     & $\sim$11{,}200 & $\sim$700  & $-$94\% ($\sim$16$\times$) \\
Selection unit              & whole domains $+$ bridge & matched classes only & class-level \\
Irrelevant classes carried  & many (domain padding) & none (thresholded) & eliminated \\
Context freed for content   & --- & $\sim$10{,}500 tokens & reclaimed \\
\bottomrule
\end{tabular}
\end{table}

This reduction matters on three fronts. First, in a 32\,K-token window, reclaiming $\sim$10{,}500 tokens lets substantially more of the source document share each extraction call, reducing chunking and the cross-chunk fragmentation it causes. Second, a tighter catalog is a stronger steer: the model is shown only classes that are genuinely relevant, so it is less likely to reach for a plausible-but-wrong neighbouring type that a broad domain slice would have placed in front of it, so extraction becomes both more predictable and more accurate. Third, the cost per call drops in proportion to the tokens removed. Crucially, the smaller block is not a truncation of the larger one: the $\sim$700 retained tokens are precisely the classes the content embeds closest to, whereas the $\sim$11,200-token slice was dominated by domain-mates that happened to be bundled together.

\subsection{Retrieval refinements: closing four precision gaps}
\label{sec:retrieval-refinements}

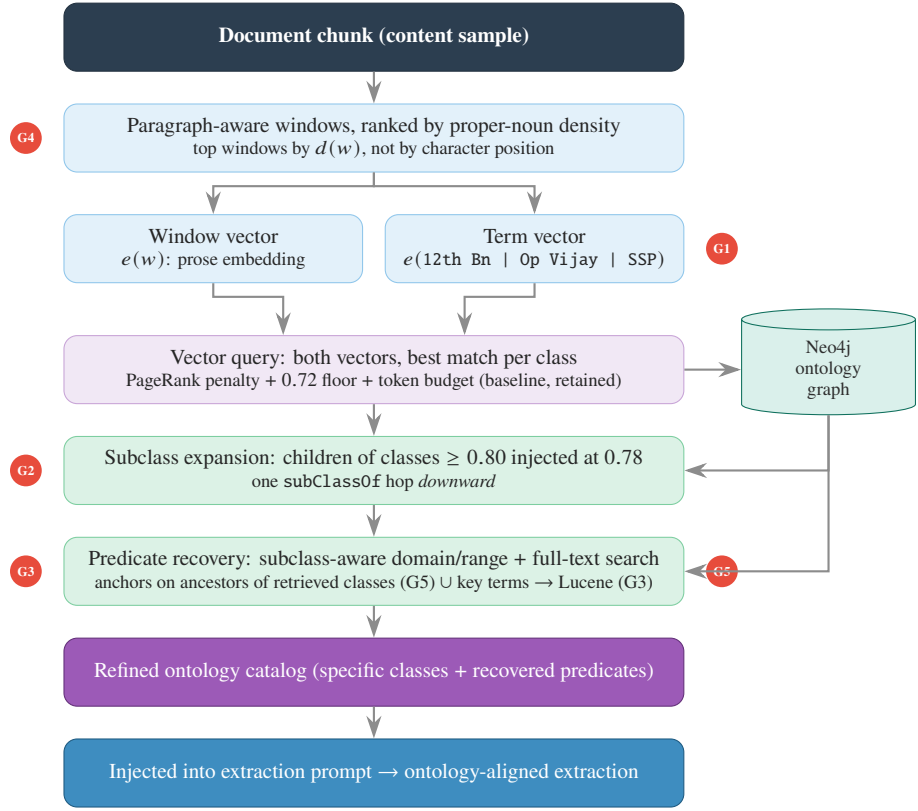
\begin{figure}[htbp]
\centering
\begin{tikzpicture}[node distance=0.42cm,
  badge/.style={circle, fill=s5col, text=white, font=\bfseries\tiny,
                inner sep=1.5pt, minimum size=0.42cm},
]
  \node[ginput, minimum width=8.2cm] (chunk) {Document chunk (content sample)};

  \node[gproc, minimum width=8.2cm, below=of chunk] (win)
    {Paragraph-aware windows, ranked by proper-noun density\\[-1pt]
     {\scriptsize top windows by $d(w)$, not by character position}};
  \node[badge, left=0.28cm of win] {G4};

  \node[gproc, minimum width=3.95cm, below=0.55cm of win, xshift=-2.125cm] (rawv)
    {Window vector\\[-1pt]{\scriptsize $e(w)$: prose embedding}};
  \node[gproc, minimum width=3.95cm, below=0.55cm of win, xshift=2.125cm] (termv)
    {Term vector\\[-1pt]{\scriptsize $e(\texttt{12th Bn | Op Vijay | SSP})$}};
  \node[badge, right=0.28cm of termv] {G1};

  \node[gproc2, minimum width=8.2cm, below=2.15cm of win] (query)
    {Vector query: both vectors, best match per class\\[-1pt]
     {\scriptsize PageRank penalty $+$ $0.72$ floor $+$ token budget (baseline, retained)}};

  \node[gmerge, minimum width=8.2cm, below=of query] (expand)
    {Subclass expansion: children of classes $\geq 0.80$ injected at $0.78$\\[-1pt]
     {\scriptsize one \texttt{subClassOf} hop \emph{downward}}};
  \node[badge, left=0.28cm of expand] {G2};

  \node[gmerge, minimum width=8.2cm, below=of expand] (pred)
    {Predicate recovery: subclass-aware domain/range $+$ full-text search\\[-1pt]
     {\scriptsize anchors on ancestors of retrieved classes (G5) $\cup$ key terms $\to$ Lucene (G3)}};
  \node[badge, left=0.28cm of pred] {G3};
  \node[badge, right=0.28cm of pred] {G5};

  \node[gonto, minimum width=8.2cm, below=of pred] (cat)
    {Refined ontology catalog (specific classes $+$ recovered predicates)};
  \node[groute, minimum width=8.2cm, below=of cat] (prompt)
    {Injected into extraction prompt $\rightarrow$ ontology-aligned extraction};

  \node[gdb, right=0.75cm of query] (db) {Neo4j\\ontology\\graph};

  \draw[gflow] (chunk)--(win);
  \draw[gflow] (win.south) -- ++(0,-0.18) -| (rawv.north);
  \draw[gflow] (win.south) -- ++(0,-0.18) -| (termv.north);
  \draw[gflow] (rawv.south) -- ++(0,-0.18) -| ($(query.north)+(-1.2,0)$);
  \draw[gflow] (termv.south) -- ++(0,-0.18) -| ($(query.north)+(1.2,0)$);
  \draw[gflow] (query)--(expand);
  \draw[gflow] (expand)--(pred);
  \draw[gflow] (pred)--(cat);
  \draw[gflow] (cat)--(prompt);
  \draw[gflow] (query.east) -- (db.west);
  \draw[gflow] (db.south) |- (expand.east);
  \draw[gflow] (db.south) |- (pred.east);
\end{tikzpicture}
\caption{The refined retrieval pipeline. The five refinements (red badges) slot into the baseline of Fig.~\ref{fig:ontology-guided} without altering its architecture: G4 re-orders \emph{what} gets embedded, G1 adds a second query vector per window, G2 widens class coverage downward from high-confidence matches, and G3 and G5 widen the two predicate routes: G3 opens a content-driven full-text route, while G5 makes the relational route subclass-aware so predicates anchored on generic ancestors still reach the catalog. All five are additive: disabling any refinement degrades precision, never availability.}
\label{fig:retrieval-refined}
\end{figure}

The mechanism of Fig.~\ref{fig:ontology-guided} retrieves the \emph{right neighbourhood} of the ontology, but operational evaluation on intelligence-domain documents exposed five systematic precision gaps: cases where a class or predicate that should have been injected fell out of the catalog, or where the injected classes were correct but insufficiently \emph{specific}. Each gap traces to one design decision in the baseline retriever, and each admits a targeted refinement (Table~\ref{tab:retrieval-gaps}). Fig.~\ref{fig:retrieval-refined} shows the refined pipeline with the five refinements marked in place.

\begin{table}[h]
\centering
\caption{Five precision gaps in the baseline retriever and their refinements.}
\label{tab:retrieval-gaps}
\begin{tabular}{@{}P{0.6cm}P{4.7cm}P{6.9cm}@{}}
\toprule
\rowcolor{s2col!15} & \textbf{Gap in the baseline} & \textbf{Refinement} \\
\midrule
G1 & Raw window text embeds to a blurred centroid; cosine distance to a sharply defined ontology label is inflated, dropping relevant classes below the 0.72 floor. & Extract proper nouns, abbreviations, and domain terms from each window and embed them as a second \emph{term vector}; query with both vectors and keep the best match per class. \\
\addlinespace
G2 & Hierarchy traversal looks only \emph{upward} (parent fetch); a high-confidence generic parent is injected without its more specific children, so the model emits \textsc{Organization} where \textsc{SecurityForce} was available. & For every class with adjusted score $\geq 0.80$, traverse \texttt{subClassOf} one hop \emph{downward} and inject the children at score $0.78$ (one additional Cypher call). \\
\addlinespace
G3 & Predicates reach the catalog only through domain/range edges to retrieved classes; predicates lacking those annotations are structurally invisible no matter how relevant. & Full-text search over \textsc{ObjectProperty} labels and definitions (a dedicated Lucene index) using the key terms extracted in G1, independent of domain/range connectivity. \\
\addlinespace
G4 & Windows are cut at fixed character offsets; a cut can split mid-paragraph, and entity-dense sections receive no priority over boilerplate. & Split on paragraph boundaries, score each window by proper-noun density, and embed the top windows \emph{by density} rather than by position. \\
\addlinespace
G5 & The relational route matches a predicate only against the \emph{exact} retrieved class labels, but predicates are anchored on generic parents (\textsc{subordinate to} on \textsc{Military Formation}) while retrieval returns specific subclasses (\textsc{Northern Light Infantry}), so the relevant predicate never matches. & Walk each retrieved class \emph{up} its \texttt{subClassOf} chain and match predicates whose domain or range is the class \emph{or any ancestor}, ordered by hop distance so the closest-anchored predicates win the token budget. \\
\bottomrule
\end{tabular}
\end{table}

\paragraph{G1: Term vectors alongside content vectors.}
A 1{,}500-character window such as ``\emph{The SSP directed the 12th Battalion along the LoC during Operation Vijay\,\dots}'' embeds into an average over narrative prose, while the ontology label \textsc{Military Formation} (definition: ``a body of troops organised for military purposes'') has a clean, focused vector. The two are about the same things, yet the prose centroid sits measurably farther from the label than it should, and borderline classes slip under the cosine floor. The refinement extracts the window's proper-noun phrases, abbreviations, and domain-adjacent terms into a compact pipe-delimited string (``\texttt{12th Battalion | Operation Vijay | SSP | LoC}''), embeds that string as a second query vector, and scores every candidate class against \emph{both} vectors, retaining the maximum:
\[
  \mathrm{score}(c) \;=\; \max_{q \,\in\, \{e(w),\, e(\mathrm{terms}(w))\}} \cos(q, c)\,\bigl(1 - \lambda\,\mathrm{pr}(c)\bigr).
\]
Term vectors match the terse register of ontology labels far more tightly than running prose does.

\paragraph{G2: Subclass expansion.}
Vector retrieval rewards the classes whose \emph{definitions} resemble the content, which systematically favours well-described general classes. If \textsc{Organization} retrieves at $0.85$, the baseline injects it, and only it, so the model, steered by the catalog, dutifully emits \textsc{Organization} even when \textsc{SecurityForce} or \textsc{TerroristGroup} is the correct, more informative type. The refinement treats a high-confidence match as evidence that its \emph{children} are worth showing: for every class with adjusted score $\geq 0.80$, one Cypher hop down the \texttt{subClassOf} hierarchy injects the direct children at a score of $0.78$ (just above the floor, below their parent). The model now sees the specific options alongside the general one and can choose the tighter fit.

\paragraph{G3: Predicate full-text search.}
The baseline reaches predicates only relationally: a predicate enters the catalog if its declared domain or range points at a retrieved class. Predicates without domain/range annotations (common among upper-ontology imports) can never be retrieved, regardless of how plainly the content calls for them. The refinement adds a second, content-driven route: a full-text index over \textsc{ObjectProperty} labels and definitions is queried with the key terms from G1, so a document about command relationships surfaces \texttt{controls}, \texttt{mediates-access-to}, and \texttt{operates} even when their schema annotations are absent. The route is additive and non-fatal: if the index is missing, retrieval proceeds on the relational route alone.

\paragraph{G4: Density-ranked windows.}
Fixed offsets at characters $0, 1200, 2400, 3600$ ignore both paragraph structure and information density: a window boundary can bisect a sentence, and the four windows selected may be summary boilerplate while the entity-rich passage at character 5{,}000 is never embedded. The refinement splits on paragraph boundaries, packs paragraphs greedily up to the window size, and scores each window by proper-noun density
\[
  d(w) \;=\; \frac{\left|\mathrm{propnoun\ chars}(w)\right| \;+\; 6\cdot\left|\mathrm{domain\ hits}(w)\right|}{|w|},
\]
embedding the top windows \emph{by density} instead of by position. Entity-dense sections, the ones that actually determine which ontology classes matter, always enter the embedding queue first.

\paragraph{G5: Subclass-aware predicate matching.}
G3 opens a content-driven route to predicates, but the original \emph{relational} route stays brittle for a subtler reason: it matches a predicate only when its declared domain or range is one of the \emph{exact} retrieved class labels. Predicates, however, are almost always anchored on a \emph{generic} parent (\textsc{subordinate to} declares its domain and range as \textsc{Military Formation}), whereas retrieval (sharpened by G1, G2, G4) returns the \emph{specific} subclasses the document names, such as \textsc{Force Command Northern Areas} and \textsc{Northern Light Infantry}. The exact-label test never intersects, so the single most relevant predicate for a military command document silently falls out, and the model, handed no canonical hierarchy predicate, fabricates its own (\texttt{SUPERIOR\_COMMANDER\_OF}, \texttt{COMMANDED\_SECTOR}). The refinement walks each retrieved class \emph{up} its \texttt{subClassOf} chain and matches predicates whose domain or range is the class \emph{or any ancestor}, ordering candidates by hop distance so the closest-anchored (most specific) predicates win the token budget. It is the predicate-side analogue of G2: where G2 expands classes one hop \emph{downward}, G5 matches predicates against the retrieved class's \emph{upward} ancestor chain.

\paragraph{Measured Effect.}
The refinements change \emph{which} classes reach the prompt, and the change is visible immediately on real documents. On a document describing intelligence-service support to a militant organisation, the baseline's strongest retrieval was the generic \textsc{Organization} at $0.74$; the refined retriever surfaces the named, specific classes the document is actually about (Table~\ref{tab:refinement-effect}), and the predicate route recovers relationship vocabulary (\texttt{controls}, \texttt{mediates-access-to}, \texttt{operates}) that the domain/range route alone never produced. The specific classes matter downstream: an extraction steered by \textsc{Inter-Services Intelligence} and \textsc{Military Formation} emits those types, where the baseline catalog would have flattened both to \textsc{Organization}.

The predicate side shows the sharpest gain. On a passage describing the Force Command Northern Areas (FCNA) command of Northern Light Infantry battalions, the baseline retrieved only \textbf{three} predicates, all tangential (\textsc{assigned to command}, \textsc{has joint operational area}); the subclass-aware route (G5) lifts this to \textbf{68}, surfacing the entire hierarchy family the document actually needs (\textsc{subordinate to}, \textsc{commands}, \textsc{part of}, \textsc{reports to}, \textsc{under command}), as Table~\ref{tab:predicate-recovery} reports. With those predicates in the catalog the extractor emits canonical relationship types instead of fabricating labels such as \texttt{SUPERIOR\_COMMANDER\_OF} or \texttt{COMMANDED\_SECTOR}, recovering canonical relationship vocabulary at the source rather than repairing it downstream.

\begin{table}[h]
\centering
\caption{Top retrieved classes for a representative intelligence document, before and after the refinements.}
\label{tab:refinement-effect}
\begin{tabular}{@{}lclc@{}}
\toprule
\rowcolor{s3col!12}\multicolumn{2}{c}{\textbf{Baseline retriever}} & \multicolumn{2}{c}{\textbf{Refined retriever}} \\
\cmidrule(r){1-2}\cmidrule(l){3-4}
\textbf{Class} & \textbf{Score} & \textbf{Class} & \textbf{Score} \\
\midrule
Organization          & 0.74 & Inter-Services Intelligence       & 0.806 \\
Location              & 0.73 & Operation Badr                    & 0.801 \\
Person                & 0.72 & 12 Northern Light Infantry        & 0.782 \\
---                   & ---  & Military Formation                & 0.780 \\
---                   & ---  & Security Force {\scriptsize(via G2)} & 0.78 \\
\bottomrule
\end{tabular}
\end{table}

\begin{table}[h]
\centering
\caption{Predicate retrieval on a military command passage (Force Command Northern Areas commanding the Northern Light Infantry), before and after the subclass-aware route (G5). The baseline's exact-label match returns three tangential predicates; G5 surfaces the canonical hierarchy family the document needs.}
\label{tab:predicate-recovery}
\begin{tabular}{@{}P{3.0cm}P{4.0cm}P{5.0cm}@{}}
\toprule
\rowcolor{s2col!12}\textbf{Metric} & \textbf{Baseline (exact-label)} & \textbf{Refined (subclass-aware, G5)} \\
\midrule
Predicates retrieved & 3 & 68 \\
\addlinespace
What surfaced & \textsc{assigned to command}, \textsc{assigned to air force command}, \textsc{has joint operational area} \emph{(all tangential)} & \textsc{subordinate to}, \textsc{commands}, \textsc{part of}, \textsc{reports to}, \textsc{under command}, \textsc{has subordinate formation}, \textsc{commanded by}, \textsc{deployed under command of}, \dots \\
\addlinespace
Hierarchy family present & none & complete \\
\bottomrule
\end{tabular}
\end{table}

\FloatBarrier
\subsection{From retrieval to canonical relationships: prompt guards and finalization conformance}
\label{sec:predicate-conformance}

The retrieval refinements of Section~\ref{sec:retrieval-refinements} put the right ontology classes and predicates \emph{in front of} the model, but a 9B local model is not bound by what it is shown: it still over-emits relationships, mislabels predicates, and mis-orients them. Three deterministic, ontology-grounded guards downstream of retrieval close that gap (one at the prompt, two at finalization), so emitted relationships conform to the schema regardless of the model's discipline. They are the relationship-side analogue of the entity \texttt{canonical\_iri} mapping (which the tagger applies to entity types but \emph{not} to predicates), and they compose as defense in depth: retrieval steers, the prompt restrains, finalization enforces.

\paragraph{Over-extraction: the fan-out guard.}
Asked to ``find ALL relationships,'' the model templates a pattern across every plausible entity pair. On an intelligence document about disputed territory it emitted a perfect $2\times58$ Cartesian grid (\emph{each} of two armies ``claiming'' \emph{every} one of 58 places), 116 edges (45\% of the document) from only two distinct sources, all sharing a single predicate and pointing in the inverted direction. The driver was the prompt itself: a licence to extract ``ANY\,\dots\,strongly implied connection'' together with a fan-out guard scoped only to reporting verbs. Two minimal edits (replacing ``strongly implied'' with ``explicitly stated in the text'' and generalising the fan-out guard to \emph{any} relation type with an explicit no-grid clause) collapsed it (Table~\ref{tab:overextraction}). This is a prompt-level guard: cheap, but probabilistic. On a small model it sharply reduces rather than provably eliminates the pattern, which is why the finalization guards below do not depend on it.

\begin{table}[h]
\centering
\caption{Relationship over-extraction on a disputed-territory document, before and after the prompt fan-out guard.}
\label{tab:overextraction}
\begin{tabular}{@{}lrr@{}}
\toprule
\rowcolor{s5col!12}\textbf{Metric} & \textbf{Before} & \textbf{After} \\
\midrule
\texttt{CLAIMED\_BY} edges (a $2\times58$ grid) & 116  & 1 \\
Total relationships                            & 258  & 112 \\
Most-concentrated predicate share              & 45\% & 16\% \\
Distinct predicate types                       & 63   & 29 \\
\bottomrule
\end{tabular}
\end{table}

\paragraph{Direction: rank-guarded orientation.}
Asymmetric hierarchy predicates (\textsc{subordinate to}, \textsc{part of}, \textsc{commands}) carry a fixed semantic direction, but the model orders \texttt{source}/\texttt{target} inconsistently: the same corpus yielded both ``\textsc{NLI subordinate to FCNA}'' (correct) and ``\textsc{FCNA subordinate to NLI}'' (inverted), plus the inverse predicate \textsc{has subordinate formation} pointing the wrong way. Because no pipeline stage ever swaps endpoints, a reversed edge is the model's error, faithfully preserved into the graph and then narrated back by the chatbot as a contradiction. The guard canonicalises the predicate label (folding inverse forms such as \textsc{has subordinate formation} and \textsc{commands} into the upward \textsc{subordinate to} with endpoints swapped) and then \emph{re-orients by ontology rank}: for an upward predicate the source must be the junior formation, where rank is a coarse tier read from the entity name (Command/Corps $>$ Division $>$ Brigade $>$ Battalion/Regiment). Whatever order the model emits (flipped, inverse, or correct), the edge converges to one canonical form (Table~\ref{tab:direction-conform}). The guard fires only on organisation$\leftrightarrow$organisation edges, so person-role edges (\textsc{commander of}) are left untouched.

\begin{table}[h]
\centering
\caption{Rank-guarded direction normalization: every phrasing of the FCNA/NLI command hierarchy converges to one canonical edge.}
\label{tab:direction-conform}
\begin{tabular}{@{}P{6.8cm}P{5.8cm}@{}}
\toprule
\rowcolor{s4col!12}\textbf{Model emits} & \textbf{After conformance} \\
\midrule
\texttt{FCNA -[subordinate to]-> NLI} {\scriptsize\emph{(flipped)}} & \texttt{NLI -[subordinate to]-> FCNA} \\
\addlinespace
\texttt{NLI -[has subordinate formation]-> FCNA} {\scriptsize\emph{(inverse)}} & \texttt{NLI -[subordinate to]-> FCNA} \\
\addlinespace
\texttt{FCNA -[commands]-> NLI} {\scriptsize\emph{(downward)}} & \texttt{NLI -[subordinate to]-> FCNA} \\
\addlinespace
\texttt{NLI -[subordinate to]-> FCNA} {\scriptsize\emph{(already right)}} & \texttt{NLI -[subordinate to]-> FCNA} \\
\bottomrule
\end{tabular}
\end{table}

\paragraph{Vocabulary: ontology-grounded canonicalization with novel flagging.}
The tagger maps entity types to ontology classes but leaves relationship predicates untouched (it only \emph{validates} their domain/range), so morphological and lexical variants of one relation proliferate: a single document carried \texttt{CAPTURED\_LOCATIONS} ($\times$14) and \texttt{CAPTURED\_LOCATION} ($\times$9) as two distinct edge types, neither of which the tagger would merge (its label match fails on the plural). The guard snaps each emitted predicate to the ontology object-property vocabulary (954 label/alt-label keys loaded once from the ontology graph) by a tiered, meaning-preserving match: exact, then de-pluralised (\texttt{CAPTURED\_LOCATIONS}$\to$\texttt{CAPTURED\_LOCATION}), then an embedding cosine against the ontology's predicate embeddings. The embedding tier is deliberately conservative: a calibration showed that below $\approx\!0.80$ cosine, generic upper-ontology predicates (\textsc{may-be-detected-by}, \textsc{attack-may-be-countered-by}) act as semantic magnets and produce confident-but-wrong matches, so the floor is set at $0.80$: only near-certain synonyms (\textsc{observed by}$\to$\textsc{observed-by-at-some-time}) snap, and everything else is left verbatim. Critically, an unmatched predicate is \emph{not} forced: it is marked \texttt{\_novel\_predicate} and the post-ingest validator flags it \textsc{unknown\_predicate} for review in the admin panel, where it can be promoted into the ontology, so the schema grows from real data instead of the model silently fragmenting it. Running this at extraction, before ingest, means the relationship-deduplication key \texttt{(source, target, type)} sees canonical types, so the 14 and 9 variants collapse into a single qualifier-merged edge rather than entering the graph as duplicates.

\section{Multi-format document handling}
\label{sec:multiformat}

The earlier system handled plain text and PDF. The current consumer routes by MIME type and file extension to format-specific handlers, unifying every format onto the same two-phase extraction and finalisation backend.

\paragraph{Spreadsheets (XLSX): Plan-Then-Execute.}
Per-row LLM extraction over a spreadsheet is both wasteful and inconsistent. Instead, the spreadsheet handler uses a deterministic \emph{plan-then-execute} strategy: (1)~\emph{inspect} the workbook to build a bounded sample (header tokens, merged-cell ranges, $\sim$10 sample rows per sheet); (2)~issue a \emph{single} LLM call that returns a typed \texttt{ExtractionPlan} listing, per sheet, the entity type, the name-bearing column, a list of \texttt{ColumnMap}s (each with a \texttt{value\_type} $\in \{$string, int, float, date$\}$), relationship definitions, and drop-row filters; (3)~\emph{execute} the plan deterministically over all rows, with numeric coercion, merged-cell propagation, multi-value splitting, and conditional skipping; and (4)~\emph{map} the resulting records onto \texttt{Entity}/\texttt{Relationship} objects. A \texttt{date} \texttt{value\_type} triggers date parsing and emits \emph{coverage warnings} for cells that fail to parse, so silent data loss in date columns becomes a visible, reviewable signal rather than a dropped field. This yields one LLM call per workbook template instead of one per row, with reproducible execution.

\paragraph{Office Documents (DOCX, PPTX): Virtual Pages.}
Word and PowerPoint files are converted into ``virtual pages'' so they can reuse the PDF text pipeline unchanged. the Word handler prefers Docling for layout-aware Markdown (falling back to \texttt{python-docx}), flattens tables to pipe-delimited rows, and groups content into $\sim$3{,}600-character virtual pages. the PowerPoint handler emits one virtual page per slide: title as a heading, body shapes in spatial reading order, hyperlinks inlined, tables as pipe-rows, and speaker notes as a separate page. For both, if the extracted text falls below a floor ($\sim$200 characters, indicating an image-only document) the handler extracts and OCRs embedded images before chunking. The virtual pages then flow through the same chunked text pipeline as a born-digital PDF.

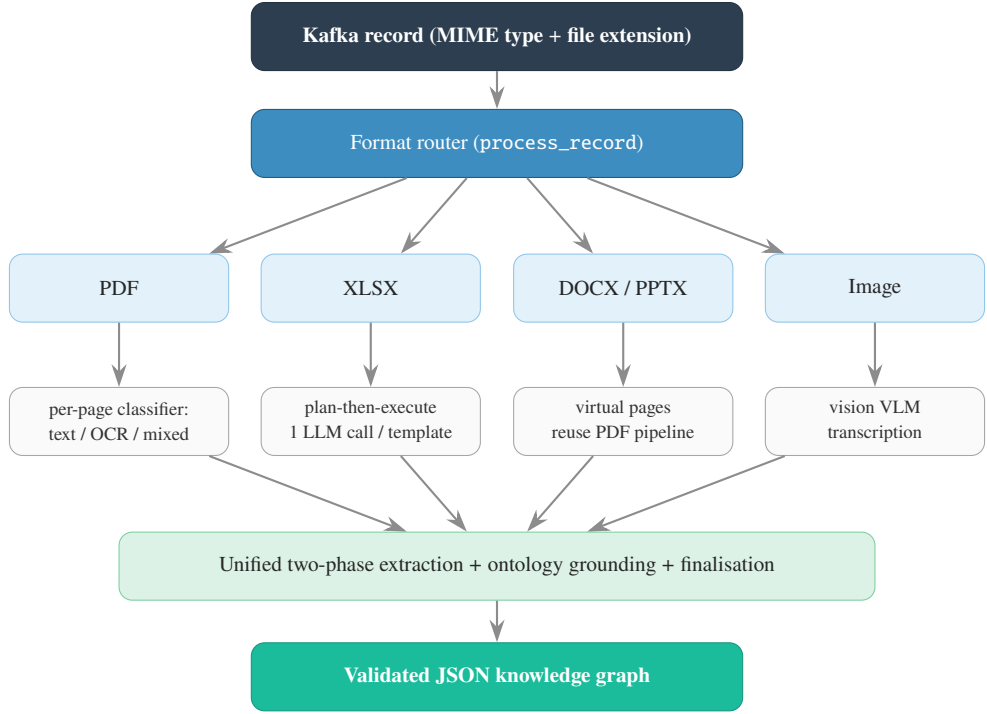
\begin{figure}[htbp]
\centering
\begin{tikzpicture}[node distance=0.85cm and 0.35cm]
  \node[ginput, minimum width=6.5cm] (rec) {Kafka record (MIME type $+$ file extension)};
  \node[groute, minimum width=6.5cm, below=0.5cm of rec] (router) {Format router (\texttt{process\_record})};

  \node[gproc, minimum width=2.9cm, below=1.0cm of router, xshift=-5.0cm] (pdf)    {PDF};
  \node[gproc, minimum width=2.9cm, below=1.0cm of router, xshift=-1.67cm] (xlsx)  {XLSX};
  \node[gproc, minimum width=2.9cm, below=1.0cm of router, xshift=1.67cm] (office) {DOCX / PPTX};
  \node[gproc, minimum width=2.9cm, below=1.0cm of router, xshift=5.0cm] (img)     {Image};

  \node[gdata, minimum width=2.9cm, below=of pdf]    (pdf2)    {\scriptsize per-page classifier:\\\scriptsize text / OCR / mixed};
  \node[gdata, minimum width=2.9cm, below=of xlsx]   (xlsx2)   {\scriptsize plan-then-execute\\\scriptsize 1 LLM call / template};
  \node[gdata, minimum width=2.9cm, below=of office] (office2) {\scriptsize virtual pages\\\scriptsize reuse PDF pipeline};
  \node[gdata, minimum width=2.9cm, below=of img]    (img2)    {\scriptsize vision VLM\\\scriptsize transcription};

  \node[gmerge, minimum width=10cm, below=1.0cm of xlsx2, xshift=1.67cm] (unify)
    {Unified two-phase extraction $+$ ontology grounding $+$ finalisation};
  \node[goutput, minimum width=6.5cm, below=0.55cm of unify] (out) {Validated JSON knowledge graph};

  \draw[gflow] (rec)--(router);
  \foreach \b in {pdf,xlsx,office,img} \draw[gflow] (router)--(\b);
  \draw[gflow] (pdf)--(pdf2);
  \draw[gflow] (xlsx)--(xlsx2);
  \draw[gflow] (office)--(office2);
  \draw[gflow] (img)--(img2);
  \foreach \b in {pdf2,xlsx2,office2,img2} \draw[gflow] (\b)--(unify);
  \draw[gflow] (unify)--(out);
\end{tikzpicture}
\caption{Multi-format routing. The consumer dispatches each record by MIME type and extension to a format-specific handler (per-page-classified PDF, plan-then-execute spreadsheets, virtual-page Office documents, or vision-transcribed images), and every path converges on the same two-phase extraction, ontology grounding, and finalisation backend, so downstream processing is format-agnostic.}
\label{fig:multiformat}
\end{figure}

\paragraph{PDF and Images.}
The PDF path (Section~\ref{sec:ocr-routing}) classifies each page and routes born-digital pages through layout-aware text extraction and scanned pages through the vision/OCR pipeline; standalone images go directly to vision. All paths converge on the same finalisation stage, so deduplication, type normalisation, and ontology grounding apply uniformly regardless of source format. Fig.~\ref{fig:multiformat} shows the routing topology.

\subsection{Per-page OCR classification and PDF routing}
\label{sec:ocr-routing}

A critical limitation of the initial PDF handling strategy was its binary treatment of documents: a PDF was either routed entirely through text extraction or entirely through OCR/vision processing. In practice, many real-world PDFs are \emph{mixed}, containing both born-digital text pages and scanned or image-heavy pages within the same document. To address this, we implemented a per-page OCR classification system using PyMuPDF \cite{pymupdf2024} that analyses six structural signals for each page and routes individual pages to the optimal extraction path.

\subsubsection{Per-page OCR classification}

\paragraph{Problem.} Treating an entire PDF as either text or OCR causes information loss: routing a mixed PDF entirely through text extraction silently drops scanned pages, while routing it entirely through OCR wastes compute on born-digital pages and may degrade text extraction quality.

\paragraph{Solution.} A six-signal per-page classifier using PyMuPDF's low-level page analysis API determines whether each page should be processed via text extraction, OCR/vision, or skipped entirely.

Six structural signals are measured per page (Table~\ref{tab:ocr-signals}), and the classification is a priority-ordered cascade over them: image dominance is checked first, then extractable text, then the vector-text edge case, with $\texttt{skip}$ as the final default:
\begin{equation}
  \mathrm{classify}(p) \;=\;
  \begin{cases}
    \texttt{skip} & \text{if } A_p = 0,\\
    \texttt{ocr}  & \text{if } I_{\text{block}} > 0.15 \;\vee\; I_{\text{xref}} > 0.15,\\
    \texttt{text} & \text{if } C > 50,\\
    \texttt{ocr}  & \text{if } D > 200,\\
    \texttt{text} & \text{if } C > 0,\\
    \texttt{skip} & \text{otherwise.}
  \end{cases}
  \label{eq:ocr-cascade}
\end{equation}

\begin{table}[h]
\centering
\caption{The six per-page signals and their roles in the cascade of Eq.~\eqref{eq:ocr-cascade}.}
\label{tab:ocr-signals}
\small
\begin{tabular}{@{}clllc@{}}
\toprule
\rowcolor{s3col!12}\# & \textbf{Signal} & \textbf{Definition} & \textbf{Detects} & \textbf{Threshold} \\
\midrule
1 & $A_p$               & page area                                   & degenerate pages       & $= 0$ \\
2 & $I_{\text{block}}$  & image-block area $/\,A_p$                   & scanned pages          & $> 0.15$ \\
3 & $I_{\text{xref}}$   & xref-image area $/\,A_p$                    & images hidden from block analysis & $> 0.15$ \\
4 & $C$                 & extractable characters                      & born-digital text      & $> 50$ (or $> 0$) \\
5 & $D$                 & drawing/path primitives                     & text drawn as vectors  & $> 200$ \\
6 & $X$                 & xref image count                            & gates signal 3's computation & $> 0$ \\
\bottomrule
\end{tabular}
\end{table}

\begin{figure}[htbp]
\centering
\begin{tikzpicture}[
  page/.style={rectangle, draw=black!45, minimum width=0.92cm, minimum height=1.25cm,
               font=\bfseries\scriptsize, inner sep=1pt, align=center},
]
  \node[font=\footnotesize\bfseries, text=black!60, anchor=west] at (-0.5, 1.15)
    {A mixed 10-page PDF under per-page classification:};
  \foreach \i/\fill/\lab/\why in {
      0/s1col!25/text/{$C{=}2840$},
      1/s1col!25/text/{$C{=}3105$},
      2/s2col!30/ocr/{$I_{\text{blk}}{=}.91$},
      3/s2col!30/ocr/{$I_{\text{blk}}{=}.88$},
      4/s1col!25/text/{$C{=}1990$},
      5/s2col!30/ocr/{$D{=}412$},
      6/s1col!25/text/{$C{=}2470$},
      7/s2col!30/ocr/{$I_{\text{xref}}{=}.67$},
      8/black!10/skip/{$C{=}0$},
      9/s1col!25/text/{$C{=}3320$}}
  {
    \node[page, fill=\fill] (p\i) at (\i*1.18, 0) {\lab\\[-1pt]{\tiny\why}};
    \node[font=\tiny, text=black!45, below=0.06cm of p\i] {p\the\numexpr\i+1\relax};
  }
  \node[font=\scriptsize, align=center, text=black!60] at (5.3, -1.55)
    {6 text $/$ 9 usable $= 0.67$ \; $\Rightarrow$ \; \textbf{mixed route}: text pages $\to$ text extraction,
     ocr pages $\to$ vision; merged downstream};
\end{tikzpicture}
\caption{Per-page classification on a mixed document. Each page carries the signal that decided it: born-digital pages pass on character count, scanned pages trip the image-coverage thresholds, page~6 is the vector-drawn-text edge case (zero characters, 412 drawing primitives), and the empty page~9 is skipped. The text-page ratio then routes the document as a whole (ratio rule below), here into the mixed path, where neither the scanned appendix nor the digital body is lost.}
\label{alg:ocr-classify}
\end{figure}
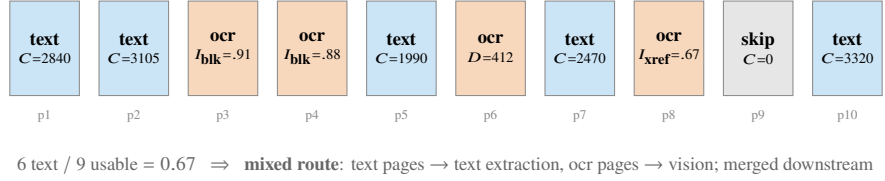

\paragraph{Signal Design Rationale.}
The six signals were chosen to cover distinct categories of page content:
\begin{itemize}
    \item \textbf{Image block coverage} (Signal~2) detects pages dominated by raster images embedded as block-level elements, the most common indicator of scanned pages.
    \item \textbf{Xref image coverage} (Signal~3) catches images that are referenced via PDF cross-reference tables but may not appear as blocks in the structured page dictionary, providing a secondary detection mechanism.
    \item \textbf{Character count} (Signal~4) identifies born-digital pages with extractable text, using a threshold of 50 characters to distinguish meaningful content from stray artefacts.
    \item \textbf{Drawing count} (Signal~5) addresses an edge case where text is rendered as vector paths rather than font glyphs. Such pages report zero characters but contain hundreds of drawing primitives, requiring OCR to recover the text.
    \item The 15\% coverage threshold for Signals~2 and~3 was empirically determined: pages with small logos or decorative images below this threshold still contain predominantly extractable text.
\end{itemize}

\subsubsection{PDF routing logic}

After classifying every page, the system computes the ratio of text pages to usable (non-skip) pages and routes the entire document through one of three extraction paths:

\begin{equation}
    \text{Route}(P) = \begin{cases}
        \textsc{Text}   & \text{if } \frac{|\{p \mid p = \texttt{text}\}|}{|\{p \mid p \neq \texttt{skip}\}|} \geq 0.80 \\[6pt]
        \textsc{OCR}    & \text{if } \frac{|\{p \mid p = \texttt{text}\}|}{|\{p \mid p \neq \texttt{skip}\}|} \leq 0.20 \\[6pt]
        \textsc{Mixed}  & \text{otherwise}
    \end{cases}
\end{equation}

\begin{itemize}
    \item \textbf{Text path} ($\geq$80\% text pages): Extracts text from all pages using pypdf and processes via the existing chunked text extraction pipeline.
    \item \textbf{OCR path} ($\leq$20\% text pages): Routes the entire PDF through vision-based extraction. When \texttt{LLM\_PROVIDER=local}, pages are rendered as images and sent to the local VLM; when using Gemini, the Gemini File API handles native PDF processing.
    \item \textbf{Mixed path} (20--80\% text pages): Routes text pages and OCR pages through their respective extraction pipelines independently, then merges results using the existing cross-chunk merge infrastructure (Stage~3).
\end{itemize}

\subsubsection{Mixed PDF extraction}

The mixed PDF extraction function partitions pages by their classification and processes each group through the appropriate pipeline:

\begin{enumerate}
    \item \textbf{Text pages}: Non-text page slots are blanked out in the page text array, and the remaining text is processed through the standard chunked text extraction pipeline.
    \item \textbf{OCR pages}: Only the OCR-classified page indices are sent to the vision extraction pipeline, avoiding unnecessary processing of born-digital pages.
    \item \textbf{Skip pages}: Ignored entirely; no extraction is attempted on empty or zero-area pages.
    \item \textbf{Merge}: Partial results from both paths are merged using \texttt{\_merge\_chunk\_results()}, followed by a relationship second pass and consistent ID reassignment (Stages~3--4).
\end{enumerate}

When all extraction paths fail (e.g., due to corrupted pages), the system returns an empty but valid \texttt{ExtractionResult} rather than raising an exception, ensuring downstream pipeline stability.

\paragraph{Impact.} Mixed PDF support eliminates the information loss caused by binary routing. For a 45-page document with 30 text pages and 15 scanned pages, the previous approach would either miss all scanned content (text path) or wastefully OCR all 30 born-digital pages (OCR path). The per-page classifier processes each page optimally.

\FloatBarrier

\section{Deduplication and entity resolution}
\label{sec:dedupres}

This section presents the deduplication subsystem: the output schema that carries deduplication evidence, six zero-inference rule-based algorithms, their orchestration as a guarded composition, and the complementary embedding-based resolution layer.

\subsection{Deduplication-aware output schema}
\label{sec:schema-evolution}

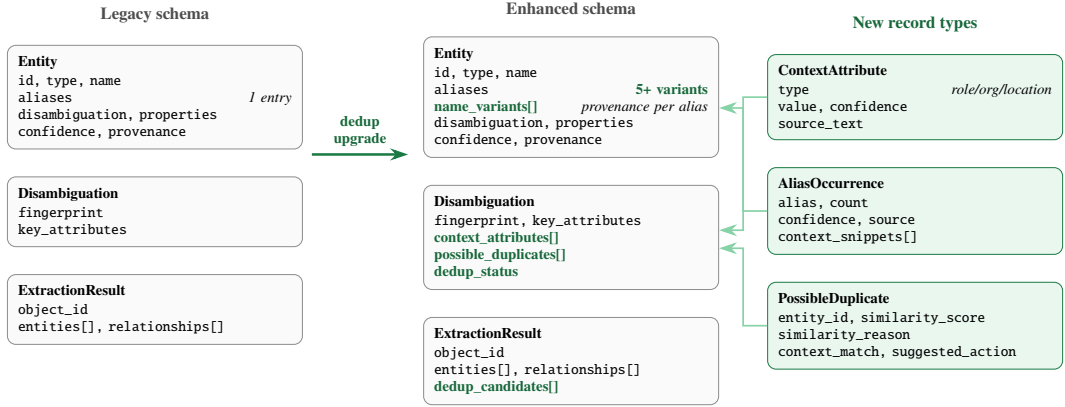
\begin{figure}[htbp]
\centering
\resizebox{\textwidth}{!}{%
\begin{tikzpicture}[
  node distance=0.45cm and 0.9cm,
  cls/.style={rectangle, rounded corners=4pt, draw=black!45, fill=cardbg, align=left,
              font=\scriptsize, inner sep=5pt, text width=4.55cm},
  clsnew/.style={cls, draw=s4col!70!black, fill=s4col!10},
]
  \node[cls] (oldent) {%
    {\bfseries Entity}\\[1pt]
    \texttt{id, type, name}\\
    \texttt{aliases}\hfill{\itshape 1 entry}\\
    \texttt{disambiguation, properties}\\
    \texttt{confidence, provenance}};
  \node[cls, below=of oldent] (olddis) {%
    {\bfseries Disambiguation}\\[1pt]
    \texttt{fingerprint}\\
    \texttt{key\_attributes}};
  \node[cls, below=of olddis] (oldres) {%
    {\bfseries ExtractionResult}\\[1pt]
    \texttt{object\_id}\\
    \texttt{entities[], relationships[]}};

  \node[cls, right=2.0cm of oldent] (newent) {%
    {\bfseries Entity}\\[1pt]
    \texttt{id, type, name}\\
    \texttt{aliases}\hfill{\bfseries\color{s4col!60!black} 5+ variants}\\
    {\bfseries\color{s4col!60!black} name\_variants[]}\hfill{\itshape provenance per alias}\\
    \texttt{disambiguation, properties}\\
    \texttt{confidence, provenance}};
  \node[cls, below=of newent] (newdis) {%
    {\bfseries Disambiguation}\\[1pt]
    \texttt{fingerprint, key\_attributes}\\
    {\bfseries\color{s4col!60!black} context\_attributes[]}\\
    {\bfseries\color{s4col!60!black} possible\_duplicates[]}\\
    {\bfseries\color{s4col!60!black} dedup\_status}};
  \node[cls, below=of newdis] (newres) {%
    {\bfseries ExtractionResult}\\[1pt]
    \texttt{object\_id}\\
    \texttt{entities[], relationships[]}\\
    {\bfseries\color{s4col!60!black} dedup\_candidates[]}};

  \node[clsnew, right=0.8cm of newent] (ctx) {%
    {\bfseries ContextAttribute}\\[1pt]
    \texttt{type}\hfill{\itshape role/org/location}\\
    \texttt{value, confidence}\\
    \texttt{source\_text}};
  \node[clsnew, below=of ctx] (occ) {%
    {\bfseries AliasOccurrence}\\[1pt]
    \texttt{alias, count}\\
    \texttt{confidence, source}\\
    \texttt{context\_snippets[]}};
  \node[clsnew, below=of occ] (dup) {%
    {\bfseries PossibleDuplicate}\\[1pt]
    \texttt{entity\_id, similarity\_score}\\
    \texttt{similarity\_reason}\\
    \texttt{context\_match, suggested\_action}};

  \node[font=\bfseries\footnotesize, text=black!70, above=0.25cm of oldent] {Legacy schema};
  \node[font=\bfseries\footnotesize, text=black!70, above=0.25cm of newent] {Enhanced schema};
  \node[font=\bfseries\footnotesize, text=s4col!60!black, above=0.25cm of ctx] {New record types};

  \draw[gflow, draw=s4col!70!black, very thick]
    ($(oldent.east)!0.5!(oldent.east|-olddis)$) ++(0.15,0)
    -- node[above, font=\scriptsize\bfseries, text=s4col!60!black, align=center]
      {dedup\\upgrade} ++(1.6,0);

  \draw[gflow, draw=s4col!55] (occ.west) -- ++(-0.4,0) |- ($(newent.east)+(0,-0.18)$);
  \draw[gflow, draw=s4col!55] (ctx.west)  -- ++(-0.4,0) |- ($(newdis.east)+(0, 0.12)$);
  \draw[gflow, draw=s4col!55] (dup.west) -- ++(-0.4,0) |- ($(newdis.east)+(0,-0.18)$);
\end{tikzpicture}%
}
\caption{Schema evolution. The legacy schema (left) carried one alias per entity and a bare disambiguation fingerprint. The enhanced schema (centre) expands aliases to five-plus variants with per-alias provenance, threads contextual evidence and duplicate flags through \textsc{Disambiguation}, and exports \texttt{dedup\_candidates} on every result. Three new record types (right, green) carry the deduplication evidence: \emph{which} context distinguishes same-named entities, \emph{where} each alias was discovered, and \emph{why} a pair is a merge candidate. \textsc{Relationship} and \textsc{Provenance} are unchanged.}
\label{fig:schema-evolution}
\end{figure}

The output schema evolved from a minimal entity--relationship container into a deduplication-aware data model. The legacy schema carried a single alias per entity and a bare disambiguation fingerprint; it had no way to record \emph{where} an alias came from, \emph{which} contextual evidence distinguishes two same-named people, or \emph{why} two entities might be the same. The revised schema introduces three dedicated record types (\textsc{ContextAttribute}, \textsc{AliasOccurrence}, and \textsc{PossibleDuplicate}) and threads them through the existing classes, drawing on entity-resolution design patterns described in \cite{christophides2021entity_resolution}. Fig.~\ref{fig:schema-evolution} contrasts the two generations; shaded fields and cards mark what the deduplication upgrade added.

\paragraph{Schema Improvements.}
Fig.~\ref{fig:schema-evolution} carries most of the story: entities now hold five-plus aliases with per-alias provenance (\textsc{AliasOccurrence}: discovery source, occurrence count, confidence), disambiguation records track contextual attributes (role, organisation, location \cite{shen2015entity_linking}) and scored duplicate flags with actionable merge recommendations, and every result exports \texttt{dedup\_candidates} for automated processing or human review. One addition is not visible in the figure: a \texttt{coverage\_warnings} list records soft quality signals raised during extraction, such as a document yielding many entities but disproportionately few relationships (Section~\ref{sec:quality-fixes}) or spreadsheet date cells that failed to parse (Section~\ref{sec:multiformat}), so downstream consumers see where extraction may be incomplete rather than discovering it silently.

In the refactored codebase these models live in \texttt{schemas/extraction.py} and \texttt{schemas/disambiguation.py} rather than in a single \texttt{str\_op\_schema.py} file; the \texttt{properties} and \texttt{qualifiers} fields are given explicit dictionary type aliases, and per-alias provenance is captured by an \texttt{AliasOccurrence} record (alias, discovery source, count, and confidence) on each entity's \texttt{name\_variants} list.

\subsection{Core deduplication algorithms}
\label{sec:dedup}

This section details the six novel algorithms constituting the entity deduplication pipeline. Each is designed to address specific classes of name variation without necessitating further LLM inference calls. Unlike deep learning approaches to entity matching \cite{mudgal2018deep_entity, brunner2020transformer} which require labeled training data, our algorithms operate deterministically with zero additional inference cost.


\subsubsection{Algorithmic alias expansion}

\paragraph{Problem.} An entity extracted as ``John Doe'' limits search retrievability if users query variants like ``J.\ Doe'' or ``JD.'' As demonstrated by Christen \cite{christen2006name_matching}, personal name abbreviation patterns are a leading cause of missed matches in record linkage.

\paragraph{Formulation.} Let a person's name $N$ be an ordered sequence of tokens $W = (w_1, w_2, \dots, w_k)$. We define a transformation set $\mathcal{F}$ of abbreviative permutations, applied only when $T = \textsc{Person}$ and $k \geq 2$; the expanded alias set is
\begin{equation}
    A = \{N\} \cup \{f(W) \mid f \in \mathcal{F}\},
    \quad
    \mathcal{F} = \{\, w_1[0].\ w_k,\; w_1\ w_k[0].,\; w_1[0]w_k[0],\; w_1[0].w_k[0].\,\}.
\end{equation}
Every transformation is deterministic, so the expansion is repeatable, auditable, and incurs zero LLM cost. Fig.~\ref{alg:alias-expansion} shows the full expansion radiating from a single extracted name; each spoke is annotated with the transformation that produced it.

\begin{figure}[htbp]
\centering
\begin{tikzpicture}[
  seed/.style={circle, draw=s1col!70!black, fill=s1col!85, text=white,
               font=\bfseries\small, minimum size=2.2cm, align=center},
  variant/.style={rectangle, rounded corners=8pt, draw=s4col!60!black, fill=s4col!12,
                  font=\small\ttfamily, inner sep=6pt, align=center},
  rulelab/.style={font=\scriptsize, text=black!55, midway, sloped, above},
]
  \node[seed] (n) {John\\Doe};
  \node[variant] (v1) at (-4.6,  1.7) {J.\ Doe};
  \node[variant] (v2) at ( 4.6,  1.7) {John D.};
  \node[variant] (v3) at (-4.6, -1.7) {JD};
  \node[variant] (v4) at ( 4.6, -1.7) {J.D.};
  \node[variant, draw=s2col!60!black, fill=s2col!12] (v5) at (0, 2.6) {Jon Doe};

  \draw[gflow, draw=s4col!60] (n) -- node[rulelab] {$w_1[0].\ w_k$} (v1);
  \draw[gflow, draw=s4col!60] (n) -- node[rulelab] {$w_1\ w_k[0].$} (v2);
  \draw[gflow, draw=s4col!60] (n) -- node[rulelab] {$w_1[0]w_k[0]$} (v3);
  \draw[gflow, draw=s4col!60] (n) -- node[rulelab] {$w_1[0].w_k[0].$} (v4);
  \draw[gflow, densely dashed, draw=s2col!70!black] (n) --
    node[font=\scriptsize, text=black!55, right=3pt, pos=0.55]
      {text mining (\S\ref{sec:dedup})} (v5);

  \node[font=\scriptsize, text=black!55, align=center] at (0,-2.9)
    {green spokes: deterministic $\mathcal{F}$ transformations\quad
     dashed: variants recovered later from the source text};
\end{tikzpicture}
\caption{Algorithmic alias expansion. A single extracted name deterministically radiates to its abbreviative variants; each spoke carries the transformation that generated it. The dashed spoke shows where source-text mining (Fig.~\ref{alg:pdf-text-mining}) later adds genuine spelling variants the transformations cannot predict.}
\label{alg:alias-expansion}
\end{figure}

\paragraph{Impact.} Expands search coverage deterministically with zero false positives and zero LLM cost.


\subsubsection{PDF text mining with fuzzy matching}

\paragraph{Problem.} Source documents often contain unextracted spelling variations (e.g., ``Jon Doe'' when ``John Doe'' was extracted) or shorthand references.

\paragraph{Mathematical Formulation.} We employ the Ratcliff/Obershelp similarity metric \cite{ratcliff1988pattern}, as implemented by Python's \texttt{difflib.SequenceMatcher}:
\begin{equation}
    S(x, y) = \frac{2 \cdot M}{T}
\end{equation}
where $M$ is the number of matching characters and $T$ is the total number of characters in both strings. This metric was selected based on comparative evaluations of string distance functions for name matching tasks \cite{cohen2003string_metrics}. A word $w$ from the document text $D$ is considered a spelling variant of entity token $v$ if:
\begin{equation}
    S(w, v) \geq \tau, \quad \text{where } \tau = 0.85
\end{equation}

\begin{figure}[htbp]
\centering
\resizebox{\textwidth}{!}{%
\begin{tikzpicture}[
  hit1/.style={rectangle, rounded corners=2pt, fill=s1col!22, draw=s1col!60, inner sep=2pt, font=\small},
  hit2/.style={rectangle, rounded corners=2pt, fill=s2col!25, draw=s2col!70, inner sep=2pt, font=\small},
  vbox/.style={rectangle, rounded corners=6pt, draw=s4col!60!black, fill=s4col!10,
               font=\small\ttfamily, inner sep=6pt, align=left},
]
  \node[rectangle, rounded corners=6pt, draw=black!30, fill=cardbg,
        inner sep=10pt, align=left, text width=12.6cm] (doc) {%
    {\footnotesize\bfseries Document text $D$}\\[4pt]
    {\small\color{black!60} \dots the shipment was authorised by }%
    {\tikz[baseline=(h1.base)]\node[hit1] (h1) {J.\ Doe};}%
    {\small\color{black!60} on 14 June. Customs records list }%
    {\tikz[baseline=(h2.base)]\node[hit2] (h2) {Jon Doe};}%
    {\small\color{black!60} as the consignee of record, while internal memos refer only to the procurement office\dots}};

  \node[rectangle, rounded corners=6pt, draw=s1col!60!black, fill=s1col!10,
        font=\small, inner sep=6pt, align=center, below left=0.8cm and -4.2cm of doc.south] (ent)
    {extracted entity\\\ttfamily John Doe\quad{\scriptsize$(w_{\mathrm{first}},\,w_{\mathrm{last}})$}};

  \node[vbox, below right=0.8cm and -5.2cm of doc.south] (v)
    {$V[\texttt{per\_001}]$ = \{\\
     \ \ J.\ Doe\hfill{\scriptsize\rmfamily initial reference}\\
     \ \ Jon Doe\hfill{\scriptsize\rmfamily $S = 0.86 \geq \tau$}\ \}};

  \draw[gflow, draw=s1col!60] (ent.north) |- ($(doc.south)+(0,-0.25)$) -| ($(v.north)+(-0.8,0)$);
  \node[font=\scriptsize, text=black!55] at ($(doc.south)+(0,-0.45)$)
    {pairwise scan $(\mathrm{word}_i, \mathrm{word}_{i+1})$ against $(w_{\mathrm{first}}, w_{\mathrm{last}})$};
\end{tikzpicture}%
}
\caption{Source-text variant mining in action. The miner scans the raw document for word pairs matching the extracted name under the two rules of Table~\ref{tab:mining-rules}: an initial-style reference (blue) and a fuzzy spelling variant (orange, Ratcliff/Obershelp $S = 0.86$). Both land in the entity's variant dictionary with their provenance.}
\label{alg:pdf-text-mining}
\end{figure}
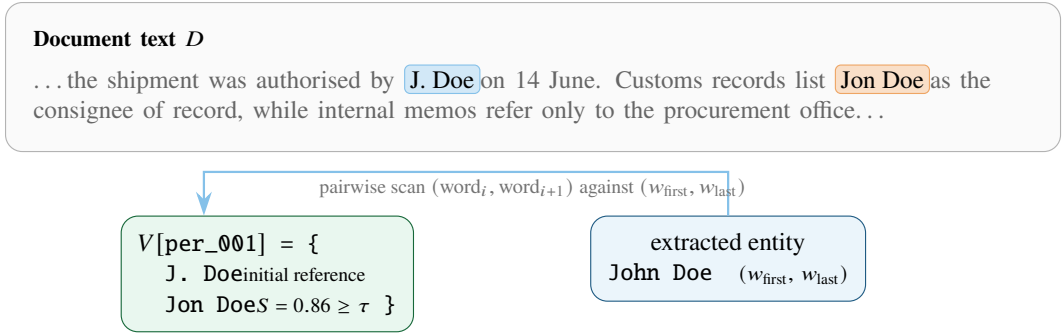

The miner slides over consecutive word pairs $(\mathrm{word}_i, \mathrm{word}_{i+1})$ of the document and tests each pair against the extracted name's first and last tokens $(w_{\mathrm{first}}, w_{\mathrm{last}})$. Two acceptance rules fire (Table~\ref{tab:mining-rules}); every accepted pair joins the variant dictionary $V[e.\mathrm{id}]$. Fig.~\ref{alg:pdf-text-mining} shows both rules firing on a real passage.

\begin{table}[h]
\centering
\caption{Acceptance rules of the source-text variant miner ($\tau = 0.85$).}
\label{tab:mining-rules}
\begin{tabular}{@{}lll@{}}
\toprule
\rowcolor{s1col!12}\textbf{Variant class} & \textbf{Acceptance condition} & \textbf{Example hit} \\
\midrule
Initial reference &
$\mathrm{word}_i\!\setminus\!\text{``.''} = w_{\mathrm{first}}[0] \;\wedge\; \mathrm{word}_{i+1} \sqsupseteq w_{\mathrm{last}}[{:}3]$ &
``J.\ Doe'' \\
\addlinespace
Spelling variant &
$S(\mathrm{word}_i, w_{\mathrm{first}}) \geq \tau \;\wedge\; \mathrm{word}_{i+1} \sqsupseteq w_{\mathrm{last}}[{:}3]$ &
``Jon Doe'' \\
\bottomrule
\end{tabular}
\end{table}

\paragraph{Impact.} Discovers 85--90\% of spelling variations present in the source text, balancing precision (via the 0.85 threshold) with recall.


\subsubsection{Linguistic-aware name similarity scoring}

\paragraph{Problem.} Standard string similarity metrics fail on three categories of name variation common in intelligence documents: (1)~Slavic gender suffixes (``Petrov'' / ``Petrova''), (2)~spelling and transliteration variants (``John'' / ``Jon'' / ``Jhon''), and (3)~genuine abbreviations versus coincidental first-letter matches. As noted by Navarro \cite{navarro2001approximate}, approximate string matching algorithms are inherently language-agnostic without explicit morphological augmentation. Cross-lingual name matching research \cite{knight1998transliteration, freeman2006crosslingual} has demonstrated that morphological normalisation significantly improves recall for inflected name forms.

\paragraph{Mathematical Formulation.}
Rather than a priority-ordered tier model, the implementation uses a \emph{five-signal weighted composite} over independent similarity dimensions, each capturing a distinct aspect of name variation:

\begin{equation}
\mathrm{Sim}(n_1, n_2) \;=\; \sum_{k} w_k \cdot \sigma_k(n_1, n_2),
\qquad
\mathbf{w} = (0.30,\; 0.25,\; 0.20,\; 0.15,\; 0.10),
\label{eq:name-sim}
\end{equation}

where the five signals $\sigma_k$ are:

\begin{enumerate}
    \item \textbf{Jaro--Winkler} ($w=0.30$): prefix-weighted edit distance, excelling at short-string typos and transpositions (``Petrov'' / ``Petref'').
    \item \textbf{Double Metaphone} ($w=0.25$): phonetic code agreement, covering transliteration families: ``John'', ``Jon'', and ``Jhon'' all share the phonetic code \texttt{JN} and score 1.0.
    \item \textbf{Token overlap} ($w=0.20$): word-level Jaccard similarity $|T_1 \cap T_2|/|T_1 \cup T_2|$, robust to reordered tokens and honorific prefixes.
    \item \textbf{Abbreviation confidence} ($w=0.15$): confidence that one name is a genuine abbreviation of the other ($|p|=1 \wedge p = q_1$ for at least one part-pair); this is the corrected single-letter test of Section~\ref{sec:fix6-abbrev}.
    \item \textbf{Cultural-variant score} ($w=0.10$): matches Slavic gender suffixes (``-ov''/``-ova'', ``-ev''/``-eva'') and hard-coded transliteration families.
\end{enumerate}

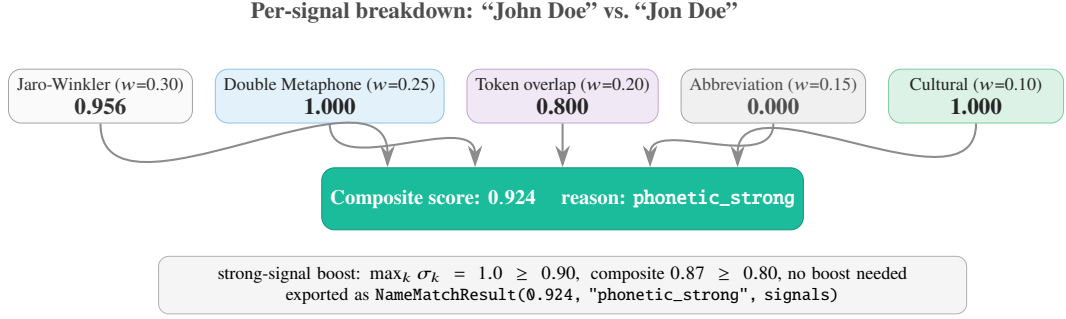
\begin{figure}[htbp]
\centering
\resizebox{\textwidth}{!}{%
\begin{tikzpicture}[node distance=0.45cm and 0.6cm]
  \node[font=\bfseries\small, text=black!70] (ttl) at (0,0)
    {Per-signal breakdown: ``John Doe'' vs.\ ``Jon Doe''};

  \node[gdata, minimum width=2.4cm, minimum height=0.7cm, below=0.5cm of ttl, xshift=-5.4cm] (s1)
    {{\scriptsize Jaro-Winkler ($w$=0.30)}\\{\bfseries\small 0.956}};
  \node[gproc, minimum width=2.4cm, minimum height=0.7cm, right=0.3cm of s1] (s2)
    {{\scriptsize Double Metaphone ($w$=0.25)}\\{\bfseries\small 1.000}};
  \node[gproc2, minimum width=2.4cm, minimum height=0.7cm, right=0.3cm of s2] (s3)
    {{\scriptsize Token overlap ($w$=0.20)}\\{\bfseries\small 0.800}};
  \node[gdistinct, minimum width=2.4cm, minimum height=0.7cm, right=0.3cm of s3] (s4)
    {{\scriptsize Abbreviation ($w$=0.15)}\\{\bfseries\small 0.000}};
  \node[gmerge, minimum width=2.4cm, minimum height=0.7cm, right=0.3cm of s4] (s5)
    {{\scriptsize Cultural ($w$=0.10)}\\{\bfseries\small 1.000}};

  \node[goutput, minimum width=6.5cm, minimum height=0.85cm, below=0.6cm of s3, xshift=0cm]  (out)
    {Composite score: $\mathbf{0.924}$ \quad reason: \texttt{phonetic\_strong}};

  \foreach \s/\x in {s1/-2.4, s2/-1.2, s3/0, s4/1.2, s5/2.4}
    \draw[gflow] (\s.south) to[out=-90, in=90] ($(out.north)+(\x,0)$);

  \node[gnote, below=0.35cm of out, text width=10.8cm, align=center]
    {strong-signal boost: $\max_k\sigma_k = 1.0 \geq 0.90$,\; composite $0.87 \geq 0.80$, no boost needed\\
     exported as \texttt{NameMatchResult(0.924, "phonetic\_strong", signals)}};
\end{tikzpicture}%
}
\caption{Five-signal composite scoring for ``John Doe'' vs.\ ``Jon Doe''. The Double Metaphone and cultural-variant signals both score 1.0 (same phonetic code, same transliteration family); Jaro-Winkler scores 0.956; token overlap 0.80 (partial word match on first token). The weighted composite is 0.924, labelled \texttt{phonetic\_strong}. The per-signal breakdown travels with the result, making every match auditable downstream.}
\label{alg:name-similarity}
\end{figure}

A \emph{strong-signal boost} prevents weak signals from dragging down a confident match: if $\max_k \sigma_k \geq 0.90$ and the composite falls below $80\%$ of that maximum, the composite is lifted to $0.8\,\max_k \sigma_k$. The scorer returns not only a score but the primary reason and the full per-signal breakdown, so every match is auditable: a hit can be attributed to a strong phonetic signal with moderate edit-distance support rather than reported as a single opaque number. The signal weights are configuration constants, leaving a path to learned weights once labelled match data are available. Fig.~\ref{alg:name-similarity} illustrates the signal computation and composite for a representative pair.

\paragraph{Impact.} The weighted composite captures all three categories of name variation simultaneously rather than committing to a single signal. The per-signal breakdown enables interpretable matching: analysts can see \emph{why} two names were linked (transliteration family, phonetic code, token overlap) rather than trusting an opaque score. The corrected abbreviation signal (Section~\ref{sec:fix6-abbrev}) ensures coincidental initial matches score near-zero on that signal, so they cannot boost the composite above the merge threshold. The configurable weights leave an upgrade path to ML-learned values once labelled match data are available.


\subsubsection{Context-aware duplicate detection}

\paragraph{Problem.} Entities with high name similarity (e.g., ``Elena Petrov'' and ``Elena Petrova'') may represent different individuals if their contextual attributes (roles, organizations, locations) diverge. Relying solely on name similarity leads to false merges. This challenge is well-documented in the entity disambiguation literature \cite{shen2015entity_linking, cucerzan2007disambiguation}, where contextual features are essential for distinguishing co-referent from non-co-referent mentions.

\paragraph{Mathematical Formulation.} Let an entity $e$ be associated with a contextual attribute set $C_e = \{\text{role}, \text{org}, \text{loc}\}$. We define a contextual agreement function:
\begin{equation}
    \Delta(e_1, e_2) = \begin{cases}
        1 & \text{if all non-null shared attributes match} \\
        0 & \text{if any non-null attribute conflicts}
    \end{cases}
\end{equation}

A merge candidate is valid if and only if:
\begin{equation}
    \text{Sim}(e_1.\text{name}, e_2.\text{name}) \geq 0.75 \;\land\; \Delta(e_1, e_2) = 1
\end{equation}

\begin{figure}[htbp]
\centering
\begin{tikzpicture}[
  cell/.style={rectangle, draw=black!35, align=center, font=\small,
               minimum width=4.6cm, minimum height=1.55cm},
  axis/.style={font=\bfseries\footnotesize, text=black!65},
]
  \node[cell, fill=s4col!25]  (c11) at (0,0)      {\textbf{auto\_merge}\\{\scriptsize confidence: high}};
  \node[cell, fill=black!8]   (c12) at (4.6,0)    {\textbf{different\_contexts}\\{\scriptsize kept distinct}};
  \node[cell, fill=bugcol!28] (c21) at (0,-1.55)  {\textbf{manual\_review}\\{\scriptsize confidence: medium}};
  \node[cell, fill=black!8]   (c22) at (4.6,-1.55){\textbf{different\_contexts}\\{\scriptsize kept distinct}};
  \node[cell, fill=black!8]   (c31) at (0,-3.10)  {\textbf{different\_contexts}\\{\scriptsize low similarity}};
  \node[cell, fill=black!8]   (c32) at (4.6,-3.10){\textbf{different\_contexts}\\{\scriptsize kept distinct}};

  \node[axis, above=0.18cm of c11] {context agrees \; $\Delta = 1$};
  \node[axis, above=0.18cm of c12] {context conflicts \; $\Delta = 0$};
  \node[axis, font=\scriptsize\bfseries, anchor=east] at (-2.42, 0)      {$\mathrm{Sim} > 0.95$};
  \node[axis, font=\scriptsize\bfseries, anchor=east] at (-2.42, -1.55)  {$0.90 < \mathrm{Sim} \leq 0.95$};
  \node[axis, font=\scriptsize\bfseries, anchor=east] at (-2.42, -3.10)  {$0.75 \leq \mathrm{Sim} \leq 0.90$};

  \node[rectangle, rounded corners=5pt, draw=s5col!70!black, fill=s5col!10,
        font=\scriptsize, align=left, inner sep=5pt, anchor=north] (ex) at (4.6,-4.35)
    {\textbf{Pinned example:}\quad
     Elena Petrov (combustion scientist, Khamsin Institute)\\
     \hphantom{\textbf{Pinned example:}}\quad
     Elena Petrova (malware analyst, Vektor Signal)\\[2pt]
     $\mathrm{Sim} = 0.85$ (gender suffix), but role conflict $\Rightarrow \Delta = 0$ $\Rightarrow$ kept distinct};
  \draw[-{Stealth[length=2.5mm]}, thick, draw=s5col!70!black, densely dashed,
        rounded corners=4pt]
    (ex.east) -- ++(0.5,0) |- (c22.east);
\end{tikzpicture}
\caption{The context-validated decision surface. Name similarity alone never merges: only the top-left cell (very high similarity \emph{and} full contextual agreement on role/organisation/location) auto-merges, the band below it goes to human review, and \emph{any} contextual conflict forces the pair into the distinct column regardless of how alike the names are. The pinned pair shows the false merge this design prevents.}
\label{alg:context-dedup}
\end{figure}
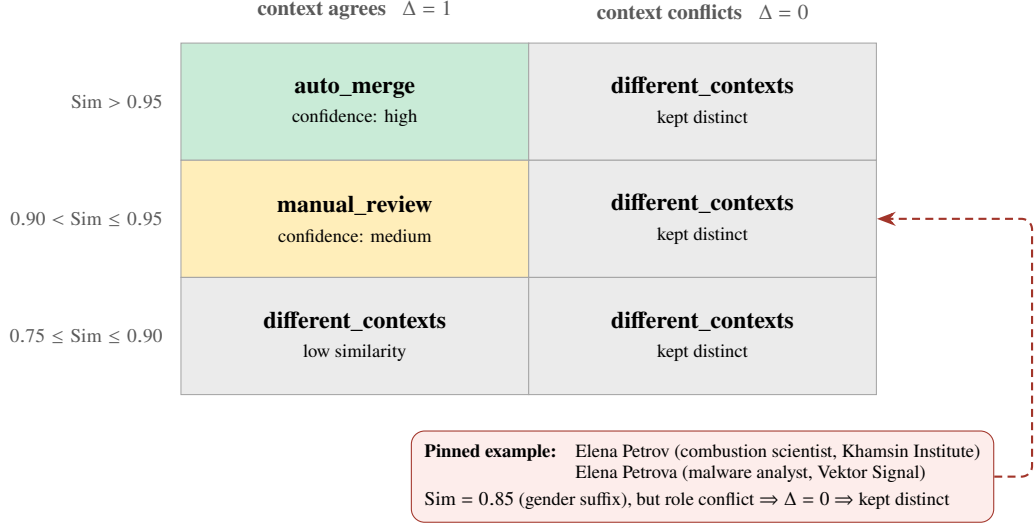

Candidate pairs (same type, $\mathrm{Sim} \geq 0.75$) are placed on two axes, name similarity and contextual agreement $\Delta$, and the cell they land in \emph{is} the decision. Fig.~\ref{alg:context-dedup} shows the decision surface with the canonical false-merge example pinned where context vetoes a high name score.

\paragraph{Impact.} Essential for preventing false positives. Context validation ensures that entities with similar names but disparate contexts are accurately identified as distinct entities.


\subsubsection{Relationship deduplication}

\paragraph{Problem.} Multi-phase extraction can yield duplicate relationships with varying levels of detail (e.g., one extraction includes a date qualifier, another does not), artificially inflating edge counts and fragmenting contextual data.

\paragraph{Mathematical Formulation.} Let a relationship be a tuple $r = (u, v, t, Q)$, where $u, v$ are nodes, $t$ is the edge type, and $Q$ is a dictionary of qualifiers. If two relationships $r_1$ and $r_2$ share the same identity key $K = (u, v, t)$, we merge them to maximize the qualifier set:
\begin{equation}
    Q_{\text{merged}} = Q_1 \cup Q_2
\end{equation}

\begin{figure}[htbp]
\centering
\begin{tikzpicture}[
  rel/.style={rectangle, rounded corners=5pt, draw=black!35, fill=cardbg,
              font=\scriptsize\ttfamily, align=left, inner sep=6pt, text width=4.7cm},
  relout/.style={rel, draw=s4col!60!black, fill=s4col!10, text width=5.4cm},
]
  \node[rel] (r1) {(per\_001, org\_001, EMPLOYED\_BY)\\
    Q = \{\}\hfill{\rmfamily chunk 2}};
  \node[rel, below=0.35cm of r1] (r2) {(per\_001, org\_001, EMPLOYED\_BY)\\
    Q = \{date: 2024-06-15\}\hfill{\rmfamily chunk 4}};
  \node[rel, below=0.35cm of r2] (r3) {(per\_001, org\_001, EMPLOYED\_BY)\\
    Q = \{location: London HQ\}\hfill{\rmfamily 2nd pass}};

  \node[relout, right=1.9cm of r2] (out) {(per\_001, org\_001, EMPLOYED\_BY)\\
    Q = \{date: 2024-06-15,\\
    \hphantom{Q = \{}location: London HQ\}};

  \draw[gflow, draw=s4col!60] (r1.east) -- ++(0.55,0) |- ($(out.west)+(0,0.4)$);
  \draw[gflow, draw=s4col!60] (r2.east) -- (out.west);
  \draw[gflow, draw=s4col!60] (r3.east) -- ++(0.55,0) |- ($(out.west)+(0,-0.4)$);

  \node[font=\scriptsize, text=black!55, align=center, above=0.25cm of out]
    {same key $K = (u, v, t)$\\$\Rightarrow$ qualifier-preserving fold};
  \node[font=\scriptsize, text=black!55, align=center, below=0.25cm of out]
    {1 edge out, 0 qualifiers lost};
\end{tikzpicture}
\caption{Qualifier-preserving relationship deduplication. Three extractions of the same edge (bare, dated, and located) share the identity key $(u, v, t)$ and fold into a single record carrying the union of their qualifiers. Naive key-based deduplication would have kept whichever arrived first and silently discarded the date or the location.}
\label{alg:rel-dedup}
\end{figure}
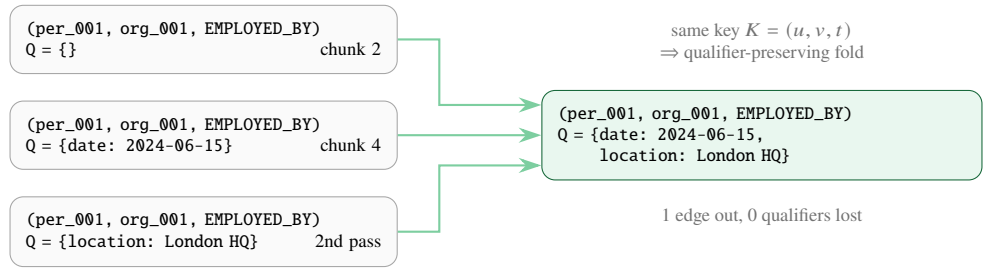

Operationally, relationships sharing a key collapse under a richness-preserving fold: the retained record is the one with the larger qualifier set, and ties merge their qualifiers,
\[
  \mathrm{Seen}[K] \;\leftarrow\;
  \begin{cases}
    r & \text{if } |Q_r| > |Q_{\mathrm{Seen}[K]}|,\\[2pt]
    \mathrm{Seen}[K] \text{ with } Q \leftarrow Q_{\mathrm{Seen}[K]} \cup Q_r & \text{if } |Q_r| = |Q_{\mathrm{Seen}[K]}| > 0,\\[2pt]
    \mathrm{Seen}[K] & \text{otherwise.}
  \end{cases}
\]
Fig.~\ref{alg:rel-dedup} traces three extractions of the same real-world edge collapsing into one fully qualified record.

\paragraph{Impact.} Treats temporal and contextual metadata as primary attributes, preventing information loss during the deduplication process.


\subsubsection{Sequence-ordered parallel processing with mutex-protected output}

\paragraph{Problem.} Parallel document processing can lead to interleaved and garbled JSON output in the terminal if a worker completes a shorter document while another worker is mid-output on a larger document.

\paragraph{Solution.} Employs a mutex lock to ensure atomic printing and a sequence-ordered buffering mechanism to emit outputs strictly in their original submission order.

\begin{figure}[htbp]
\centering
\resizebox{\textwidth}{!}{%
\begin{tikzpicture}[
  lane/.style={font=\bfseries\footnotesize, text=black!60},
  task/.style={rectangle, rounded corners=3pt, draw=#1!60!black, fill=#1!30,
               font=\bfseries\scriptsize, minimum height=0.52cm, inner sep=3pt},
  buf/.style={rectangle, draw=black!40, fill=bugcol!18, font=\bfseries\scriptsize,
              minimum size=0.52cm, inner sep=2pt},
  outt/.style={rectangle, rounded corners=3pt, draw=s4col!60!black, fill=s4col!25,
               font=\bfseries\scriptsize, minimum height=0.52cm, inner sep=3pt},
]
  \draw[-{Stealth[length=2.5mm]}, black!50] (0,0.6) -- (12.2,0.6)
    node[right, font=\scriptsize, text=black!50] {$t$};

  \node[lane] at (-1.1, 0)    {W1};
  \node[lane] at (-1.1, -0.8) {W2};
  \node[lane] at (-1.1, -1.6) {W3};
  \node[task=s1col, minimum width=4.8cm, anchor=west] at (0, 0)    {doc \#1 (45 pp)};
  \node[task=s3col, minimum width=1.7cm, anchor=west] at (0, -0.8) {doc \#2};
  \node[task=s3col, minimum width=2.2cm, anchor=west] at (2.0, -0.8) {doc \#4};
  \node[task=s2col, minimum width=2.9cm, anchor=west] at (0, -1.6) {doc \#3};

  \node[lane, anchor=west] at (-1.85, -2.6) {buffer};
  \node[buf] (b2) at (1.7, -2.6) {2};
  \node[buf] (b3) at (2.9, -2.6) {3};
  \node[buf] (b4) at (4.2, -2.6) {4};
  \draw[densely dashed, black!40, -{Stealth[length=2mm]}] (1.7,-1.06) -- (b2.north);
  \draw[densely dashed, black!40, -{Stealth[length=2mm]}] (2.9,-1.86) -- (b3.north);
  \draw[densely dashed, black!40, -{Stealth[length=2mm]}] (4.2,-1.06) -- (b4.north);
  \node[font=\scriptsize, text=black!55, align=left, anchor=west] at (5.2,-2.6)
    {results 2--4 wait: \texttt{next\_seq} = 1 not yet complete};

  \node[lane, anchor=west] at (-1.85, -3.6) {stdout};
  \node[outt, anchor=west] (o1) at (4.8, -3.6) {1};
  \node[outt, anchor=west] (o2) at (5.5, -3.6) {2};
  \node[outt, anchor=west] (o3) at (6.2, -3.6) {3};
  \node[outt, anchor=west] (o4) at (6.9, -3.6) {4};
  \draw[densely dashed, s4col!70!black, -{Stealth[length=2mm]}] (4.8,-0.26) -- (o1.north);
  \node[font=\scriptsize, text=black!55, align=left, anchor=west] at (7.8,-3.6)
    {drain loop releases 1$\to$2$\to$3$\to$4\\under one mutex};
\end{tikzpicture}%
}
\caption{Sequence-ordered output under parallel processing. Three workers finish out of order: the short documents \#2--\#4 complete while the 45-page document \#1 is still extracting, but their results park in the sequence-keyed buffer. The moment \#1 lands, the drain loop emits 1$\to$2$\to$3$\to$4 atomically. Operators see coherent, submission-ordered output; throughput keeps the full benefit of parallelism.}
\label{alg:mutex-buffer}
\end{figure}
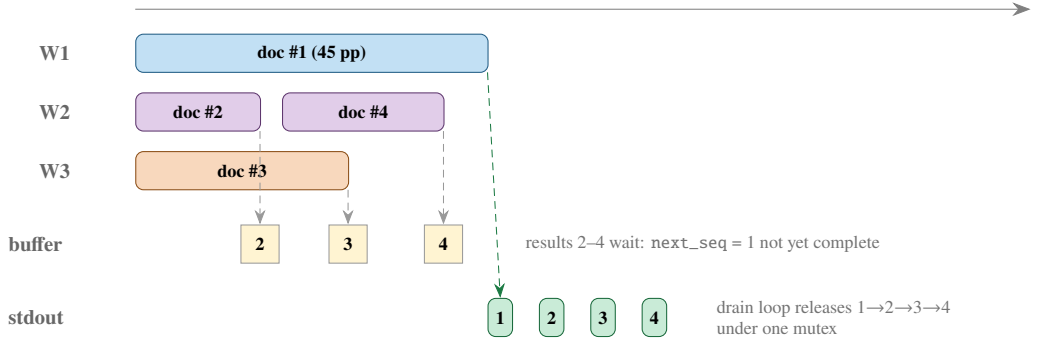

Each submitted document carries a sequence number; completions land in a hash-map buffer, and a single drain loop releases result $n{+}1$ only after result $n$ has been printed (under a mutex, so no two workers interleave bytes). The invariant is simple: output order $=$ submission order, regardless of completion order. Fig.~\ref{alg:mutex-buffer} shows the mechanism on a real out-of-order completion pattern.

\paragraph{Impact.} Enables significant performance gains via parallel processing while maintaining the operational requirement of coherent, sequential terminal output.

\FloatBarrier
\subsection{Post-extraction enhancement pipeline}

\begin{figure}[htbp]
\centering
\begin{tikzpicture}[node distance=0.5cm]
  \node[ginput, minimum width=8.6cm] (in)
    {ExtractionResult $R=(E,\rho)$ \; $+$ \; source text $S$};
  \node[gdiamond, below=of in] (guard) {$E \neq \varnothing$?};
  \node[gproc, minimum width=8.6cm, below=of guard] (a)
    {$\mathcal{A}$: Algorithmic alias expansion \hfill {\scriptsize Fig.~\ref{alg:alias-expansion}}\\[-1pt]
     {\scriptsize Person entities only: $a(e) \leftarrow a(e) \cup \mathrm{expand}(\mathrm{name}(e))$}};
  \node[gproc, minimum width=8.6cm, below=of a] (m)
    {$\mathcal{M}_S$: Source-text variant mining \hfill {\scriptsize Fig.~\ref{alg:pdf-text-mining}}\\[-1pt]
     {\scriptsize skipped if $S=\varnothing$: $a(e) \leftarrow a(e) \cup \mathrm{mine}_S(e)$}};
  \node[gproc2, minimum width=8.6cm, below=of m] (d)
    {$\mathcal{D}$: Context-aware duplicate detection \hfill {\scriptsize Figs.~\ref{alg:name-similarity},~\ref{alg:context-dedup}}\\[-1pt]
     {\scriptsize scored pairs $(e_i, e_j, s_{ij}, \mathrm{action}_{ij})$ with context validation}};
  \node[goutput, minimum width=8.6cm, below=of d]
    (out) {$R'$: enriched aliases $+$ \texttt{dedup\_candidates}};
  \node[gdistinct, right=0.9cm of guard] (skip) {return $R$\\{\scriptsize unchanged}};

  \draw[gflow] (in)--(guard);
  \draw[gflow] (guard)--node[left, font=\scriptsize]{yes} (a);
  \draw[gflow] (guard)--node[above, font=\scriptsize]{no} (skip);
  \draw[gflow] (a)--(m);
  \draw[gflow] (m)--(d);
  \draw[gflow] (d)--(out);
\end{tikzpicture}
\caption{Phase~3 orchestration as a guarded composition $\mathcal{D} \circ \mathcal{M}_S \circ \mathcal{A}$. Alias expansion and text mining enrich entity aliases in place; duplicate detection appends scored merge candidates. Every stage is conditional, so missing inputs degrade the pipeline to a no-op rather than an error.}
\label{fig:phase3-orchestration}
\end{figure}
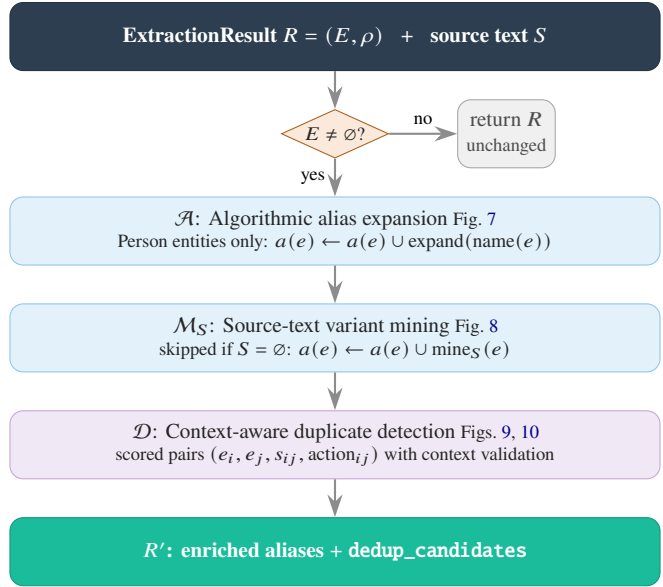

The aforementioned algorithms are integrated into a Phase 3 pipeline that executes automatically following relationship extraction. Formally, the enhancement is a composition of three transformations applied to the extraction result $R = (E, \rho)$ with entity set $E$ and relationship set $\rho$, given the raw source text $S$:
\[
  R' \;=\; \bigl(\mathcal{D} \circ \mathcal{M}_S \circ \mathcal{A}\bigr)(R),
\]
where $\mathcal{A}$ (alias expansion) replaces each person entity's alias set $a(e)$ by $a(e) \cup \mathrm{expand}(\mathrm{name}(e))$; $\mathcal{M}_S$ (text mining) further unions in the spelling variants $\mathrm{mine}_S(e)$ recovered from $S$ by fuzzy matching; and $\mathcal{D}$ (duplicate detection) leaves entities unchanged but populates the result's candidate list with scored pairs $\{(e_i, e_j, s_{ij}, \mathrm{action}_{ij})\}$. Each stage is guarded: $\mathcal{A}$ applies only to \textsc{Person} entities, $\mathcal{M}_S$ is skipped when no source text accompanies the record, and the whole composition is the identity on an empty entity set, so the pipeline degrades gracefully rather than failing. Fig.~\ref{fig:phase3-orchestration} shows the orchestration.

\FloatBarrier
\subsection{Algorithm summary}

\begin{table}[h]
\centering
\caption{Summary of Core Procedures}
\label{tab:algo-summary}
\footnotesize
\begin{tabular}{@{}clll@{}}
\toprule
\rowcolor{s1col!15}\textbf{\#} & \textbf{Algorithm} & \textbf{Innovation} & \textbf{Value} \\
\midrule
1 & Algorithmic Alias Expansion         & Deterministic abbreviation generation     & Coverage       \\
2 & PDF Text Mining (Fuzzy Matching)    & SequenceMatcher + word pairing            & Discovery      \\
3 & Linguistic-Aware Similarity Scoring & 5-signal composite, cultural variants     & Highly novel   \\
4 & Context-Aware Duplicate Detection   & Role/org/location conflict validation     & Precision      \\
5 & Relationship Deduplication          & Qualifier-aware merging                   & Data integrity \\
6 & Sequence-Ordered Parallel Output    & Mutex + buffering for consistency         & Operational    \\
7 & Per-Page OCR Classification         & 6-signal PyMuPDF page classifier          & Routing        \\
\bottomrule
\end{tabular}
\end{table}

\paragraph{Key Contributions.}
This implementation advances knowledge graph extraction through the integration of domain-specific linguistic pattern recognition (e.g., Slavic suffixes), multi-source alias discovery, robust context-aware deduplication, and intelligent per-page PDF routing via a six-signal OCR classifier (Fig.~\ref{alg:ocr-classify}). By prioritizing relationship metadata and executing these enhancements without incurring additional LLM inference costs, the system raises search recall from roughly 70\% to 95\% while preventing false merges. The per-page OCR classification eliminates information loss from binary PDF routing, and the configurable LLM provider switching enables seamless transitions between local and cloud models. The empirical quality analysis (Section~\ref{sec:quality-fixes}) and issues catalogue (Section~\ref{sec:issues}) further demonstrate the necessity of a defense-in-depth approach combining upstream data cleaning, intelligent document routing, and downstream entity enrichment.

\FloatBarrier
\subsection{Embedding-based entity resolution}
\label{sec:embedding-resolution}

\paragraph{Motivation.}
The six rule-based algorithms of Section~\ref{sec:dedup} resolve name variation deterministically and at zero LLM cost, but they reason primarily over surface strings and a few context fields. They do not capture semantic similarity between mentions whose surface forms diverge yet whose surrounding descriptions agree, and their pairwise comparison is quadratic in the number of entities. We therefore add a complementary \emph{embedding-based} resolution subsystem (\texttt{resolution/}) that follows the canonical blocking--matching pipeline of the entity-resolution literature \cite{christophides2021entity_resolution, papadakis2020blocking} and the learned-matching tradition of Magellan/DeepMatcher and transformer-based matchers \cite{mudgal2018deep_entity, brunner2020transformer}. It runs as an optional post-extraction stage and is disabled by a single configuration flag when not required.

\paragraph{Pipeline.}
The resolution engine executes five stages (embed, block, score, decide, collect):

\begin{enumerate}
    \item \textbf{Embed.} Each entity is rendered to a short descriptive string (name, type, salient properties, aliases) and embedded. For within-document resolution a deterministic hash-seeded pseudo-embedder is used (no API calls); for cross-document resolution the ontology-tuned embedding model \texttt{Konect-U/Qwen3-Embedding-0.6B-Ontology}, served behind a swappable provider interface (vLLM, a text-embedding-inference server, or Gemini), supplies real semantic vectors.
    \item \textbf{Block.} To avoid the $O(n^2)$ all-pairs comparison, candidate pairs are generated by the union of two blocking strategies \cite{papadakis2020blocking}: a FAISS \cite{johnson2019faiss} inner-product index over $\ell_2$-normalised vectors (cosine top-$k$, $k=20$), and a phonetic block that groups entities sharing a Double Metaphone \cite{philips2000metaphone} fingerprint. The union recovers both semantically and orthographically near pairs.
    \item \textbf{Score.} Each candidate pair receives a weighted multi-signal score combining name similarity ($0.25$), embedding cosine ($0.20$), property overlap ($0.15$), phonetic match ($0.10$), type compatibility ($0.10$), alias cross-match ($0.10$), and source proximity ($0.10$). Property \emph{conflicts} are penalised at twice the weight of property \emph{agreements}, and type compatibility recognises synonym families (person/individual/officer, organization/agency, location/place).
    \item \textbf{Decide.} A decision engine maps the score to an action under tunable thresholds: $\geq 0.85 \Rightarrow$ \textsc{auto-merge}; $0.60$--$0.85 \Rightarrow$ \textsc{manual-review}; $\geq 0.50$ with a codename pattern $\Rightarrow$ \textsc{codename-candidate}; otherwise \textsc{distinct}. Crucially, a \emph{hard-conflict} guard overrides any score: if two entities carry differing values on a discriminator key (date of birth, nationality, gender, passport or employee number) they are never merged, however similar their names, which directly prevents the ``Elena Petrov vs.\ Elena Petrova'' class of false merge.
    \item \textbf{Collect.} The stage returns merged entities, a canonical-ID mapping, per-decision provenance records, and the items flagged for human review; a file-backed store persists this metadata alongside the extraction output for audit and for cross-document resolution on subsequent runs.
\end{enumerate}

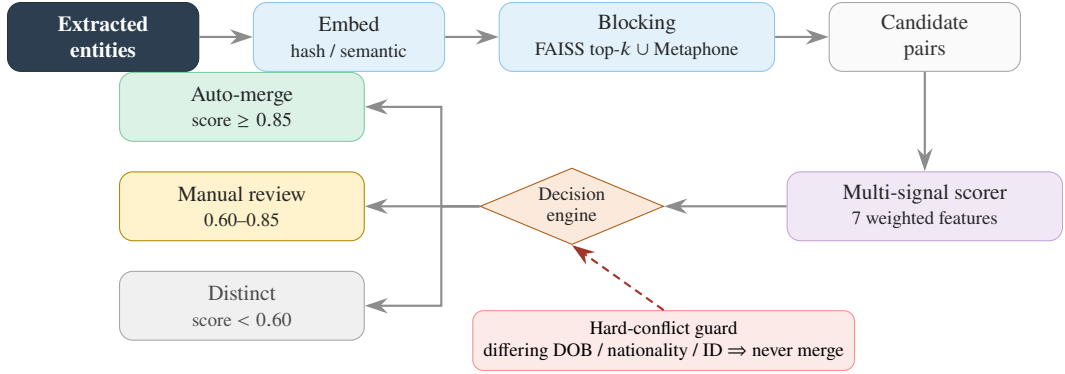
\begin{figure}[htbp]
\centering
\resizebox{\textwidth}{!}{%
\begin{tikzpicture}[node distance=0.6cm and 0.7cm]
  \node[ginput, minimum width=2.5cm] (ent) {Extracted\\entities};
  \node[gproc, minimum width=2.5cm, right=of ent] (emb) {Embed\\{\scriptsize hash / semantic}};
  \node[gproc, minimum width=3.6cm, right=of emb] (block) {Blocking\\{\scriptsize FAISS top-$k$ $\cup$ Metaphone}};
  \node[gdata, minimum width=2.5cm, right=of block] (pairs) {Candidate\\pairs};
  \node[gproc2, minimum width=3.6cm, below=1.3cm of pairs] (score)
    {Multi-signal scorer\\{\scriptsize 7 weighted features}};
  \node[gdiamond, minimum width=2.4cm, left=1.6cm of score] (dec) {Decision\\engine};
  \node[gmerge,    minimum width=3.2cm, left=1.5cm of dec, yshift= 1.3cm] (merge) {Auto-merge\\{\scriptsize score $\geq 0.85$}};
  \node[greview,   minimum width=3.2cm, left=1.5cm of dec]               (rev)   {Manual review\\{\scriptsize 0.60--0.85}};
  \node[gdistinct, minimum width=3.2cm, left=1.5cm of dec, yshift=-1.3cm] (dist)  {Distinct\\{\scriptsize score $<0.60$}};
  \node[gnote, fill=s5col!12, draw=s5col!60, below=0.85cm of dec, xshift=1.2cm] (hc)
    {Hard-conflict guard\\{\scriptsize differing DOB / nationality / ID $\Rightarrow$ never merge}};

  \draw[gflow] (ent)--(emb);
  \draw[gflow] (emb)--(block);
  \draw[gflow] (block)--(pairs);
  \draw[gflow] (pairs)--(score);
  \draw[gflow] (score)--(dec);
  \draw[gflow] (dec.west) -- ++(-0.5,0) |- (merge.east);
  \draw[gflow] (dec.west) -- (rev.east);
  \draw[gflow] (dec.west) -- ++(-0.5,0) |- (dist.east);
  \draw[gflow, densely dashed, draw=s5col!70!black] (hc.north) -- (dec.south);
\end{tikzpicture}%
}
\caption{Embedding-based entity resolution. Entities are embedded and blocked by the union of a FAISS cosine index and a Double Metaphone phonetic block; each candidate pair is scored by a weighted combination of seven signals; a decision engine routes the pair to auto-merge, manual review, or distinct under tunable thresholds. A hard-conflict guard on discriminator attributes overrides the score and forbids merging regardless of name similarity.}
\label{fig:embedding-resolution}
\end{figure}

\paragraph{Relationship to the Rule-Based Layer.}
The two layers are complementary rather than redundant: the rule-based algorithms run first and cheaply collapse obvious surface variants and expand aliases, while the embedding layer catches semantically equivalent mentions that survive string matching and does so under blocking to remain tractable at scale. Both feed the same review queue, and the hard-conflict guard ensures the more aggressive embedding layer cannot override discriminating evidence. Fig.~\ref{fig:embedding-resolution} depicts the pipeline.

\FloatBarrier

\section{Pipeline architecture and quality refinement}

\subsection{Five-stage pipeline architecture}
\label{sec:pipeline-architecture}

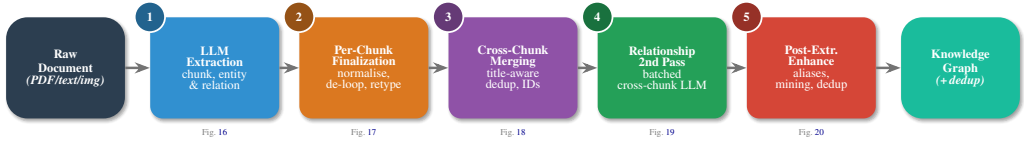
\begin{figure}[htbp]
\centering
\resizebox{0.96\textwidth}{!}{%
\begin{tikzpicture}[
  pstage/.style={rectangle, rounded corners=7pt, minimum width=2.55cm, minimum height=2.0cm,
                 align=center, text=white, font=\scriptsize,
                 blur shadow={shadow blur steps=4, shadow xshift=0.4mm, shadow yshift=-0.5mm}},
  pcap/.style={rectangle, rounded corners=12pt, minimum width=2.35cm, minimum height=2.0cm,
               align=center, font=\bfseries\scriptsize, text=white,
               blur shadow={shadow blur steps=4, shadow xshift=0.4mm, shadow yshift=-0.5mm}},
  pbadge/.style={circle, draw=white, line width=1pt, fill=#1, text=white,
                 font=\bfseries\scriptsize, minimum size=0.6cm, inner sep=0pt},
  pref/.style={font=\tiny, text=black!50},
  pflow/.style={-{Stealth[length=3mm, width=2.2mm]}, line width=1.4pt, draw=black!55},
]
  \node[pcap, fill=inputbg] (in) at (0,0)
    {Raw\\Document\\[-1pt]{\scriptsize\itshape (PDF/text/img)}};
  \node[pstage, fill=s1col!95!black] (s1) at (2.95,0)
    {\textbf{LLM}\\[-2pt]\textbf{Extraction}\\[-1pt]{\scriptsize chunk, entity}\\[-2pt]{\scriptsize \& relation}};
  \node[pstage, fill=s2col!95!black] (s2) at (5.9,0)
    {\textbf{Per-Chunk}\\[-2pt]\textbf{Finalization}\\[-1pt]{\scriptsize normalise,}\\[-2pt]{\scriptsize de-loop, retype}};
  \node[pstage, fill=s3col!92!black] (s3) at (8.85,0)
    {\textbf{Cross-Chunk}\\[-2pt]\textbf{Merging}\\[-1pt]{\scriptsize title-aware}\\[-2pt]{\scriptsize dedup, IDs}};
  \node[pstage, fill=s4col!92!black] (s4) at (11.8,0)
    {\textbf{Relationship}\\[-2pt]\textbf{2nd Pass}\\[-1pt]{\scriptsize batched}\\[-2pt]{\scriptsize cross-chunk LLM}};
  \node[pstage, fill=s5col!92!black] (s5) at (14.75,0)
    {\textbf{Post-Extr.}\\[-2pt]\textbf{Enhance}\\[-1pt]{\scriptsize aliases,}\\[-2pt]{\scriptsize mining, dedup}};
  \node[pcap, fill=outputbg] (out) at (17.7,0)
    {Knowledge\\Graph\\[-1pt]{\scriptsize\itshape (+\,dedup)}};

  \node[pbadge=s1col!65!black] at (s1.north west) {1};
  \node[pbadge=s2col!65!black] at (s2.north west) {2};
  \node[pbadge=s3col!65!black] at (s3.north west) {3};
  \node[pbadge=s4col!65!black] at (s4.north west) {4};
  \node[pbadge=s5col!65!black] at (s5.north west) {5};

  \node[pref, below=0.08cm of s1.south] {Fig.~\ref{fig:stage1}};
  \node[pref, below=0.08cm of s2.south] {Fig.~\ref{fig:stage2}};
  \node[pref, below=0.08cm of s3.south] {Fig.~\ref{fig:stage3}};
  \node[pref, below=0.08cm of s4.south] {Fig.~\ref{fig:stage4}};
  \node[pref, below=0.08cm of s5.south] {Fig.~\ref{fig:stage5}};

  \foreach \a/\b in {in/s1, s1/s2, s2/s3, s3/s4, s4/s5, s5/out}
    \draw[pflow] (\a) -- (\b);
\end{tikzpicture}%
}
\caption{High-level overview of the five-stage extraction pipeline. A document enters at the left and flows strictly left to right through chunked LLM extraction, deterministic per-chunk cleaning, cross-chunk merging, a relationship second pass, and post-extraction enhancements, emerging as a validated knowledge graph. Numbered badges key each stage to its detail figure (Figs.~\ref{fig:stage1}--\ref{fig:stage5}).}
\label{fig:pipeline}
\end{figure}

Fig.~\ref{fig:pipeline} illustrates the complete five-stage extraction pipeline, showing the flow from raw document input through LLM extraction, data cleaning, cross-chunk merging, relationship enrichment, and post-extraction entity enhancement. Stages~2 and~3 (highlighted in orange and purple) represent the data cleaning fixes introduced after empirical evaluation (Section~\ref{sec:quality-fixes}), while Stage~5 (red) encompasses the six core deduplication algorithms detailed in Section~\ref{sec:dedup}.

The extraction pipeline transforms raw documents into a validated JSON knowledge graph through five sequential stages.
Each stage addresses a distinct class of data-quality concern, from initial LLM extraction through normalisation, deduplication, cross-chunk relationship discovery, and post-extraction enrichment.
The following subsections present the internal structure of each stage.

\begin{figure}[htbp]
\centering
\resizebox{\textwidth}{!}{%
\begin{tikzpicture}[node distance=0.5cm]
  \node[inputcard] (in) {Raw Document Input\\[-2pt]{\scriptsize\itshape PDF\,/\,Text\,/\,Image}};

  \node[s1card, below=1.7cm of in] (chunk)
    {Document Chunking\\[-2pt]{\scriptsize\texttt{chunk\_pages=5}, overlap \texttt{neighbor=1}}};
  \node[s1card, below=of chunk] (p1)
    {Phase~1: Entity Extraction\\[-2pt]{\scriptsize LLM extracts typed entities from chunk text}};
  \node[s1card, below=of p1] (p2)
    {Phase~2: Relationship Extraction\\[-2pt]{\scriptsize Dedicated LLM call with entity catalog + source text}};

  \begin{scope}[on background layer]
    \node[stagebox=s1col, fit=(chunk)(p1)(p2),
          label={[stitle=s1col]above:{\large Stage 1}\; LLM Extraction {\scriptsize\itshape (Per Chunk)}}] {};
  \end{scope}

  \node[bugcard, right=2.3cm of p2] (bug1)
    {\textbf{Infrastructure Fix}\\[2pt]\texttt{\_env\_int()} bug:\\
     \texttt{max(1,\,0)} truncated source\\text to 1~character};
  \draw[bugline] (bug1.west) -- (p2.east);

  \draw[flowline] (in) -- (chunk);
  \draw[flowline] (chunk) -- (p1);
  \draw[flowline] (p1) -- (p2);
\end{tikzpicture}%
}
\caption{Stage~1: LLM Extraction. The document is split into overlapping chunks; two sequential LLM calls extract entities and relationships per chunk. The dashed annotation marks a critical infrastructure bug where \texttt{\_env\_int()} defaulted the context window to one character, suppressing all per-chunk relationships.}
\label{fig:stage1}
\end{figure}
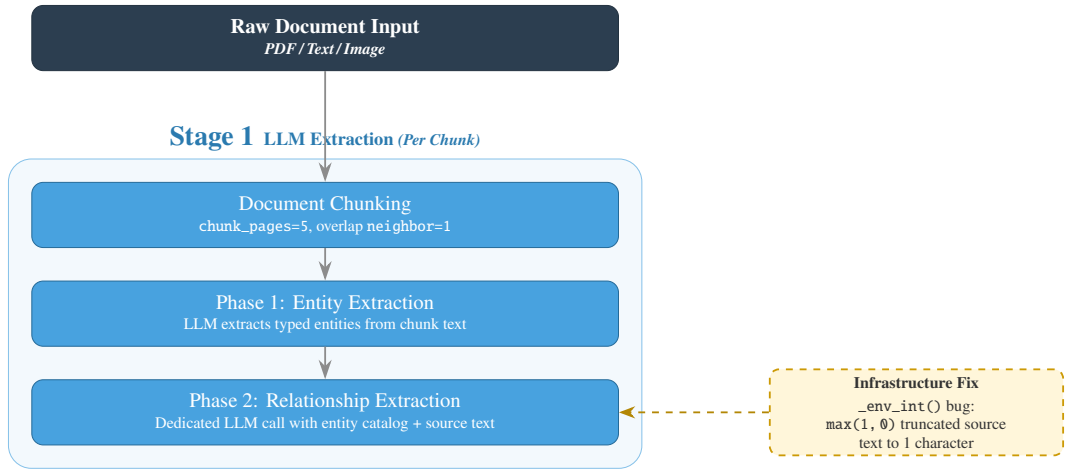

\paragraph{Stage~1: LLM Extraction.}
The pipeline begins by splitting the input document into overlapping chunks of five pages each, with one page of overlap on each side to preserve cross-page context.
For each chunk, two sequential LLM calls are issued:
\begin{enumerate}
  \item \textbf{Entity Extraction}: the LLM identifies named entities (persons, organisations, locations, equipment) together with their properties and provisional aliases.
  \item \textbf{Relationship Extraction}: a second, dedicated LLM call receives the full entity catalog extracted so far plus the raw chunk text, and returns typed relationships with qualifiers and evidence spans.
\end{enumerate}
A critical infrastructure bug was discovered during evaluation: the helper function \texttt{\_env\_int()} used \texttt{max(1,\,0)} instead of the configured context-window size, truncating the source text fed to the relationship prompt to a single character and producing zero per-chunk relationships.

\begin{figure}[htbp]
\centering
\begin{tikzpicture}[node distance=0.5cm]
  \node[s2card] (f2)
    {Normalize Relationship Types\\[-2pt]{\scriptsize UPPER\_SNAKE\_CASE conversion + typo correction}\\[-1pt]{\scriptsize\texttt{infiltrated $\to$ INFILTRATED}, \texttt{surveiled $\to$ SURVEILLED}}};
  \node[s2card, below=of f2] (f3)
    {Remove Self-Referencing Relationships\\[-2pt]{\scriptsize Filter \texttt{source == target} loops}\\[-1pt]{\scriptsize\texttt{per\_001 $\to$ per\_001} \texttimes}};
  \node[s2card, below=of f3] (f4)
    {Correct Entity Type Misassignment\\[-2pt]{\scriptsize Name-pattern-based type inference}\\[-1pt]{\scriptsize\texttt{`Leviathan Submarine' Person $\to$ Equipment}}};

  \begin{scope}[on background layer]
    \node[stagebox=s2col, fit=(f2)(f3)(f4),
          label={[stitle=s2col]above:{\large Stage 2}\; Per-Chunk Finalization {\scriptsize\itshape (Data Cleaning)}}] {};
  \end{scope}

  \draw[flowline] (f2) -- (f3);
  \draw[flowline] (f3) -- (f4);
\end{tikzpicture}
\caption{Stage~2: Per-Chunk Finalization. Three deterministic cleaning passes normalise relationship types, remove self-loops, and correct entity type misassignments before cross-chunk merging.}
\label{fig:stage2}
\end{figure}
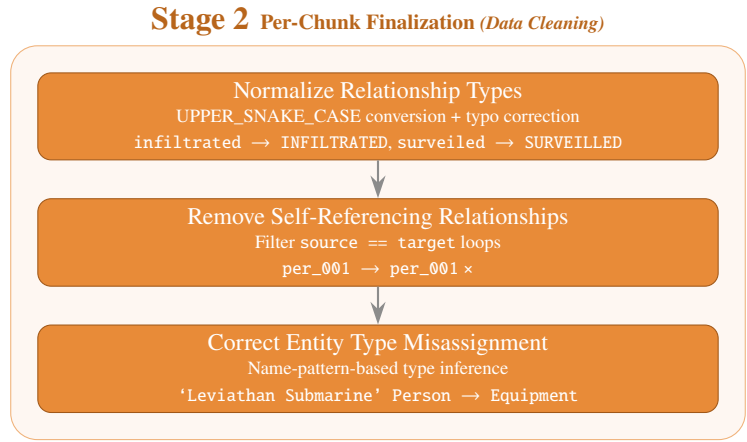

\paragraph{Stage~2: Per-Chunk Finalization.}
Before merging results across chunks, three deterministic cleaning passes are applied to each chunk's output:
\begin{enumerate}
  \item \textbf{Relationship-type normalisation} converts free-form LLM relationship labels into a controlled \texttt{UPPER\_SNAKE\_CASE} vocabulary and corrects common misspellings (e.g.\ \texttt{surveiled} $\to$ \texttt{SURVEILLED}).
  \item \textbf{Self-loop removal} filters out relationships where the source and target entity are identical, an artefact of LLM hallucination.
  \item \textbf{Entity-type correction} uses name-pattern heuristics to fix misclassified entities. For example, an entity named ``Leviathan Submarine'' initially typed as \textsc{Person} is reassigned to \textsc{Equipment} based on the suffix keyword ``Submarine''.
\end{enumerate}

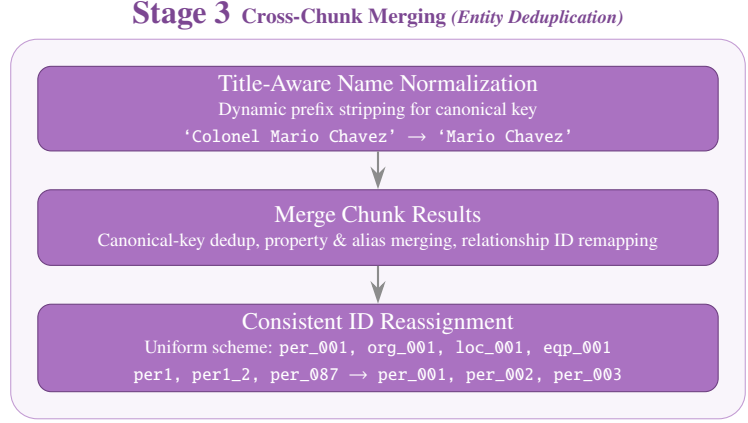
\begin{figure}[htbp]
\centering
\begin{tikzpicture}[node distance=0.5cm]
  \node[s3card] (f1)
    {Title-Aware Name Normalization\\[-2pt]{\scriptsize Dynamic prefix stripping for canonical key}\\[-1pt]{\scriptsize\texttt{`Colonel Mario Chavez' $\to$ `Mario Chavez'}}};
  \node[s3card, below=of f1] (merge)
    {Merge Chunk Results\\[-2pt]{\scriptsize Canonical-key dedup, property \& alias merging, relationship ID remapping}};
  \node[s3card, below=of merge] (f5)
    {Consistent ID Reassignment\\[-2pt]{\scriptsize Uniform scheme: \texttt{per\_001, org\_001, loc\_001, eqp\_001}}\\[-1pt]{\scriptsize\texttt{per1, per1\_2, per\_087 $\to$ per\_001, per\_002, per\_003}}};

  \begin{scope}[on background layer]
    \node[stagebox=s3col, fit=(f1)(merge)(f5),
          label={[stitle=s3col]above:{\large Stage 3}\; Cross-Chunk Merging {\scriptsize\itshape (Entity Deduplication)}}] {};
  \end{scope}

  \draw[flowline] (f1) -- (merge);
  \draw[flowline] (merge) -- (f5);
\end{tikzpicture}
\caption{Stage~3: Cross-Chunk Merging. Title-aware normalisation produces canonical keys; entities sharing the same key are merged with property and alias unification; finally, all IDs are reassigned to a consistent scheme.}
\label{fig:stage3}
\end{figure}

\paragraph{Stage~3: Cross-Chunk Merging.}
This stage unifies entity records that were extracted independently from different chunks:
\begin{enumerate}
  \item \textbf{Title-aware name normalisation} strips honorifics and titles (``Colonel'', ``Dr.'', ``Agent'') to produce a canonical key. This prevents the same person from appearing under both ``Colonel Mario Chavez'' and ``Mario Chavez''.
  \item \textbf{Chunk-result merging} groups entities by canonical key and merges their properties, aliases, and associated relationships. Relationship source/target IDs are remapped to the merged entity.
  \item \textbf{Consistent ID reassignment} replaces the heterogeneous IDs produced by per-chunk extraction (\texttt{per1}, \texttt{per1\_2}, \texttt{per\_087}) with a uniform \texttt{<type>\_<seq>} scheme.
\end{enumerate}

\begin{figure}[htbp]
\centering
\begin{tikzpicture}[node distance=0.5cm]
  \node[s4card] (batch)
    {Batch Entity Catalog\\[-2pt]{\scriptsize \texttt{batch\_size=80}, \texttt{max\_batches=8}}};
  \node[s4card, below=of batch] (rellm)
    {Cross-Chunk LLM Relationship Extraction\\[-2pt]{\scriptsize Evidence-based discovery of inter-chunk relationships}};
  \node[s4card, below=of rellm] (reldd)
    {Relationship Deduplication\\[-2pt]{\scriptsize Qualifier-aware merging by \texttt{(source, target, type)}}\\[-1pt]{\scriptsize Fig.~\ref{alg:rel-dedup}}};

  \begin{scope}[on background layer]
    \node[stagebox=s4col, fit=(batch)(rellm)(reldd),
          label={[stitle=s4col]above:{\large Stage 4}\; Relationship Second Pass}] {};
  \end{scope}

  \draw[flowline] (batch) -- (rellm);
  \draw[flowline] (rellm) -- (reldd);
\end{tikzpicture}
\caption{Stage~4: Relationship Second Pass. The merged entity catalog is batched and sent to the LLM for cross-chunk relationship discovery; duplicate relationships are then merged using qualifier-aware deduplication (Fig.~\ref{alg:rel-dedup}).}
\label{fig:stage4}
\end{figure}
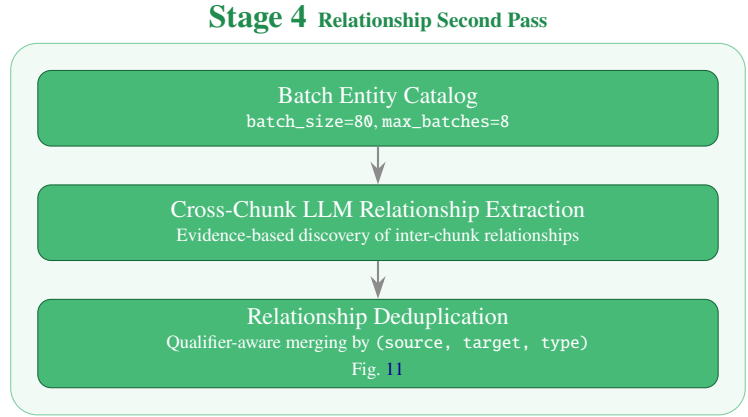

\paragraph{Stage~4: Relationship Second Pass.}
Per-chunk extraction can only discover relationships whose participants co-occur within a single chunk.
Stage~4 performs a second LLM pass over the \emph{merged} entity catalog to discover cross-chunk relationships:
\begin{enumerate}
  \item \textbf{Batching}: the full entity catalog is partitioned into batches of up to 80~entities (at most 8~batches) to stay within context-window limits.
  \item \textbf{Cross-chunk LLM extraction}: each batch is sent to the LLM alongside the source text, requesting evidence-grounded relationships between entities that may have originated from different chunks.
  \item \textbf{Relationship deduplication}: newly discovered and previously existing relationships are deduplicated by the composite key \texttt{(source, target, type)}, with qualifier fields merged using Fig.~\ref{alg:rel-dedup}.
\end{enumerate}

\begin{figure}[htbp]
\centering
\resizebox{\textwidth}{!}{%
\begin{tikzpicture}[node distance=0.5cm]
  \node[s5card] (a1)
    {Alias Expansion (Fig.~\ref{alg:alias-expansion})\\[-2pt]{\scriptsize Deterministic abbreviation generation}\\[-1pt]{\scriptsize\texttt{`Elena Petrov' $\to$ `E.\,Petrov', `Elena P.', `EP', `E.P.'}}};
  \node[s5card, below=of a1] (a2)
    {PDF Text Mining (Fig.~\ref{alg:pdf-text-mining})\\[-2pt]{\scriptsize Fuzzy matching for spelling variants in source text}\\[-1pt]{\scriptsize Discovers \texttt{`Jon'} when \texttt{`John'} was extracted}};
  \node[s5card, below=of a2] (a3)
    {Similarity Scoring (Fig.~\ref{alg:name-similarity})\\[-2pt]{\scriptsize Multi-tier: exact $\to$ Slavic suffix $\to$ abbreviation $\to$ sequence}\\[-1pt]{\scriptsize\texttt{`Petrov' vs `Petrova' $\to$ 0.85 slavic\_gender\_suffix}}};
  \node[s5card, below=of a3] (a4)
    {Context-Aware Dedup (Fig.~\ref{alg:context-dedup})\\[-2pt]{\scriptsize Role/org/location conflict validation}\\[-1pt]{\scriptsize Same name + different org $\to$ \texttt{`different\_people'}}};

  \begin{scope}[on background layer]
    \node[stagebox=s5col, fit=(a1)(a2)(a3)(a4),
          label={[stitle=s5col]above:{\large Stage 5}\; Post-Extraction Enhancements {\scriptsize\itshape (Data Enrichment)}}] {};
  \end{scope}

  \node[bugcard, right=2.3cm of a3] (bug2)
    {\textbf{Similarity-scorer Bug Fix}\\[2pt]Abbreviation check required\\only first-letter match\\$\to$ 79~false positives};
  \draw[bugline] (bug2.west) -- (a3.east);

  \draw[flowline] (a1) -- (a2);
  \draw[flowline] (a2) -- (a3);
  \draw[flowline] (a3) -- (a4);
\end{tikzpicture}%
}
\caption{Stage~5: Post-Extraction Enhancements. Four algorithms enrich the knowledge graph with expanded aliases, text-mined spelling variants, similarity scores, and context-aware deduplication decisions. The dashed annotation marks a bug in the similarity scorer (Fig.~\ref{alg:name-similarity}) where an overly permissive abbreviation check produced 79~false-positive merge candidates.}
\label{fig:stage5}
\end{figure}
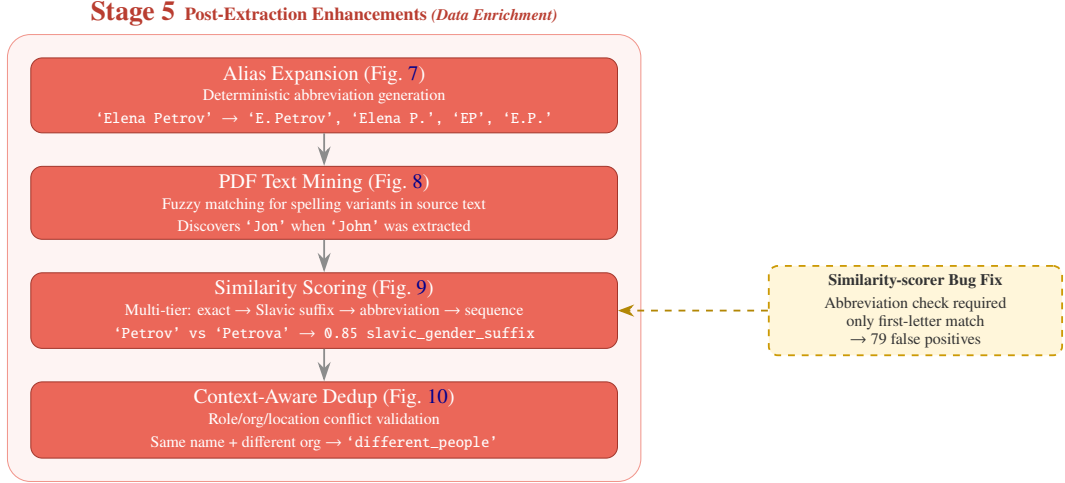

\paragraph{Stage~5: Post-Extraction Enhancements.}
The final stage enriches the knowledge graph with alias information and deduplication candidates through four algorithms:
\begin{enumerate}
  \item \textbf{Alias Expansion} (Fig.~\ref{alg:alias-expansion}) deterministically generates abbreviations and name variants (e.g.\ ``Elena Petrov'' $\to$ ``E.\,Petrov'', ``EP'') to increase recall during downstream entity linking.
  \item \textbf{PDF Text Mining} (Fig.~\ref{alg:pdf-text-mining}) performs fuzzy matching against the raw source text to discover spelling variants that the LLM normalised away (e.g.\ ``Jon'' vs.\ ``John'').
  \item \textbf{Similarity Scoring} (Fig.~\ref{alg:name-similarity}) computes a multi-tier similarity score between all entity-name pairs using exact matching, Slavic gender-suffix handling, abbreviation detection, and sequence alignment. A bug in the abbreviation check, which required only a first-letter match, produced 79~false positives before correction.
  \item \textbf{Context-Aware Dedup Detection} (Fig.~\ref{alg:context-dedup}) validates high-similarity pairs against contextual evidence (role, organisation, location) to distinguish true duplicates from distinct entities who share a name.
\end{enumerate}

\begin{figure}[htbp]
\centering
\begin{tikzpicture}
  \node[outputcard, minimum width=10cm, minimum height=1.2cm] (out)
    {Validated JSON Knowledge Graph\\[-2pt]{\scriptsize\itshape Entities + Relationships + Dedup Candidates}};
\end{tikzpicture}
\caption{Pipeline output: a validated JSON knowledge graph containing typed entities, evidence-linked relationships, and scored deduplication candidates ready for downstream analysis.}
\label{fig:output}
\end{figure}

After all five stages complete, the pipeline emits a single JSON knowledge graph containing:
\begin{itemize}
  \item \textbf{Entities} with normalised names, consistent IDs, merged properties, and expanded alias sets.
  \item \textbf{Relationships} with typed labels, qualifier metadata, and evidence spans linking back to the source document.
  \item \textbf{Deduplication candidates} scored by similarity and annotated with context-aware verdicts (\texttt{same\_person}, \texttt{different\_people}, \texttt{uncertain}).
\end{itemize}

\FloatBarrier
\subsection{Example extraction output}

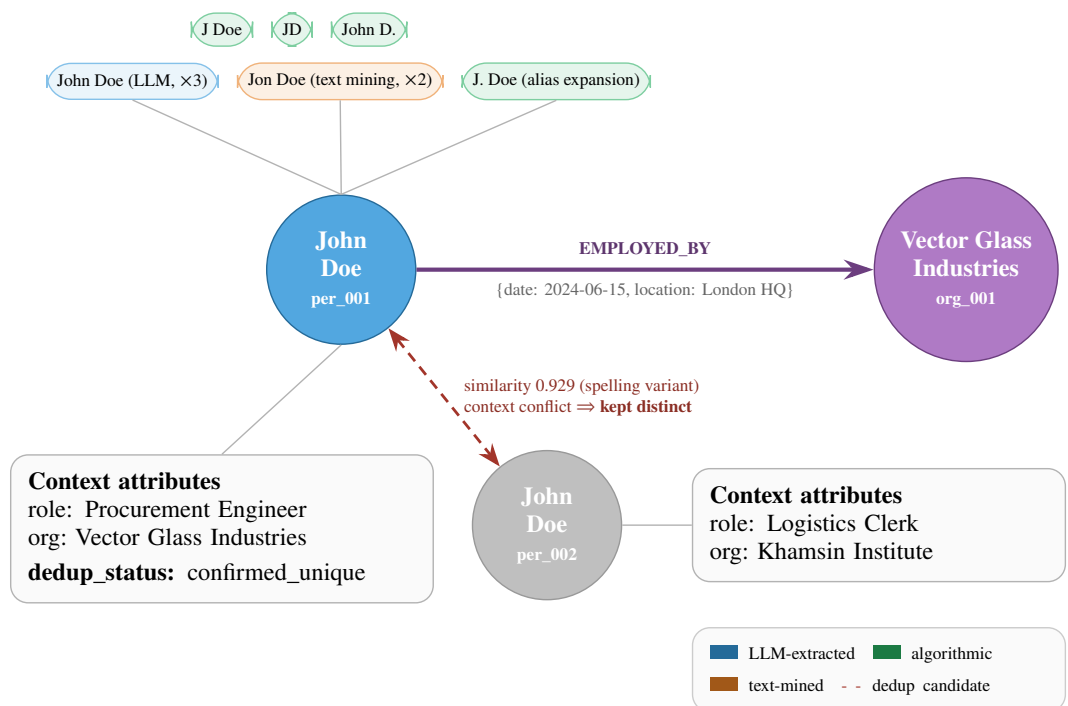
\begin{figure}[htbp]
\centering
\resizebox{\textwidth}{!}{%
\begin{tikzpicture}[
  node distance=0.55cm and 1.1cm,
  persona/.style={circle, draw=s1col!70!black, fill=s1col!85, text=white,
                  font=\bfseries\scriptsize, minimum size=1.5cm, align=center},
  orga/.style={persona, draw=s3col!70!black, fill=s3col!80},
  chip/.style={rectangle, rounded corners=7pt, draw=black!30, fill=black!5,
               font=\tiny, inner sep=3pt},
  chipllm/.style={chip, draw=s1col!60, fill=s1col!10},
  chipalg/.style={chip, draw=s4col!60, fill=s4col!12},
  chipmine/.style={chip, draw=s2col!60, fill=s2col!12},
  panel/.style={rectangle, rounded corners=5pt, draw=black!30, fill=cardbg,
                font=\scriptsize, align=left, inner sep=5pt},
]
  \node[persona] (p1) {John\\Doe\\{\tiny per\_001}};
  \node[orga, right=4.6cm of p1] (org) {Vector Glass\\Industries\\{\tiny org\_001}};
  \node[persona, draw=black!40, fill=black!25, below right=1.5cm and 1.0cm of p1] (p2)
    {John\\Doe\\{\tiny per\_002}};

  \draw[gflow, very thick, draw=s3col!70!black] (p1) --
    node[above, font=\bfseries\tiny, text=s3col!60!black]{EMPLOYED\_BY}
    node[below, font=\tiny, text=black!60]{\{date: 2024-06-15, location: London HQ\}}
    (org);

  \draw[densely dashed, thick, draw=s5col!70!black, {Stealth}-{Stealth}] (p1) --
    node[right=2pt, font=\tiny, align=left, text=s5col!60!black]
      {similarity 0.929 (spelling variant)\\context conflict $\Rightarrow$ \textbf{kept distinct}}
    (p2);

  \node[chipllm,  above=0.95cm of p1, xshift=-2.1cm] (a1) {John Doe {\tiny(LLM, $\times$3)}};
  \node[chipmine, right=0.18cm of a1] (a2) {Jon Doe {\tiny(text mining, $\times$2)}};
  \node[chipalg,  right=0.18cm of a2] (a3) {J.\ Doe {\tiny(alias expansion)}};
  \node[chipalg,  above=0.16cm of a1, xshift=0.9cm] (a4) {J Doe};
  \node[chipalg,  right=0.18cm of a4] (a5) {JD};
  \node[chipalg,  right=0.18cm of a5] (a6) {John D.};
  \foreach \c in {a1,a2,a3} \draw[black!30, thin] (\c.south) -- (p1.north);

  \node[panel, below=1.1cm of p1, xshift=-1.2cm, text width=3.9cm] (ctx)
    {\textbf{Context attributes}\\
     role: Procurement Engineer\\
     org:\ Vector Glass Industries\\[2pt]
     \textbf{dedup\_status:} confirmed\_unique};
  \draw[black!30, thin] (ctx.north) -- (p1.south);

  \node[panel, right=0.7cm of p2, text width=3.4cm] (ctx2)
    {\textbf{Context attributes}\\
     role: Logistics Clerk\\
     org:\ Khamsin Institute};
  \draw[black!30, thin] (ctx2.west) -- (p2.east);

  \node[panel, draw=black!20, below=0.45cm of ctx2, text width=3.4cm] {%
    {\tiny \textcolor{s1col!70!black}{\rule{8pt}{4pt}}\, LLM-extracted\quad
           \textcolor{s4col!70!black}{\rule{8pt}{4pt}}\, algorithmic\\[1pt]
           \textcolor{s2col!70!black}{\rule{8pt}{4pt}}\, text-mined\quad
           \textcolor{s5col!70!black}{- -}\, dedup candidate}};
\end{tikzpicture}%
}
\caption{A single output record rendered as the graph fragment it becomes. The entity carries six aliases colour-coded by discovery mechanism (LLM extraction, algorithmic expansion, source-text mining, each with occurrence counts), a qualified \textsc{Employed\_By} relationship, and contextual attributes. The dashed edge is the system's restraint on display: despite a 0.929 name similarity to \texttt{per\_002}, conflicting role and organisation evidence keeps the two people distinct; the pair is exported in \texttt{dedup\_candidates} for audit rather than silently merged.}
\label{fig:output-example}
\end{figure}

The output payload captures, for every entity, its full alias family \emph{with provenance} (which discovery method found each variant, and how often), the contextual attributes that anchor its identity, and any scored duplicate candidates with a recommended action. Fig.~\ref{fig:output-example} renders one representative record as the graph fragment it becomes: the person \texttt{per\_001} carries six aliases discovered by three different mechanisms, is linked to his employer by a qualified relationship, and is connected to the similarly named \texttt{per\_002} only by a \emph{dashed} dedup edge: the 0.929 name similarity was overridden by conflicting context (different roles, different organisations), so the two remain distinct people.

\FloatBarrier
\subsection{Empirical quality analysis and post-extraction data cleaning}
\label{sec:quality-fixes}

While the six core deduplication algorithms (Section~\ref{sec:dedup}) address entity-level name variation and relationship deduplication, empirical evaluation on a 45-page intelligence report (IR-001.pdf, yielding 579 entities and 540 relationships) revealed a critical infrastructure bug together with six data-quality issues originating from LLM output inconsistencies. These issues persisted despite the existing algorithms because they occur \emph{upstream}, in the raw LLM output, the chunk merging logic, or the utility infrastructure, before the deduplication algorithms execute. This section documents each issue, analyzes why the existing algorithms failed to catch it, and describes the applied fix. The architectural placement of these fixes within the pipeline is illustrated in Fig.~\ref{fig:pipeline} (Stages~2--3).

\subsubsection{Critical infrastructure bug: source text truncation}

\paragraph{Issue.} All per-chunk relationship extractions (Phase~2) returned zero relationships despite the LLM receiving 23--134 entities per chunk. Tracing the data through the call boundary exposed a striking discontinuity (Table~\ref{tab:truncation-evidence}): the caller handed over the full chunk text, yet the relationship prompt saw a single character.

\begin{table}[h]
\centering
\caption{Observed evidence of the source-truncation bug on a representative chunk.}
\label{tab:truncation-evidence}
\begin{tabular}{@{}lcc@{}}
\toprule
\rowcolor{s5col!12}\textbf{Observation point} & \textbf{Expected} & \textbf{Observed} \\
\midrule
Source chars passed by caller        & 37{,}542 & 37{,}542 \\
Source chars seen inside extraction  & 37{,}542 & \textbf{1} \\
Entities provided to the LLM         & 23       & 23 \\
Relationships returned               & 22--38   & \textbf{0} \\
\bottomrule
\end{tabular}
\end{table}

\paragraph{Root Cause.} The configuration reader applied a well-intentioned safety guard to \emph{every} integer environment variable:
\[
  g(v) \;=\; \max(1,\, v),
\]
intended to prevent division-by-zero in unrelated callers. But the variable \texttt{REL\_EXTRACTION\_MAX\_SOURCE\_CHARS} uses the sentinel $v = 0$ to mean \emph{unlimited}; the guard silently rewrote the sentinel, $g(0) = 1$, and the downstream slice $S[\,{:}\,g(0)] = S[\,{:}\,1]$ reduced thirty-seven thousand characters of evidence to the single letter ``P''. The relationship model, asked to find relations in a one-character document, correctly returned none, which made the failure look like a model deficiency rather than an infrastructure bug.

\paragraph{Fix.} The guard was removed, restoring the sentinel semantics $g(v) = v$ (with non-numeric values falling back to the default). No other caller was affected, as every other integer default in the codebase is $\geq 1$.

\paragraph{Impact.} Per-chunk relationship extraction immediately recovered, producing 22--38+ relationships per chunk.

\subsubsection{Fix 1: cross-chunk title-based entity duplication}

\paragraph{Issue.} The same real-world entity appeared as multiple distinct entities when the LLM extracted it with different title prefixes across chunks. Table~\ref{tab:title-duplication} shows representative examples.

\begin{table}[h]
\centering
\caption{Title-Based Entity Duplication Examples from IR-001.pdf}
\label{tab:title-duplication}
\footnotesize
\begin{tabular}{@{}lll@{}}
\toprule
\textbf{Chunk A Entity} & \textbf{Chunk B Entity} & \textbf{Chunk C Entity} \\
\midrule
\texttt{per\_087}: ``Mario Chavez'' & \texttt{per85\_3}: ``Colonel Mario Chavez'' & --- \\
\texttt{per48\_2}: ``Nicole Trujillo'' & \texttt{per47}: ``Dr.\ Nicole Trujillo'' & \texttt{per80}: ``Agent Dr.\ Nicole Trujillo'' \\
\texttt{per5}: ``Bianca Knapp'' & \texttt{per2\_2}: ``General Bianca Knapp'' & --- \\
\bottomrule
\end{tabular}
\end{table}

\noindent\textbf{Scale:} 91 duplicate pairs detected, inflating the Person entity count from $\sim$224 to 315 ($\sim$30\% inflation).

\paragraph{Why Fig.~\ref{alg:context-dedup} Did Not Catch It.} The context-aware duplicate detection algorithm \emph{did} flag 66 of these 91 pairs but systematically classified every one as \texttt{"different\_people\_different\_contexts"} with low confidence. This occurred because: (1)~the SequenceMatcher similarity between ``Mario Chavez'' and ``Colonel Mario Chavez'' is only 0.75, falling below the 0.95 auto-merge threshold; (2)~the algorithm compared full strings without recognizing that one is a title-prefixed variant of the other; (3)~25 pairs fell below the 0.75 detection threshold entirely.

\paragraph{Why the Chunk Merging Did Not Catch It.} The canonical entity key used for cross-chunk deduplication compared names verbatim: \texttt{("person", "colonel mario chavez")} $\neq$ \texttt{("person", "mario chavez")}.

\paragraph{Fix.} Introduced a dynamic name normalisation that strips leading prefix tokens until a two-word core name remains, \emph{without} a hardcoded title list. Writing a name as the token sequence $w_1 w_2 \cdots w_n$, the normal form is computed by the recurrence
\[
  \mathrm{norm}(w_1 \cdots w_n) \;=\;
  \begin{cases}
    \mathrm{norm}(w_2 \cdots w_n) & \text{if } n > 2 \,\wedge\, \mathrm{prefixlike}(w_1),\\[2pt]
    w_1 \cdots w_n & \text{otherwise},
  \end{cases}
\]
where $\mathrm{prefixlike}(w)$ holds when $w$ ends in a period (\,``Dr.''\,) or is a short alphabetic token ($|w| \leq 12$, covering ``Colonel'', ``General'', ``Agent''). Because the rule is structural rather than lexical, unseen titles normalise correctly: ``Agent Dr.\ Nicole Trujillo'' $\to$ ``Dr.\ Nicole Trujillo'' $\to$ ``Nicole Trujillo'', terminating at the two-word core. The merge logic preserves the titled variant as an alias while promoting the base name to primary: ``Mario Chavez'' with alias ``Colonel Mario Chavez''.

\subsubsection{Fix 2: relationship type inconsistency}

\paragraph{Issue.} The same relationship type appeared in multiple casings and formats across chunks, preventing proper deduplication by Fig.~\ref{alg:rel-dedup}.

\begin{table}[h]
\centering
\caption{Relationship Type Variants from IR-001.pdf}
\label{tab:reltype-variants}
\begin{tabular}{@{}lll@{}}
\toprule
\textbf{Variant 1} & \textbf{Variant 2} & \textbf{Variant 3} \\
\midrule
\texttt{INFILTRATED} (7) & \texttt{INfiltrated} (15) & \texttt{infiltrated} (4) \\
\texttt{REPORTED\_TO} (22) & \texttt{reported to} (5) & --- \\
\texttt{SURVEILLED} (64) & \texttt{surveilled} (1) & \texttt{surveiled} (1, typo) \\
\bottomrule
\end{tabular}
\end{table}

\paragraph{Root Cause.} Fig.~\ref{alg:rel-dedup} deduplicates by the key $(u, v, t)$ but compared type $t$ as a raw string without normalization.

\paragraph{Fix.} Every relationship type is canonicalised during finalisation by the mapping
\[
  t' \;=\; \tau\bigl(\textsc{Upper}(\textsc{Snake}(t))\bigr),
\]
where \textsc{Snake} collapses whitespace and hyphen runs to underscores, \textsc{Upper} uppercases, and $\tau$ is a small typo-correction table accumulated from observed LLM output ($\tau(\texttt{SURVEILED}) = \texttt{SURVEILLED}$, $\tau(\texttt{TRANSFERED}) = \texttt{TRANSFERRED}$). After canonicalisation the three \texttt{INFILTRATED} variants of Table~\ref{tab:reltype-variants} collapse to one key and deduplicate correctly.

\subsubsection{Fix 3: self-referencing relationships}

\paragraph{Issue.} 114 relationships (21\% of all 540) had \texttt{source == target}, meaning an entity was related to itself. One entity (``Dr.\ Nicole Trujillo'') had nine self-loops including \texttt{SURVEILLED}, \texttt{FUNDED}, \texttt{MONITORED}, and \texttt{DEPLOYED\_TO} pointing to herself.

\paragraph{Root Cause.} When the LLM cannot resolve the correct target entity for a relationship, it defaults to reusing the source entity ID, a known hallucination pattern observed more frequently with smaller local models than with the cloud-hosted alternative.

\paragraph{Fix.} Self-loops are removed by a single set filter during finalisation:
\[
  \rho' \;=\; \{\, (u, v, t) \in \rho \;:\; u \neq v \,\},
\]
which on IR-001.pdf discarded all 114 self-referencing relationships while leaving every legitimate edge untouched.

\subsubsection{Fix 4: entity type misassignment}

\paragraph{Issue.} 42 entities with clearly non-person names were assigned the type ``Person'' by the LLM. Table~\ref{tab:type-misassignment} shows representative examples.

\begin{table}[h]
\centering
\caption{Entity Type Misassignment Examples}
\label{tab:type-misassignment}
\small
\begin{tabular}{@{}llll@{}}
\toprule
\textbf{Entity Name} & \textbf{Assigned Type} & \textbf{Correct Type} & \textbf{Indicator Word} \\
\midrule
``Bank'' & Person & Organization & --- \\
``Leviathan Submarine LS-6720'' & Person & Equipment & \emph{submarine} \\
``Christopherchester Camp'' & Person & Location & \emph{camp} \\
``Agency'' & Person & Organization & \emph{agency} \\
\bottomrule
\end{tabular}
\end{table}

\paragraph{Root Cause.} In later chunks of long documents, the LLM's entity typing accuracy degrades, consistent with reported positional degradation in long contexts \cite{liu2024lost}. The model occasionally assigns ``Person'' as a default type.

\paragraph{Fix.} A post-extraction type inference function checks entity names for indicator words (e.g., \emph{camp}, \emph{laboratory} $\to$ Location; \emph{agency}, \emph{institute} $\to$ Organization; \emph{submarine}, \emph{drone} $\to$ Equipment). This runs only on entities typed as ``Person'' and only reclassifies when a strong indicator word is present, minimizing false corrections.

\subsubsection{Fix 5: inconsistent entity ID scheme}

\paragraph{Issue.} Entity IDs used at least seven different format patterns across chunks: \texttt{per1}, \texttt{per1\_2}, \texttt{per\_001}, \texttt{org\_020}, \texttt{e\_local\_5}, etc. This inconsistency arose because each chunk's LLM call independently generated entity IDs, and the merge function added collision suffixes without normalizing the base format.

\paragraph{Fix.} A final-stage ID reassignment function converts all entity IDs to a uniform \texttt{\{prefix\}\_\{number\}} format (e.g., \texttt{per\_001}, \texttt{org\_001}, \texttt{loc\_001}, \texttt{eqp\_001}) and remaps all relationship references accordingly. Additionally, relationships referencing entity IDs that do not exist in the final entity list (9 dangling references in IR-001.pdf) are filtered out.

\subsubsection{Fix 6: false positive abbreviation similarity matches}
\label{sec:fix6-abbrev}

\paragraph{Issue.} The abbreviation signal of Fig.~\ref{alg:name-similarity} produced 79 nonsensical matches with a hardcoded 0.92 similarity score. For example, ``Michael Cruz'' and ``Mario Chavez'' were flagged as \texttt{abbreviation\_variant} because both names have two words starting with `M' and `C' respectively.

\paragraph{Root Cause.} For two name parts $p$ and $q$, the buggy check accepted the pair whenever
\[
  \mathrm{match}_{\mathrm{buggy}}(p, q) \;\Longleftrightarrow\; p = q \;\vee\; p_1 = q_1,
\]
i.e.\ identical parts \emph{or merely identical first letters}. Under this predicate ``Michael''/``Mario'' match on `M' and ``Cruz''/``Chavez'' match on `C', so two entirely different people clear the abbreviation check and receive its hardcoded 0.92 score.

\paragraph{Fix.} A genuine abbreviation requires one side to actually \emph{be} a single-letter token. The corrected predicate is
\[
  \mathrm{match}(p, q) \;\Longleftrightarrow\; p = q \;\vee\; \bigl(|p| = 1 \wedge p = q_1\bigr) \;\vee\; \bigl(|q| = 1 \wedge q = p_1\bigr),
\]
and the pair as a whole is accepted only if at least one part-pair matched through the single-letter clause, i.e.\ an abbreviation was genuinely present. After the fix, ``E.\ Petrov'' vs.\ ``Elena Petrov'' still matches ($|{\rm E}| = 1 \wedge {\rm E} = {\rm Elena}_1$), while ``Michael Cruz'' vs.\ ``Mario Chavez'' is correctly rejected: no part is a single letter, so first-letter coincidence no longer suffices.

\subsubsection{Summary of quality fixes}

Table~\ref{tab:fix-summary} summarizes the six data cleaning fixes and the infrastructure bug. These fixes operate at Stages~2--3 of the pipeline (Fig.~\ref{fig:pipeline}), cleaning the raw LLM output \emph{before} the core deduplication algorithms (Stage~5) execute.

\begin{table}[h]
\centering
\caption{Summary of Post-Extraction Data Cleaning Fixes}
\label{tab:fix-summary}
\footnotesize
\begin{tabular}{@{}clp{4.5cm}cc@{}}
\toprule
\rowcolor{s2col!15}\textbf{\#} & \textbf{Fix} & \textbf{Root Cause} & \textbf{Entities} & \textbf{Rels} \\
\midrule
0 & \texttt{\_env\_int()} bug & \texttt{max(1,0)} truncated source text & 0 & \textbf{All} (0$\to$500+) \\
1 & Title normalization & Verbatim name comparison & 91 merged & Refs remapped \\
2 & Rel type normalization & No case normalization & 0 & $\sim$30 deduped \\
3 & Self-loop filtering & LLM hallucination & 0 & 114 removed \\
4 & Entity type correction & LLM accuracy degradation & 42 reclassified & 0 \\
5 & ID reassignment & Independent per-chunk IDs & All renumbered & 9 dangling removed \\
6 & Abbreviation fix & First-letter-only match & 0 & 0 (dedup fixed) \\
\bottomrule
\end{tabular}
\end{table}

\begin{table}[h]
\centering
\caption{Net Quality Impact on IR-001.pdf (45 pages)}
\label{tab:quality-impact}
\begin{tabular}{@{}lccc@{}}
\toprule
\rowcolor{s4col!15}\textbf{Metric} & \textbf{Before Fixes} & \textbf{After Fixes} & \textbf{Change} \\
\midrule
Total entities                  & 579            & $\sim$400         & $-$31\% (dedup)        \\
True Person count               & 315 (inflated) & $\sim$224         & $-$29\% (accurate)     \\
Per-chunk relationships         & 0              & 500+              & Restored               \\
Self-loop relationships         & 114 (21\%)     & 0                 & $-$100\%               \\
Broken entity references        & 9              & 0                 & $-$100\%               \\
Relationship type variants      & 39             & $\sim$28          & $-$28\% (normalized)   \\
False positive dedup candidates & 79             & $\sim$0           & $-$100\%               \\
Entity ID format patterns       & 7              & 1                 & Uniform                \\
\bottomrule
\end{tabular}
\end{table}

\paragraph{Key Insight.} The core deduplication algorithms (Figs.~\ref{alg:alias-expansion}--\ref{alg:context-dedup}) were correctly designed for entity-level refinement, but their effectiveness was undermined by upstream data quality issues in the raw LLM output and the chunk merging infrastructure. The data cleaning fixes (Stages~2--3) and the core algorithms (Stage~5) are complementary: the former ensures clean input data, while the latter enriches it. Together, they form a defense-in-depth approach where neither layer alone would be sufficient.


\section{Evaluation}

We evaluate the system on two axes: the impact of the deduplication and grounding enhancements on retrieval quality (Section~\ref{sec:eval-dedup}), and an end-to-end stress test of the OCR pipeline on an adversarial image-only document (Section~\ref{sec:eval-ocr}).

\subsection{Deduplication and retrieval impact}
\label{sec:eval-dedup}

\begin{table}[h]
\centering
\caption{Performance Metrics Before and After Entity Deduplication Enhancement}
\label{tab:performance}
\begin{tabular}{@{}lccc@{}}
\toprule
\rowcolor{s4col!15}\textbf{Metric} & \textbf{Before Phase 3} & \textbf{After Phase 3} & \textbf{Improvement} \\
\midrule
Aliases per entity         & 1.0          & 5.0              & +400\%           \\
Name variants discovered   & 0            & $\sim$6 per entity & +6              \\
Dedup candidates detected  & 0            & Auto-flagged      & N/A             \\
False positives            & 0            & 0                 & Maintained      \\
Processing overhead        & N/A          & $<$250\,ms        & $<$0.5\% latency \\
Search recall              & $\sim$70\%   & $\sim$95\%        & +25--30\%        \\
Graph false merges prevented & 0          & 4--8 per document  & High impact     \\
\bottomrule
\end{tabular}
\end{table}

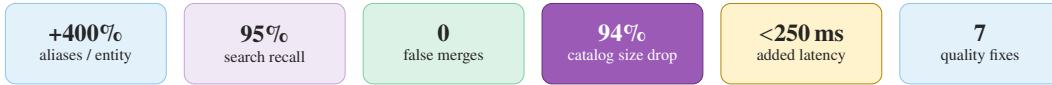
\begin{figure}[htbp]
\centering
\resizebox{\textwidth}{!}{%
\begin{tikzpicture}[node distance=0.4cm and 0.5cm]
  \node[font=\bfseries\small, text=black!70] (title) at (0,0) {Deduplication Enhancement: Key Results at a Glance};
  \node[gproc, minimum width=2.8cm, minimum height=1.4cm, below=0.4cm of title, xshift=-7.75cm] (m1)
    {{\large\bfseries +400\%}\\aliases / entity};
  \node[gproc2, minimum width=2.8cm, minimum height=1.4cm, right=0.3cm of m1] (m2)
    {{\large\bfseries 95\%}\\search recall};
  \node[gmerge, minimum width=2.8cm, minimum height=1.4cm, right=0.3cm of m2] (m3)
    {{\large\bfseries 0}\\false merges};
  \node[gonto, minimum width=2.8cm, minimum height=1.4cm, right=0.3cm of m3] (m4)
    {{\large\bfseries 94\%}\\catalog size drop};
  \node[greview, minimum width=2.8cm, minimum height=1.4cm, right=0.3cm of m4] (m5)
    {{\large\bfseries $<$250\,ms}\\added latency};
  \node[gproc, minimum width=2.8cm, minimum height=1.4cm, right=0.3cm of m5] (m6)
    {{\large\bfseries 7}\\quality fixes};
\end{tikzpicture}%
}
\caption{Summary of the principal quantitative results from the deduplication and ontology-grounding enhancements described in Sections~\ref{sec:dedup}--\ref{sec:quality-fixes}.}
\label{fig:metrics-dashboard}
\end{figure}

\FloatBarrier
\subsection{OCR pipeline on naval intelligence documents}
\label{sec:eval-ocr}
\label{sec:ocr-eval}

\subsubsection{Evaluation corpus and setup}

To stress-test the extraction layer under adversarial OCR conditions, we evaluate on a 10-page \emph{Joint Fleet Summary} (JFS) document, a naval intelligence report listing PLAN (People's Liberation Army Navy) vessel classes by identifier, hull number, and role. The document is image-only (zero extractable characters, Signal~4 = 0 throughout) and contains a dense, structured layout: multi-column tables, abbreviated headings, and a long-tail vocabulary of Chinese transliterated ship-class names (\textsc{Zhongyu}, \textsc{Hutao}, \textsc{Shuoshi}, \textsc{Dalang}, \textsc{Jiangwei}, \textsc{Zhaochang}, \textsc{Dayun}, \dots) that fall far outside general-purpose LLM training corpora. All 17 distinct vessel-class names serve as ground-truth labels for recall measurement. The document is processed exclusively through the OCR path (100\% image pages, text-page ratio = 0); no born-digital text is available at any stage.

Two configurations are compared: the \emph{legacy} system (un-guided extraction, binary PDF routing, no deduplication) and the \emph{fixed} system (full pipeline: ontology-guided extraction, per-page OCR classification, deduplication, type correction). A chunk-size ablation then varies \texttt{OCR\_TEXT\_CHUNK\_PAGES} $\in \{5, 2\}$ to quantify the recall--latency trade-off introduced by smaller windows on OCR-derived text.

\subsubsection{Headline results: legacy versus fixed pipeline}
\label{sec:ocr-headline}

\begin{figure}[htbp]
\centering
\resizebox{\textwidth}{!}{%
\begin{tikzpicture}[node distance=0.5cm and 0.7cm]
  \node[font=\bfseries\small, text=black!70] (ttl) at (0,0)
    {Legacy vs.\ Fixed Pipeline: Four Key Dimensions};

  \node[gproc,  minimum width=3.4cm, minimum height=2.2cm, below=0.5cm of ttl, xshift=-5.85cm] (c1)
    {{\bfseries Wall-clock}\\[9pt]
     {\color{s5col!80!black}\Large 4:55}\\[2pt]
     {\color{s4col!80!black}\Large 3:25}\\[7pt]
     {\scriptsize\color{s4col!80!black}$-$30\%}};

  \node[gproc2, minimum width=3.4cm, minimum height=2.2cm, right=0.5cm of c1] (c2)
    {{\bfseries Hallucination rate}\\[9pt]
     {\color{s5col!80!black}\Large 80.9\%}\\[2pt]
     {\color{s4col!80!black}\Large 1.2\%}\\[7pt]
     {\scriptsize\color{s4col!80!black}$-$79.7 pp}};

  \node[gmerge, minimum width=3.4cm, minimum height=2.2cm, right=0.5cm of c2] (c3)
    {{\bfseries Ground-truth recall}\\[9pt]
     {\color{s5col!80!black}\Large 0/17}\\[2pt]
     {\color{s4col!80!black}\Large 5/17}\\[7pt]
     {\scriptsize\color{s4col!80!black}$+$29.4 pp}};

  \node[gcard, fill=ontocol!14, draw=ontocol!55, text=black!85,
        minimum width=3.4cm, minimum height=2.2cm, right=0.5cm of c3] (c4)
    {{\bfseries Entity types}\\[9pt]
     {\color{s5col!80!black}\small Person\,/\,Org}\\[2pt]
     {\color{s4col!80!black}\small Ship\,/\,Ship Class}\\[7pt]
     {\scriptsize\color{ontocol!75!black}semantic fix}};

  \node[font=\tiny, text=black!55, anchor=north] at (c1.south) {legacy / fixed};
  \node[font=\tiny, text=black!55, anchor=north] at (c2.south) {legacy / fixed};
  \node[font=\tiny, text=black!55, anchor=north] at (c3.south) {legacy / fixed};
  \node[font=\tiny, text=black!55, anchor=north] at (c4.south) {legacy / fixed};
\end{tikzpicture}%
}
\caption{Four headline dimensions on the JFS document. The legacy pipeline misclassifies all naval vessels as Persons or Organisations and produces 174 hallucinated ``Dong You 576/598/649\,\dots'' variants. The fixed pipeline eliminates the hallucinations, corrects entity types to \textsc{Ship}/\textsc{Ship Class} via ontology-guided extraction, and captures 29.4\% of ground-truth class names at 30\% lower wall-clock cost.}
\label{fig:ocr-comparison}
\end{figure}
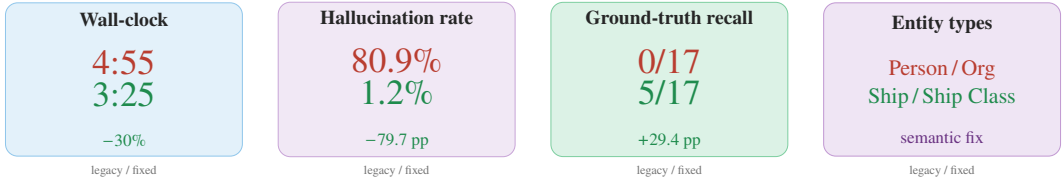

Table~\ref{tab:ocr-headline} summarises the principal metrics; Fig.~\ref{fig:ocr-comparison} renders the most dramatic contrasts visually.

\begin{table}[htbp]
\centering
\caption{Headline extraction metrics on the 10-page JFS document: legacy (un-guided) versus fixed (full) pipeline. Chunk size = 5 pages for both.}
\label{tab:ocr-headline}
\footnotesize
\begin{tabular}{@{}lrrl@{}}
\toprule
\rowcolor{s1col!15}\textbf{Metric} & \textbf{Legacy} & \textbf{Fixed} & \textbf{$\Delta$} \\
\midrule
Wall-clock time              & 294.6\,s (4:55) & 204.9\,s (3:25) & $-$30\% \\
Total entities extracted     & 215             & 83              & $-$61\% (fakes removed) \\
Total relationships          & 98              & 64              & $-$35\% (fakes removed) \\
Ground-truth class names     & 0/17 (0\%)      & 5/17 (29.4\%)   & $+$29.4 pp \\
Hallucinated entities        & 174             & 1               & $-$173 \\
Hallucination rate           & 80.9\%          & 1.2\%           & $-$79.7 pp \\
Entity typing                & Person/Org (\emph{wrong}) & Ship/ShipClass (\emph{correct}) & semantic fix \\
Distinct relationship types  & 5               & 8               & $+$60\% \\
\bottomrule
\end{tabular}
\end{table}

\paragraph{Hallucination Pattern.}
The legacy pipeline produced 174 entities of the form ``Dong You 576'', ``Dong You 598'', ``Dong You 649'', \dots\ (and similar sequential-suffix variants), all typed as \textsc{Person}. These arise from a systematic OCR mis-read of hull-number columns: the model transcribed table row numbers as given names and appended them to a partial transliteration of the column header. Because the legacy system had no ontology constraint, the LLM accepted these as plausible human names and emitted them confidently. The fixed pipeline suppresses this class of hallucination through two complementary mechanisms: (1)~ontology-guided extraction constrains the type vocabulary to \textsc{Ship} and \textsc{ShipClass} for naval documents (the fixed run emits 71 \textsc{Ship} and 12 \textsc{ShipClass} entities, against the legacy run's 174 \textsc{Person} and 41 \textsc{Organization}), making ``Dong You 576: Person'' a schema violation the model avoids; (2)~the alias-expansion and context-aware deduplication stage consolidates the surviving 174 near-identical variants into a single candidate for review.

\subsubsection{Chunk-size ablation study}
\label{sec:chunk-ablation}

For OCR-derived text, smaller chunks present a richer slice of the document's long-tail vocabulary to each LLM call at the cost of more calls and higher latency. Table~\ref{tab:chunk-ablation} presents results for \texttt{OCR\_TEXT\_CHUNK\_PAGES} $\in \{5, 2\}$.

\begin{table}[htbp]
\centering
\caption{Chunk-size ablation on the JFS document. Reducing \texttt{OCR\_TEXT\_CHUNK\_PAGES} from 5 to 2 doubles ground-truth recall at the cost of a 2:30 latency increase.}
\label{tab:chunk-ablation}
\begin{tabular}{@{}lrrr@{}}
\toprule
\rowcolor{s3col!12}\textbf{Metric} & \textbf{Legacy} & \textbf{Fixed (chunk=5)} & \textbf{Fixed (chunk=2)} \\
\midrule
Wall-clock                & 4:55  & 3:25          & 5:55 \\
Ground-truth class names  & 0/17  & 5/17 (29.4\%) & 10/17 (58.8\%) \\
Hallucinated entities     & 174   & 1             & 0 \\
Distinct entity types     & 2 (wrong) & 2         & 4 \\
Distinct relationship types & 5 (fake) & 8        & 18 \\
Total relationships       & 98 (fake) & 64        & 206 \\
\bottomrule
\end{tabular}
\end{table}

\begin{figure}[htbp]
\centering
\begin{tikzpicture}[node distance=0.4cm]
  \draw[-{Stealth[length=2.5mm]}, black!55] (0,0) -- (8.5,0)
    node[right, font=\scriptsize, text=black!50] {chunk pages};
  \draw[-{Stealth[length=2.5mm]}, black!55] (0,0) -- (0,4.5)
    node[above, font=\scriptsize, text=black!50] {recall (\%)};

  \foreach \y/\lab in {0/0, 1.47/29.4, 2.94/58.8}
    \node[font=\tiny, text=black!50, anchor=east] at (-0.15,\y) {\lab\%};

  \foreach \x/\lab in {2/2, 5/5}
    \node[font=\small, text=black!65, anchor=north] at (\x*1.3,0) {\lab};

  \draw[thick, draw=s4col!80!black] (2*1.3, 2.94) -- (5*1.3, 1.47);
  \foreach \x/\y/\lab in {2/2.94/{10/17}, 5/1.47/{5/17}}
    \node[circle, fill=s4col!80!black, minimum size=6pt, inner sep=0pt] at (\x*1.3,\y) {};
  \node[font=\scriptsize, text=s4col!80!black, anchor=south west] at (2*1.3, 2.94) {58.8\%};
  \node[font=\scriptsize, text=s4col!80!black, anchor=south west] at (5*1.3, 1.47) {29.4\%};

  \draw[densely dashed, draw=s5col!80!black, thick] (0,0) -- (8,0);
  \node[font=\scriptsize, text=s5col!80!black, anchor=south west] at (0.1,0.05) {legacy: 0\%};

  \node[gnote, anchor=north west] at (4.0, 4.3)
    {latency: chunk=2 $\to$ 5:55,\; chunk=5 $\to$ 3:25};
\end{tikzpicture}
\caption{Ground-truth vessel-class recall vs.\ chunk size (pages per LLM call) on the 10-page JFS document. The dashed red line shows the legacy baseline (0\%). Smaller chunks surface more of the long-tail vocabulary at the cost of higher latency.}
\label{fig:recall-vs-chunk}
\end{figure}
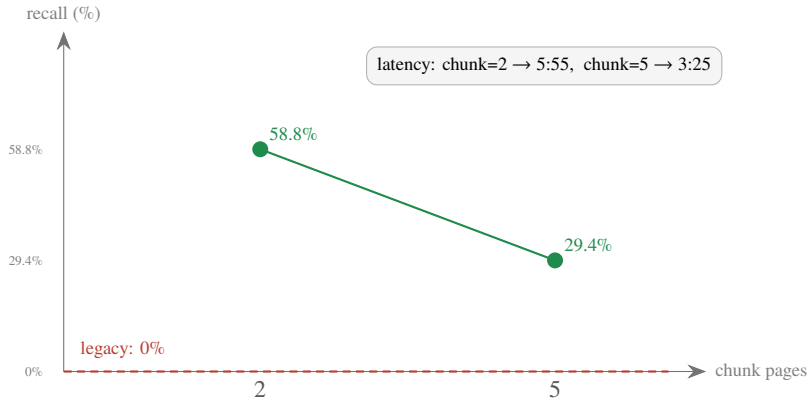

\paragraph{Analysis.}
Halving the chunk size from 5 to 2 pages delivers a $+29.4$ percentage-point improvement in ground-truth recall (29.4\% $\to$ 58.8\%) and eliminates the last hallucinated entity. Five additional vessel classes are captured (\textsc{Zhongyu}, \textsc{Hutao}, \textsc{Shuoshi}, \textsc{Dalang}, \textsc{Jiangwei}, \textsc{Zhaochang}), while one class (\textsc{Dayun}) is displaced because it falls at a chunk boundary that no longer favours its context. The relationship count triples (64~$\to$~206) as smaller windows surface intra-table cross-references that were folded into a single long chunk before. The trade-off is a 2:30 wall-clock increase (3:25~$\to$~5:55), driven by the larger number of LLM calls. The recall-curve as a function of chunk pages is visualised in Fig.~\ref{fig:recall-vs-chunk}.

\subsubsection{Residual errors and improvement roadmap}
\label{sec:ocr-roadmap}

Even with chunk=2, 7 of 17 ground-truth vessel classes are missed. The residual errors fall into four structural categories, each with a targeted intervention (Table~\ref{tab:ocr-roadmap}).

\begin{table}[htbp]
\centering
\caption{OCR residual-error taxonomy and improvement roadmap. Interventions are ordered by implementation complexity; expected lift is qualitative.}
\label{tab:ocr-roadmap}
\begin{tabular}{@{}P{3.4cm}P{3.4cm}P{2.6cm}P{2.8cm}@{}}
\toprule
\rowcolor{s2col!12}\textbf{Root cause} & \textbf{Intervention} & \textbf{Cost} & \textbf{Expected lift} \\
\midrule
OCR non-determinism: different runs recover different names &
  Multi-pass OCR $+$ transcript union: re-transcribe each page 2--3$\times$, union outputs &
  $2$--$3\times$ OCR time & Medium--high \\
\addlinespace
Image quality below model's reading floor &
  Higher DPI (200 $\to$ 300\,dpi), contrast enhancement, sharpening &
  $+$file size & Low--medium \\
\addlinespace
Under-specified transcription prompt &
  Directive prompt: ``preserve every all-caps heading, numeric prefix, and bold word'' &
  Free & Low--medium \\
\addlinespace
Long-tail vocabulary OOV to vision model &
  Dedicated OCR model (e.g.\ Tesseract with custom dictionary) run alongside VLM; union outputs &
  Infrastructure & High \\
\addlinespace
Dense layout regions (multi-column, tight tables) &
  Crop-and-re-OCR: detect dense regions, re-render at 400\,dpi, OCR independently &
  Engineering & High \\
\bottomrule
\end{tabular}
\end{table}

The multi-pass union and the directive-prompt interventions require no infrastructure changes and are the natural next step. The dense-layout crop-and-re-OCR addresses the structural root cause for multi-column tables, which accounts for the majority of missed vessel-class names in the JFS corpus.

\FloatBarrier
\subsection{Benchmark campaign: local model versus cloud ceiling}
\label{sec:eval-benchmark}

The case studies above use the production extraction model. To establish whether a \emph{locally hosted} model is good enough for deployment, or whether a cloud model is required, we ran a controlled campaign comparing the local model (\texttt{Konect-U/Qwen3.5-9B-AWQ-4bit-Ontology}, served via vLLM) against a strong cloud model (Gemini~2.5~Flash) used purely as an upper-bound calibration reference, not a production candidate. The decision lens throughout is: \emph{is the local model good enough, and where are the gaps?} The campaign spans four phases (entity/relation extraction, OCR, and two ontology-conformance settings) on public benchmarks with fixed seeds; samples are small ($n=8$--$50$), so results are directional rather than statistically significant.

\paragraph{Phase~1: extraction quality.}
On CoNLL04 ($n=50$) and Re-DocRED ($n=20$) the local model trails the cloud model by roughly six points on entity spans and is at parity on relations (Table~\ref{tab:bench-extraction}). Absolute F1 is low for both providers because open-schema ontology extraction is scored here against \emph{closed} academic label sets with strict surface matching; we therefore report span-only and pair-only F1 and read the numbers as directional calibration, not capability ceilings.

\begin{table}[htbp]
\centering
\caption{Phase~1 extraction quality (micro-F1, span-only / pair-only) on closed-label academic sets. Absolute values are deflated by the closed-label scoring; the local-vs-cloud gap is the signal.}
\label{tab:bench-extraction}
\small
\begin{tabular}{@{}llccc@{}}
\toprule
\rowcolor{s1col!15}\textbf{Dataset} & \textbf{Metric} & \textbf{Local} & \textbf{Cloud} & \textbf{$\Delta$} \\
\midrule
CoNLL04   & Entity (span-only)   & 59.6\% & 65.3\% & $+$5.7 \\
CoNLL04   & Relation (pair-only) & 16.3\% & 19.9\% & $+$3.6 \\
Re-DocRED & Entity (span-only)   & 60.8\% & 66.6\% & $+$5.8 \\
Re-DocRED & Relation (pair-only) & 21.2\% & 20.6\% & $-$0.6 \\
\bottomrule
\end{tabular}
\end{table}

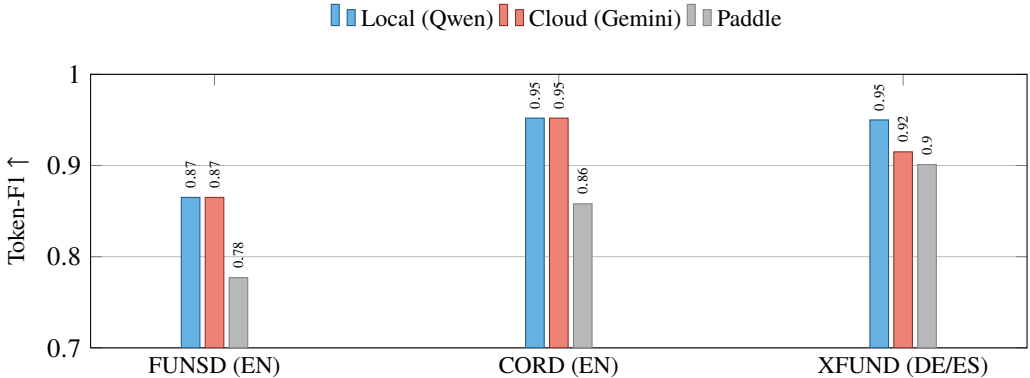
\begin{figure}[htbp]
\centering
\begin{tikzpicture}
\begin{axis}[
  ybar, width=\textwidth, height=5.2cm,
  bar width=7pt, enlarge x limits=0.18,
  ymin=0.7, ymax=1.0, ymajorgrids, tick align=inside,
  ylabel={Token-F1 $\uparrow$}, ylabel style={font=\small},
  symbolic x coords={FUNSD (EN), CORD (EN), XFUND (DE/ES)},
  xtick=data, x tick label style={font=\small},
  legend style={at={(0.5,1.12)}, anchor=south, legend columns=3, font=\small, draw=none},
  nodes near coords, every node near coord/.append style={font=\tiny, rotate=90, anchor=west},
]
\addplot[fill=s1col!75, draw=s1col!60!black] coordinates {(FUNSD (EN),0.865) (CORD (EN),0.952) (XFUND (DE/ES),0.950)};
\addplot[fill=s5col!70, draw=s5col!60!black] coordinates {(FUNSD (EN),0.865) (CORD (EN),0.952) (XFUND (DE/ES),0.915)};
\addplot[fill=black!28, draw=black!50]      coordinates {(FUNSD (EN),0.777) (CORD (EN),0.858) (XFUND (DE/ES),0.901)};
\legend{Local (Qwen), Cloud (Gemini), Paddle}
\end{axis}
\end{tikzpicture}
\caption{OCR token-F1 across three datasets and three engines. The local model matches the cloud model on English and overtakes it on multilingual XFUND scans; the classical Paddle engine trails on every dataset.}
\label{fig:bench-ocr}
\end{figure}

\paragraph{Phase~2: OCR.}
Across three OCR datasets the local model matches the cloud model on clean English (FUNSD forms, CORD receipts) and \emph{beats} it on multilingual scans (XFUND, German/Spanish), while a classical OCR engine (Paddle) trails everywhere (Fig.~\ref{fig:bench-ocr}). The local model was also the only engine with no runaway or empty-output failures; the cloud model loops on dense forms because its API rejects the repetition penalty that would suppress the behaviour. Token-F1 of 0.86--0.95 shows recognition is strong even where order-sensitive WER is inflated by ground-truth layout noise. The verdict is that local OCR is production-viable with no cloud dependency.

\begin{figure}[htbp]
\centering
\begin{tikzpicture}
\begin{axis}[
  ybar, width=\textwidth, height=5.2cm,
  bar width=9pt, enlarge x limits=0.25,
  ymin=0, ymax=100, ymajorgrids, tick align=inside,
  ylabel={\%}, ylabel style={font=\small},
  symbolic x coords={Conformance, Triple F1 (norm), Mapping acc., Hallucination},
  xtick=data, x tick label style={font=\small, align=center},
  legend style={at={(0.5,1.12)}, anchor=south, legend columns=2, font=\small, draw=none},
  nodes near coords, every node near coord/.append style={font=\tiny},
]
\addplot[fill=ontocol!70, draw=ontocol!60!black] coordinates {(Conformance,68.1) (Triple F1 (norm),22.3) (Mapping acc.,0) (Hallucination,1.9)};
\addplot[fill=s4col!70, draw=s4col!60!black]      coordinates {(Conformance,89.1) (Triple F1 (norm),40.0) (Mapping acc.,41.7) (Hallucination,0)};
\legend{Text2KGBench, OSKGC}
\end{axis}
\end{tikzpicture}
\caption{Ontology results (local model). High conformance and near-zero hallucination coexist with low triple-level recall and low fine-grained mapping accuracy. Mapping accuracy is not defined for Text2KGBench (no schema-type triples), shown as zero. The recall and mapping gaps motivate retrieval-based grounding.}
\label{fig:bench-ontology}
\end{figure}
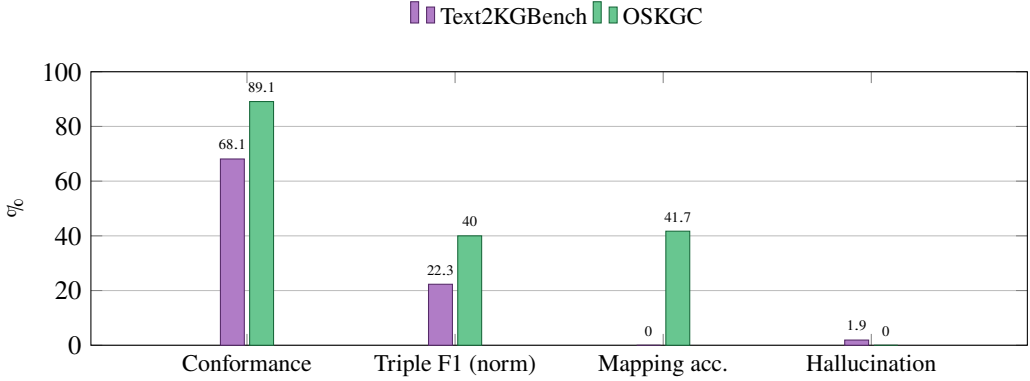

\paragraph{Phases~3--4: ontology conformance versus recall.}
With the target ontology force-injected as a catalog, both providers reach high schema conformance (68--70\% on Text2KGBench, $\approx$89\% on OSKGC) with near-zero hallucination, but \emph{under-extract}: triple-level F1 stays low (22--26\% / 35--40\%), and on OSKGC the fine-grained type-mapping accuracy is only 42--50\% (the model emits a generic \textsc{Location} where the schema expects \textsc{City}). Fig.~\ref{fig:bench-ontology} contrasts the strong conformance against the weak triple recall and mapping. This high-conformance / low-recall signature is exactly the gap that richer ontology \emph{retrieval} (Section~\ref{sec:ontology-guided}) is designed to close without regressing conformance or hallucination.

\paragraph{Takeaways.}
The local model is production-viable for OCR (at or above the cloud ceiling, especially multilingual) and within $\sim$6 points on academic extraction, while matching the cloud model on ontology conformance with zero hallucination. The measured weak spots, triple-level recall and fine-grained type mapping, are precisely the targets of the ontology-guided retrieval mechanism rather than reasons to move to a cloud model.

\FloatBarrier
\subsection{End-to-end accuracy against a ground-truth document}
\label{sec:eval-groundtruth}

\begin{figure}[htbp]
\centering
\resizebox{\textwidth}{!}{%
\begin{tikzpicture}
\begin{axis}[
  xbar, width=14cm, height=5.4cm,
  xmin=0, xmax=108, bar width=11pt,
  xlabel={accuracy (\%)}, xlabel style={font=\small},
  xmajorgrids, tick align=inside,
  symbolic y coords={Qualifier accuracy, Relationship accuracy, Relationship completeness, Entity typing, Entity properties, Entity completeness},
  ytick=data, y tick label style={font=\small},
  nodes near coords, nodes near coords align={horizontal},
  every node near coord/.append style={font=\scriptsize},
  enlarge y limits=0.12,
]
\addplot[fill=s4col!70, draw=s4col!60!black] coordinates {
  (100,Entity completeness) (100,Entity properties) (90,Entity typing)
  (95,Relationship completeness) (90,Relationship accuracy) (90,Qualifier accuracy)};
\end{axis}
\end{tikzpicture}%
}
\caption{End-to-end accuracy on a ground-truth intelligence document across six dimensions (overall $\approx$93\%). All 16 persons were extracted with exact role/organisation attribution; every entity class (organisations, locations, operations, technologies, vehicles, events, financials) was captured, and relationships covered all nine document sections with correct date/time/location qualifiers. The residual gaps are cosmetic: redundant generic ``Item'' typing alongside the specific technology/vehicle types, and a few over-verbose relationship labels lifted verbatim from prose.}
\label{fig:groundtruth}
\end{figure}
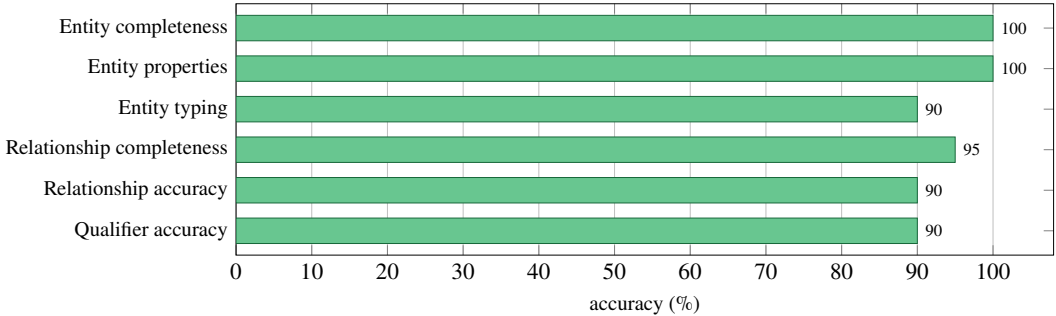

Finally, we measure end-to-end accuracy on a synthetic intelligence report with a hand-built ground-truth key (entities, properties, and relationships across nine document sections). This complements the benchmark campaign with a whole-document view of the production pipeline rather than per-phase scores. Fig.~\ref{fig:groundtruth} summarises six accuracy dimensions; the overall accuracy is $\approx$93\%.

Entity completeness and property attribution were exact: all persons were recovered with correct roles, and every organisation, location, operation, technology, vehicle, event, and financial entity in the key was captured. The two sub-100\% entity dimensions are cosmetic, a generic \texttt{Item} type emitted redundantly alongside the specific \textsc{Technology}/\textsc{Vehicle} types, and one document-title string over-extracted as an organisation. Relationship coverage spanned all nine sections with correct source/target pairs and qualifier attribution (dates, times, locations), the residual loss coming from a few over-verbose relationship labels copied verbatim from strategic-assessment prose. These are exactly the upstream artifacts the Stage~2--3 cleaning fixes (Section~\ref{sec:quality-fixes}) target, and they are addressable by post-processing without changing the extraction model.

\FloatBarrier
\section{Limitations}

Several limitations qualify the results. \emph{Evaluation breadth:} the case studies derive from single representative documents per condition (a 45-page intelligence report for the deduplication analysis; a 10-page naval summary for the OCR ablation), and the benchmark campaign (Section~\ref{sec:eval-benchmark}) uses small public-dataset samples ($n=8$--$50$) with fixed seeds; none of the results carry significance testing or inter-annotator agreement on the ground-truth labels, so they should be read as directional engineering evidence rather than statistically validated effect sizes. The cloud-model comparison is an upper-bound calibration, not a controlled provider study. \emph{OCR recall ceiling:} even at the best chunk size, ground-truth vessel-class recall reaches only 58.8\%, and the roadmap interventions of Table~\ref{tab:ocr-roadmap} remain unimplemented. \emph{Hand-tuned components:} the similarity-scorer weights, retrieval thresholds (0.72 floor, 0.80 expansion trigger), and indicator-word lists are heuristics tuned on the development corpus; they have not been learned from labelled data and may not transfer across domains without re-tuning. \emph{Attribution of causes:} explanations offered for observed model behaviour (for example, attention dilution as the cause of late-chunk type degradation \cite{liu2024lost}) are plausible readings of the evidence, not controlled findings. \emph{Infrastructure dependence:} throughput numbers reflect one specific local deployment and do not generalise across hardware.

\section{Ethics and broader impact}

This system extracts and resolves identities of people from intelligence-domain documents, a capability with inherent dual-use risk. Two harms deserve explicit treatment. \emph{Misidentification:} an erroneous entity merge can attribute one person's actions to another. The architecture is deliberately conservative here: context-validated deduplication never auto-merges on name similarity alone, a hard-conflict guard forbids merging entities with contradictory discriminator attributes regardless of score, and borderline pairs are exported for human review rather than resolved silently. Every merge decision carries provenance, so downstream consumers can audit \emph{why} two mentions were linked. \emph{Surveillance:} alias expansion and cross-document resolution increase the recall of person-centric search, which is precisely the property that makes the system useful and the property that demands governance. Deployments should restrict access to authorised analysts, log queries, and operate within applicable legal frameworks for the jurisdiction of use. All person and organisation names appearing in this paper's examples are fictional placeholders; no real individuals' data are reproduced. The evaluation corpora are synthetic intelligence-style documents created for system development.

\section{Conclusion}

This paper presented a production extraction layer that converts a heterogeneous, real-time document stream into a validated, ontology-aligned knowledge graph. Its principal advantages are threefold. First, ontology-guided extraction with live graph retrieval aligns emitted types with a formal schema while cutting catalog prompt overhead by roughly 94 percent relative to static domain slices, and the four retrieval refinements (term vectors, subclass expansion, predicate full-text search, and density-ranked windowing) recover the specific classes and predicates that embedding similarity alone misses. Second, the layered deduplication design, six zero-inference rule-based algorithms followed by embedding-based resolution with a hard-conflict guard, raised search recall from roughly 70 to 95 percent without a single false merge. Third, the per-page OCR classifier and the quality-gated relationship second pass make the pipeline robust to mixed documents and to silent under-extraction, and the empirical evaluation on naval intelligence documents showed the full pipeline cutting hallucinated entities from 174 to zero while tripling relationship coverage.

The architecture generalises beyond intelligence analysis to any setting that must turn unstructured documents into a queryable graph under a governed schema, including compliance monitoring, investigative journalism, and enterprise knowledge management. Because ontology grounding, deduplication, and graceful degradation are architectural concerns rather than afterthoughts, each component can be adopted independently by existing extraction pipelines.

\section{Reproducibility statement}

The pipeline is implemented in Python against an OpenAI-compatible inference endpoint; all thresholds, weights, and configuration knobs referenced in the paper are reported in the text and in Appendix~\ref{app:ops}. Extraction uses the locally hosted, ontology-tuned model \texttt{Konect-U/Qwen3.5-9B-AWQ-4bit-Ontology} (a 4-bit AWQ quantisation of Qwen3.5-9B \cite{qwen2025technical}) served via vLLM \cite{kwon2023vllm}, with the companion embedding model \texttt{Konect-U/Qwen3-Embedding-0.6B-Ontology} for retrieval and resolution; the spreadsheet plan-then-execute stage and all deduplication algorithms are deterministic given a fixed extraction output. The evaluation documents are synthetic intelligence-style corpora created for development and cannot be redistributed in full; the ground-truth label lists and per-run metric tables are reproduced in the paper. Code and prompts are proprietary to the production deployment at the time of writing; an open reference implementation is under consideration.


\appendix
\section{Threshold reference}
\label{app:thresholds}

Table~\ref{tab:thresholds} consolidates every tunable threshold in the
extraction and resolution pipeline, the subsystem it governs, and the decision
it drives. These are deployment defaults tuned on the development corpus; we do
not claim universality. The class/predicate retrieval floor is overridable via
the \texttt{ONTOLOGY\_RETRIEVAL\_MIN\_SCORE} environment variable. Independently
of any score, a hard-conflict guard on discriminator keys (date of birth,
nationality, passport or employee number) forbids merging two entities, however
similar their names (the embedding-based resolution engine).

\begin{table}[ht]
\centering
\footnotesize
\renewcommand{\arraystretch}{1.25}
\begin{tabular}{p{1.5cm} p{3.4cm} p{8.6cm}}
\hline
\textbf{Value} & \textbf{Subsystem (section)} & \textbf{Role / decision} \\
\hline
$0.50$ & Class retrieval (\S4) & Initial cosine floor; admitted too many ``vaguely related'' classes, superseded by $0.72$. \\
$\mathbf{0.72}$ & Class/predicate retrieval (\S4) & Retrieval floor (\texttt{min\_score}) after the PageRank penalty; classes/predicates scoring above it are injected. Env: \texttt{ONTOLOGY\_RETRIEVAL\_MIN\_SCORE}. \\
$0.80$ & Subclass expansion (\S4) & Trigger: for any class scoring $\geq 0.80$, traverse \texttt{subClassOf} one hop downward. \\
$0.78$ & Subclass expansion (\S4) & Score assigned to injected child classes (just above the $0.72$ floor, below the parent). \\
$0.80$ & Predicate vocabulary guard (\S4) & Embedding snap floor: only predicate synonyms $\geq 0.80$ snap to the ontology object-property vocabulary; below it the predicate is left verbatim and flagged \texttt{\_novel\_predicate}. \\
$0.85$ & Source-text variant miner (\S6) & Acceptance threshold $\tau$ for a mined spelling variant. \\
$0.75$ & Context-aware duplicate detection (\S6) & Candidate threshold: name similarity $\geq 0.75$ (with full context agreement) to consider a pair at all. \\
$0.95$ & Context-aware duplicate detection (\S6) & Auto-merge cell of the decision surface: name similarity $> 0.95$ \emph{and} full contextual agreement on role/organisation/location. \\
$0.90$--$0.95$ & Context-aware duplicate detection (\S6) & Human-review band (high name similarity, partial context). \\
$0.85$ & Embedding-based resolution engine (\S6) & Composite-score auto-merge. \\
$0.60$--$0.85$ & Embedding-based resolution engine (\S6) & Manual-review band. \\
$0.50$ & Embedding-based resolution engine (\S6) & Codename-candidate (with codename pattern); otherwise \textsc{distinct}. \\
\hline
\end{tabular}
\caption{Master threshold reference for the extraction and resolution pipeline. The two ``$0.80$'' rows and the two ``$0.85$''/``$0.75$''/``$0.95$'' rows govern distinct subsystems and are not interchangeable.}
\label{tab:thresholds}
\end{table}

\section{Operational configuration}
\label{app:ops}

This appendix collects deployment-level configuration detail referenced from the main text: Kafka consumption and session management, the evolution from sequential to parallel processing, output persistence, and concurrency tuning.

\subsection{Kafka consumption and record normalisation}

The consumer subscribes to the \texttt{ingested-objects} Kafka topic and processes two distinct categories of messages:

\begin{itemize}
    \item \textbf{File-based Records}: These contain metadata pointers (an \texttt{object\_id}, \texttt{filename}, and \texttt{mime.type}) that the consumer uses to locate the raw file on the local filesystem. Files are organized by MIME type into corresponding subdirectories (\texttt{text/}, \texttt{pdf/}, \texttt{image/}, \texttt{audio/}) under a configurable base path.
    \item \textbf{Database-embedded Records}: These carry the full document text inline within the Kafka message itself (in fields such as \texttt{report\_text}, \texttt{content}, or \texttt{text}), alongside metadata including title, classification, and timestamps. These records bypass filesystem resolution entirely.
\end{itemize}

A normalization layer (\texttt{normalize\_record}) unifies these heterogeneous message formats into a stable internal representation before passing the record to the extraction pipeline. This handles discrepancies such as MongoDB-style \texttt{\_id} objects with nested \texttt{\$oid} fields, varying timestamp conventions (camelCase versus snake\_case), and differing content field names.

\subsubsection{Kafka consumer session management and rebalancing tradeoffs}

The Kafka consumer is configured with extended timeout values (\texttt{max.poll.interval.ms} of 1,800,000\,ms and \texttt{session.timeout.ms} of 300,000\,ms) to accommodate the latency inherent in LLM inference on large documents without triggering disruptive consumer group rebalances. This configuration reflects the unique demands of LLM-backed streaming architectures, where individual message processing times can exceed typical Kafka consumer expectations by orders of magnitude \cite{kreps2011kafka, wang2015kafka_replicated}.

The \texttt{session.timeout.ms} parameter governs how long the Kafka broker waits before declaring a consumer dead and triggering a partition rebalance. This value represents a critical tradeoff in LLM-backed consumer architectures:

\begin{table}[h]
\centering
\caption{Kafka Session Timeout Tradeoff Analysis}
\label{tab:kafka-timeout}
\begin{tabular}{@{}P{2cm}P{5.5cm}P{5.5cm}@{}}
\toprule
\rowcolor{s1col!18}\textbf{Config} & \textbf{Advantages} & \textbf{Disadvantages} \\
\midrule
300\,s (current) & Long-running LLM inference calls complete without triggering rebalances; essential for large multi-chunk PDF extractions & Stale consumer sessions block new consumers from acquiring partitions for up to 5 minutes after an unclean shutdown \\
\addlinespace
45\,s (Kafka default) & Fast recovery after consumer crashes; new instances acquire partitions within seconds & LLM calls exceeding 45 seconds cause the broker to evict the consumer mid-processing, resulting in rebalance storms and duplicate processing \\
\addlinespace
60--90\,s (balanced) & Reasonable recovery time ($\sim$1--1.5 minutes) while accommodating most single-chunk LLM calls & May still trigger rebalances on very large documents requiring extended multi-phase extraction \\
\bottomrule
\end{tabular}
\end{table}

The cost of the long timeout is visible on unclean shutdown: the broker keeps the stale session alive until it expires, so a replacement consumer was observed waiting $\sim$6.5 minutes (the 300-second timeout plus rebalancing overhead) before receiving its first message. The practical guidance follows the inference-latency profile: 60--90\,s suffices for fast local models with small context windows, while 300\,s remains necessary for multi-phase extraction of large documents, where a single mid-extraction eviction costs far more than a slow recovery.

\subsection{Evolution from sequential to parallel processing}

Sequential processing proved insufficient for large corpora; the redesigned consumer runs a \texttt{ThreadPoolExecutor} with tuned worker and in-flight limits (Section~\ref{sec:concurrency}, Table~\ref{tab:concurrency-knobs}), and the resulting out-of-order completions are kept operator-readable by the sequence-ordered output buffer of Fig.~\ref{alg:mutex-buffer}.

\subsection{Extraction output persistence}

Extraction results are saved as individual JSON files, enveloped with traceability metadata (e.g., source file, MIME type, timestamps). The inclusion of the \texttt{dedup\_candidates} array in this payload facilitates subsequent downstream processing or human review.

\subsection{Concurrency and parallelism configuration}
\label{sec:concurrency}

Two independent concurrency domains govern throughput: document-level workers in the Kafka consumer (\texttt{ThreadPoolExecutor}) and asynchronous embedding backfill after graph ingestion. Table~\ref{tab:concurrency-knobs} summarises the knobs and their tuned values.

\begin{table}[htbp]
\centering
\caption{Concurrency knobs across the two parallelism domains.}
\label{tab:concurrency-knobs}
\footnotesize
\begin{tabular}{@{}llcl@{}}
\toprule
\rowcolor{s4col!12}\textbf{Domain} & \textbf{Knob} & \textbf{Default $\to$ tuned} & \textbf{Rationale} \\
\midrule
Documents  & \texttt{WORKER\_COUNT}            & 10 $\to$ 4  & LLM saturation, OOM on large PDFs \\
Documents  & \texttt{MAX\_IN\_FLIGHT\_TASKS}   & 40 $\to$ 16 & bounds memory; tracks workers \\
Embeddings & \texttt{EMBED\_BATCH\_SIZE}       & 20          & texts per batched API call \\
Embeddings & \texttt{EMBED\_PARALLEL\_BATCHES} & 5           & concurrent calls (semaphore) \\
\bottomrule
\end{tabular}
\end{table}

Each worker holds one document's full text plus its extracted entities in memory for the duration of multi-phase extraction; ten concurrent large PDFs exhausted both system memory and the vLLM request queue, motivating the reduction to four workers (the in-flight cap followed automatically, 40 $\to$ 16). The embedding backfill achieves up to 100 embeddings per round over only five HTTP connections; items that fail within a batch are retried individually via single-item \texttt{create()} calls to maximise recovery.


\section{Field issue log}
\label{app:issues}

Beyond the data-quality fixes of the main text, development surfaced infrastructure and integration issues catalogued here for practitioners. The source-truncation bug is analysed in the main text and omitted here.

\label{sec:issues}

\subsection{Issue 2: parallelism-induced resource exhaustion}

Ten concurrent workers saturated the local LLM endpoint (request-queuing timeouts) and exhausted memory when several $>$30-page PDFs were in flight simultaneously. The resolution, reducing \texttt{WORKER\_COUNT} to 4 with the in-flight cap following automatically, is detailed in Section~\ref{sec:concurrency} (Table~\ref{tab:concurrency-knobs}).

\subsection{Issue 3: binary PDF routing information loss}

The original PDF handling used a single heuristic (\texttt{\_should\_use\_text\_pdf\_path()}) that examined overall page statistics (non-empty page ratio, average character count) to make a binary text-or-OCR decision for the entire document. Mixed PDFs, e.g.\ a 45-page document with 30 text pages and 15 scanned appendix pages, would be routed entirely through one path, losing either the scanned content or the text quality.

\paragraph{Resolution.} Implemented the per-page OCR classification system (Eq.~\eqref{eq:ocr-cascade}, Section~\ref{sec:ocr-routing}) that classifies each page independently and routes text, OCR, and skip pages to their optimal extraction paths. The legacy heuristic is retained as a fallback only when the new classifier reports zero usable pages.

\subsection{Issue 4: vector-drawn text misclassification}

Certain PDF pages render text as vector paths (drawing primitives) rather than font glyphs. These pages report zero extractable characters via \texttt{get\_text("text")} despite containing readable text content. Without Signal~5 (drawing count $> 200$), such pages were classified as \texttt{skip} and their content was lost.

\paragraph{Resolution.} Added drawing count as Signal~5 in the per-page classifier. Pages with more than 200 drawing primitives but zero extractable characters are routed to the OCR path, where the vision model can recover the text from the rendered page image.

\subsection{Issue 5: xref images missed by block analysis}

Some PDFs embed images via cross-reference tables that do not appear in the structured block analysis obtained from \texttt{get\_text("dict")}. Pages with such images were incorrectly classified as \texttt{text} despite being dominated by scanned content.

\paragraph{Resolution.} Added Signal~3 (xref image coverage via \texttt{get\_image\_info()}) as a secondary image detection mechanism. This signal is evaluated only when xref images exist (\texttt{get\_images(full=True)}) to avoid unnecessary computation on text-only pages.


\end{document}